\documentclass{article}

\PassOptionsToPackage{numbers, compress}{natbib}

\usepackage[preprint]{neurips_2025}

\usepackage[utf8]{inputenc} 
\usepackage[T1]{fontenc}    
\usepackage{hyperref}       
\usepackage{url}            
\usepackage{booktabs}       
\usepackage{amsfonts}       
\usepackage{nicefrac}       
\usepackage{microtype}      
\usepackage{xcolor}         
\usepackage{graphicx}

\usepackage{amsmath}
\usepackage{amsthm}
\usepackage{cleveref}
\usepackage{subcaption}
\usepackage{footmisc}
\usepackage{amsfonts}
\usepackage{booktabs}
\usepackage{cases}
\usepackage{yhmath}
\usepackage[algo2e,ruled]{algorithm2e} 

\title{Open Vocabulary Compositional Explanations for Neuron Alignment}

\author{%
  Biagio La Rosa \\
  Department of Computer Science and Engineering\\
  University of California, Santa Cruz\\
  \texttt{bilarosa@ucsc.edu} \\
  \And
  Leilani H. Gilpin \\
  Department of Computer Science and Engineering\\
  University of California, Santa Cruz\\
  \texttt{lgilpin@ucsc.edu} \\
}

\begin{document}

\maketitle

\begin{abstract}
Neurons are the fundamental building blocks of deep neural networks, and their interconnections allow AI to achieve unprecedented results. Motivated by the goal of understanding how neurons encode information, compositional explanations leverage logical relationships between concepts to express the spatial alignment between neuron activations and human knowledge.
However, these explanations rely on human-annotated datasets, restricting their applicability to specific domains and predefined concepts. This paper addresses this limitation by introducing a framework for the vision domain that allows users to probe neurons for arbitrary concepts and datasets. Specifically, the framework leverages masks generated by open vocabulary semantic segmentation to compute open vocabulary compositional explanations. The proposed framework consists of three steps: specifying arbitrary concepts, generating semantic segmentation masks using open vocabulary models, and deriving compositional explanations from these masks. The paper compares the proposed framework with previous methods for computing compositional explanations both in terms of quantitative metrics and human interpretability, analyzes the differences in explanations when shifting from human-annotated data to model-annotated data, and showcases the additional capabilities provided by the framework in terms of flexibility of the explanations with respect to the tasks and properties of interest. 
\end{abstract}
\section{Introduction}
\label{sec:introduction}
The black-box nature of deep neural networks (DNNs) remains an important limitation for their adoption in fields such as healthcare, finance, and autonomous systems, where understanding the rationale behind model behaviors is essential for trust and accountability~\cite{Ching2018}. In particular, the opacity of the learning process in DNNs makes it difficult to gain insights into what these models learn and to guarantee the correctness of their behavior. To address this problem, several works have focused on methods to study the knowledge encoded in DNNs and on what individual neurons learn during the training process. 
Among them, this paper focuses on methods that study the spatial alignment between neuron activations and human knowledge in the vision domain. These methods aim to understand whether artificial neurons encode and parse information similarly to humans. 
The state-of-the-art in this area is represented by compositional explanations~\cite{Mu2020}, which express the alignment between the locations of a given neuron activation range and the location of concepts through propositional logic formulas. For example,  ((Cat AND White) OR Dog) can be associated with a neuron whose activations overlap with the locations of white cats or dogs within the images. This approach has been improved over time, including more complex spatial relations~\cite{Harth2022}, knowledge bases~\cite{Massidda2023}, and multiple activation ranges~\cite{LaRosa2023Towards}.

Despite the progress, one of the main limitations of this family of methods is their dependency on concept-annotated datasets~\cite{Ramaswamy2023,Mu2020}. Specifically, for each concept, these explanations require annotations that identify its precise locations in all samples within the probing dataset. This annotation process is conducted by humans, making it both costly and prone to inconsistencies. On a practical level, only a limited number of concept-annotated datasets are available in the literature. This scarcity imposes several limitations, such as the closed-world assumption, where the model can only be evaluated on concepts present in these few datasets. 
Consequently, concepts that are not annotated or concepts with a different level of granularity may be ignored.

\textbf{This paper addresses the dependency on human annotations by proposing a framework for the vision domain that leverages open vocabulary semantic segmentation models}. These models have recently been proposed to segment \emph{any} object in images, even those not seen during training, by combining traditional segmentation architectures with foundational models~\cite{Radford2021}. Specifically, our framework is training-free and relies only on a user-specified list of concepts, without requiring any manual annotations. Based on these concepts, optionally organized into different concept sets, the proposed framework generates segmentation masks by using open vocabulary semantic segmentation models and computes compositional explanations based on the generated masks. This framework offers several advantages, such as enabling explanation generation independent of human-annotated data, supporting explanations at varying levels of granularity, improving explanations through iterative refinements, and compatibility with the open-world assumption, where there are no constraints on the concepts a user can probe the neurons for.  

In detail, the paper's contribution is threefold:
\begin{itemize}
    \item it proposes the first framework that supports open vocabulary compositional explanations in the vision domain. Compared to previous methods, the framework achieves comparable performance on datasets with human annotations, while also offering greater flexibility, and better quantitative and qualitative results on datasets without human annotations.
   \item It investigates the differences between explanations derived from human and model-annotated data and analyzes the sources of these differences in terms of misalignment and granularity levels.
    \item It showcases, through two application scenarios, the advantages of the proposed framework in supporting multiple explanation granularity levels and iterative improvement of explanations through refinements.
\end{itemize}
We will release the code upon acceptance at the following repository \url{https://github.com/aiea-lab/open_vocab_compositional_explanations}.
\section{Related Work}
\label{sec:related}
\paragraph{Open Vocabulary Semantic Segmentation} The task of semantic segmentation aims to identify semantic regions in an image based on predefined classes of interest. Open vocabulary semantic segmentation aims to achieve this goal by replacing pre-defined classes with textual descriptions, including ones not encountered during training ~\cite{liang2023open}. Existing approaches can be categorized into two main groups: zero-shot segmentation approaches~\cite{Bucher2019,Xian_2019_CVPR}, which typically rely on word embeddings to align image features with unseen classes, and approaches that leverage pre-trained multi-modal models~\cite{Radford2021} to encode both text and images in a shared embedding space and identify the combination of segmented regions and text that maximizes their alignment~\cite{li2022languagedriven,Xu_2023_CVPR,Liu_2024_CVPR,Xie_2024_CVPR}. Within the second group, we can further distinguish between two-stage approaches~\cite{liang2023open,Xu2022}, which first generate class-agnostic masks and then assign labels to them using multi-modal models, and end-to-end approaches~\cite{Cho2024CatSeg,Xu_2023_CVPR}, which integrate multi-modal models earlier in the pipeline to simultaneously identify regions of interest and assign labels. These approaches differ in the placement of the multi-modal model within the pipeline and the training procedures (e.g., alignment losses) used to adapt the models for the segmentation task. Our framework is agnostic to the specific open vocabulary segmentation model employed. However, in this paper, we focus primarily on end-to-end approaches, as they offer greater flexibility in adapting masks to different concept granularity (e.g., whole objects versus object parts).

\paragraph{Explanations for Neuron Alignment} 
In this paper, we focus on a specific family of neuron explanations: logic and alignment-based explanations. These explanations aim to find the combination of concepts that maximizes the \textbf{alignment between the locations of a given neuron activation range and the locations of those concepts}. These combinations aim to capture a high degree of polysemantic behavior (i.e., the phenomenon where neurons can fire for multiple unrelated concepts~\cite{elhage2022superposition}). The seminal work in this area is Network Dissection~\cite{Bau2017,bau2020units}, which has been extended by \cite{Mu2020}, leading to compositional explanations. These explanations map neuron activations to logical connections between recognized concepts, expressing relationships between them.  Relationships explored in the literature, typically expressed as logical operators, include co-occurrence~\cite{Bau2017}, exclusion~\cite{Mu2020}, relative position~\cite{Harth2022},  and hierarchy~\cite{Massidda2023}. 
Despite the progress, one of the main limitations of this family of methods is their dependency on concept-annotated datasets~\cite{Ramaswamy2023}, limiting their applicability~\cite{Mu2020} and making them costly in terms of human labor. The framework proposed in this paper falls into this last paradigm and addresses the dependency on human annotations. Our approach is related to \cite{bau2020units}, which employs a segmentation model trained on the probing dataset to identify the individual concept (among the ones it has been trained on) that maximizes the overlap between annotations and activations. Differently from them, we leverage open segmentation models, thus removing the requirement to train on the probing dataset, support multiple granularities, and extract logical combinations of concepts. The support for different concept granularity also generalizes the approach proposed in \cite{Massidda2023}, which leverages an ontology to infer partial annotations at a higher level of granularity (e.g., from ``\textit{cat}'' to ``\textit{animal}''). In contrast, our framework supports refinements in granularity toward both higher and lower levels.

\paragraph{Other Neuron and Concept-Based Explanations}
While this paper focuses on methods targeting neuron spatial alignment, the literature offers a wide range of explanation approaches that use concepts to derive explanations~\cite{Casper2023,Gilpin2018,LaRosa2024thesis} or to decode other types of neurons~\cite{hesse2025disentangling,Srinivas2025,bau2020units,O’Mahony2025,Bykov2023} and layer behaviors~\cite{gao2025scaling,bricken2023monosemanticity}.
Some recent studies have also explored the use of foundational models to enable open-vocabulary explanations, for instance within the Concept Bottleneck framework~\cite{oikarinenlabel,Tan2024} or in approaches that estimate correlations between concept presence and neuron activations~\cite{Oikarinen2023}.
Our work is inspired by this growing interest in open-vocabulary explanations. However, the similarities end there: these approaches belong to distinct families of interpretability methods that differ fundamentally from compositional explanations across multiple dimensions, such as their goal (e.g., alignment vs. correlation), scope (e.g., neuron clusters vs. layers), assumptions (e.g., access to internals and availability of masks), representation (e.g., logical vs. statistical explanations), and phase (post-hoc vs. ante-hoc).
Because these families are not able to measure the spatial alignment between activations and the locations of concepts, they have traditionally been regarded as complementary rather than competing categories with respect to compositional explanations. Consequently, they are not compared against them, and no established protocols exist for such comparisons.
Given these differences, and because our work specifically focuses on improving and extending compositional explanations without altering their established formulation, we leave the exploration of connections and complementarities with other interpretability paradigms for future research.

\section{Framework}
\label{sec:framework}
Let $\mathbb{D} = \{x_1, x_2, ..., x_n\}$ be a probing dataset, where each input image $\{x \in \mathbb{R}^{3,h,w}\}$ has (variable) height  $h$  and width $w$. Let $z$ be a neuron in a probed model. Let $\mathbb{C}$ be a concept set specified by the user, including concepts that may or may not be present in the probing dataset, and $\mathfrak{L}^n$ be the set of all possible logical connections of arity at maximum $n$ between concepts in the concept set $\mathbb{C}$, where concepts are chained by propositional logic connectives. Compositional explanations aim to assign to $z$ the logical combination $L \in \mathfrak{L}^n$  of concepts in $\mathbb{C}$ (e.g., ((Cat OR Dog) AND Brown)) that maximizes the alignment between the localization of a given neuron's activation range and the localization of the concepts within the probing dataset. The goal of our framework is to achieve this objective without requiring humans to manually annotate the location of every concept in the probing dataset while offering more flexibility to the user. We can distinguish three steps: identifying the concept set, generating segmentation masks, and generating compositional explanations.

\paragraph{Concept Set Identification.} In our framework, the concept set $\mathbb{C}$ corresponds to a collection of $m$ concept subsets 
\begin{equation}
\mathbb{C} = \{ C_1, \ldots, C_m \}
\end{equation}
where each subset $C_{k}$ consists of a list of $n_k$ concepts 
\begin{equation}
C_k = \{ c_{k,1}, \ldots, c_{k,n_k} \},\ \forall k \in \{1, \ldots, m\}
\end{equation}
\begin{equation}
\text{subject to} \quad C_i \cap C_j \emptyset,\ \forall i \neq j
\label{eq:diff}
\end{equation}
\Cref{eq:diff} represents the constrain that the concept sets do not share concept names, and it is necessary to avoid inconsistency in mask generation. The concepts are arbitrary and specified by the users.  Each concept subset can be used to describe different levels of concept granularity (e.g., object names, abstractness, colors, parts, shapes, etc.).

\paragraph{Masks Generation.} 
Given the probing dataset $\mathbb{D}$, a pretrained open vocabulary segmentation model $f(\cdot, \cdot)$, and a concept subset  $C_k \in \mathbb{C}$, the framework generates a set of segmentation masks 
\begin{equation}
    S_k = \{\ s^j, \forall{j \in \mathfrak{D}} : s^j = f(x^j,C_k)\}
\end{equation}
where each element in $s^j$ corresponds to the concept most likely represented by the pixel at the same position in $x$.
The specific operations performed by the function $f(\cdot , \cdot)$ depend on the implementation of the open vocabulary segmentation model. Our framework is agnostic with respect to this implementation. The only assumption is that $f(\cdot , \cdot)$ can assign an arbitrary specified concept to each pixel. 

To satisfy the requirement of the compositional explanation algorithm~\cite{Mu2020},  these masks are upsampled (or downsampled) to have the same dimensions. Each segmentation mask $s^j$ is then transformed into a set of binarized masks $M^j_{C_k}$, one for each concept $c \in C_k$:
\begin{equation}
   M^j_{C_k} = \{ b_s(s^j, c_{k,i}), \forall{c_{k,i} \in C_k}\}
\end{equation}

where $b_s(s^j, c_{k,i})$ is a function that returns a binary mask where the pixels assigned to the concept  $c_{k,i}$ are set to 1, and the others are set to 0. For each concept subset, the binarized masks are then grouped into a \textbf{single-granularity} binary mask set: \begin{equation}
    \mathbb{M}_{C_k} = \{ M^j_{C_k}, \forall{j \in \mathbb{D}} \}
\end{equation}

By aggregating the single-granularity sets for all of the desired granularities, we can obtain the \textbf{multi-granularity} binary masks set:
\begin{equation}
    \mathbb{M}_{all} = \{M_{C_k} , \forall{C_k \in \mathbb{C}}\}
\end{equation}

\paragraph{Alignment Computation.}
The first step to compute the alignment for a neuron $k$ is to collect its activations $A_k$ over the probing dataset:
\begin{equation}
    A_k = \{a_{k,j}, \forall{j \in \mathbb{D}}\}
\end{equation}
The shape of $a_{k,j}$ depends on the neuron type. In general, this shape differs from that of the input and segmentation masks, and an additional function is needed to project the activation into the proper dimensional space. Our framework is agnostic to the specific projection. In this paper, we follow the established literature on the topic~\cite{bau2020units,Mu2020} by considering bidimensional neurons in the convolutional layers and using bilinear interpolation to reshape the activations. 
Given $A_k$ we apply clustering as in ~\cite{LaRosa2023Towards} to split the activations into semantic regions and identify multiple activation ranges. Then, given an activation range $[\tau_i,\tau_l]$, the framework computes the binarized activations $\mathbb{A}$ as:
\begin{equation}
    \mathbb{A} = \{b_a(a_{k,j}, [\tau_i,\tau_l]), \forall{j \in \mathbb{D}}\}
\end{equation}
where $b_a(a_{k,j}, [\tau_i,\tau_l])$ is a function that sets to 1 all activation values within the specified range and to 0 otherwise.  

Finally, the framework computes compositional explanations by finding the concepts that maximize the alignment between the binarized masks $\mathbb{A}$ and the concepts' segmentation masks in $\mathbb{M}$. To compute these explanations, we apply the recently proposed algorithm~\cite{LaRosa2023Towards} based on a beam search guided by the MMESH spatial heuristic (see Appx. B for more details). This heuristic exploits bounding and inscribed boxes to accelerate the beam search. Formally, the algorithm identifies the label $L \in \mathfrak{L}^n$ that maximizes the following objective:

\begin{equation}
    \operatorname*{arg\,max}_{L \in \mathfrak{L}^n} IoU(L, \mathbb{A}, \mathbb{M})
\end{equation}
where the Intersection Over Union ($IoU$) measures the overlap between label annotations and neuron activations, and it is defined as:
\begin{equation}
    IoU(L, \mathbb{A}, \mathbb{M}) = \frac{|\mathbb{A} \cap \theta(\mathbb{M}, L)|}{|\mathbb{A} \cup \theta(\mathbb{M},L)|}
\label{eq:iou}
\end{equation}
and $\theta(\mathbb{M}, L)$ is a function that returns the logical combination of the masks in $\mathbb{M}$ of the concepts involved in the label $L$. Following \cite{Mu2020}, we consider AND, OR, and AND NOT as logical connectives,  computed by standard bitwise logical operators between the binary matrices in $\mathbb{M}$. Setting $\mathbb{M} = \mathbb{M}_{C_k}$ in \cref{eq:iou} results in single-granularity compositional explanations, equivalent to those computed in previous work, but based on model annotations instead of human ones. Conversely, setting $\mathbb{M}=\mathbb{M}_{all}$ enables the usage of concepts from all of the granularities. In this case, the algorithm automatically selects the granularity level that is most aligned with each neuron.

After inspecting the compositional explanations computed in this step, the user can \textbf{optionally refine} the concept set by adding or removing concepts of interest, thus providing more flexibility during the analysis. Since the framework treats the concept subsets as independent, it regenerates the masks only for the specific subsets affected by the refinement (i.e., those to which the concepts are added or removed).

\section{Experiments}
\label{sec:experiments}
This section introduces the experimental setup (\cref{sec:setup}),  evaluates the proposed framework (\cref{sec:quantitative}), and analyzes the difference between compositional explanations computed over human and model-annotated datasets (\cref{sec:analysis}).
\subsection{Setup}
\label{sec:setup}

In the following sections, we use CAT-Seg~\cite{Cho2024CatSeg} with its default parameters (Appx. I
) as the backbone open vocabulary segmentation model of our framework. However, our findings are independent of the specific model choice: in Appx. A, we show that alternative open-vocabulary segmentation models achieve similar results, underscoring the readiness of this research area for explainability tasks. We selected these specific models because they employ setups that enable a fair and meaningful analysis, and they are compatible with the technical requirements of compositional explanations (see Appx. A for further details). However, for specific tasks, settings, and real-world applications, stronger models, especially among foundational models~\cite{Kirillov_2023_ICCV,Li2024}, may be found in the literature to further improve the results reported in this paper (see Appx. A).  As competitors, we follow the previous literature and consider the following approaches to computing spatial alignment: the heuristic-guided approach proposed in \cite{LaRosa2023Towards}, which relies on human-annotated data, and the closed vocabulary approach proposed in \cite{bau2020units}.
The term ``closed vocabulary'' refers to segmentation models trained on a specific dataset and able to recognize only concepts included in that dataset. Differently from our framework, the user cannot specify the concepts of interest and this baseline will generate segmentation masks related exclusively to the concept dataset used during the training stage. 
For the closed vocabulary approach, we update their proposal by extending it to the compositional explanation case and replacing their model with a state-of-the-art segmentation model (Mask2Former~\cite{Cheng2021mask2former}) trained on COCO~\cite{Lin2014}. We do not include SAE or other open-vocabulary explanation methods (\Cref{sec:related}) as competitors, as they pursue different goals and are not designed to capture localization alignment. Evaluating them fairly would require substantial adaptations beyond the scope of this work.
 
All competitors share the same experimental settings and hyperparameters, selected as the best found by prior work (see Appx. I). Namely, we focus on the neurons of the last convolutional layer of the probed models, we set the maximum explanation length to 3 and the beam size to 5 as in  \cite{Mu2020}, and we use K-Means to identify five clusters (i.e., activation ranges) in the neuron activations, as in \cite{LaRosa2023Towards}. Regarding terminology, we associate a number with each cluster: the lower the number, the lower the activations included in that cluster.

\subsection{Quantitative and Qualitative Evaluation}
\label{sec:quantitative}
The first set of experiments evaluates our proposed framework by measuring the quality of its generated compositional explanations. Due to space constraints, we report only a subset of our experiments in this section. A more comprehensive evaluation across six additional human-annotated datasets (Appx. A.2), four alternative framework implementations (Appx. A.1), and two additional probed models (Appx. A.3) is included in Appendix A.
To measure explanation quality, we use the per-pixel metrics adopted by previous literature and a user study. In the main text, as quantitative metrics, we report \textit{IoU}, as defined in \cref{eq:iou}; Detection Accuracy~\cite{Makinwa2022} (\textit{DetAcc}), which quantifies the percentage of label annotations recognized within the activation range; and 
Activation Coverage~\cite{LaRosa2023Towards} (\textit{ActCov}), which measures the percentage of neuron activations within the annotated label regions. Additional quantitative metrics and their results, as well as further details about these metrics, can be found in Appx. A.4. As qualitative metrics, to measure human interpretability, we ask participants in the user study to rate the alignment, precision, and relevance of the explanations computed by all the competitors. Given a randomly sampled set of activation masks produced by a neuron within a specific activation range, we define a concept as aligned if it appears in at least a subset of the activated masks; precise if its level of granularity matches that of the concepts included in the activation masks; and relevant if it is perceived as discriminative for the given task.

\begin{table}[!t]
\caption{Avg. scores for compositional explanations computed by the competitors for a model trained on the Place365 dataset probed using Ade20K.}
   
    \label{tab:ade20k}
    \centering
    \begin{tabular}{cllll}
        \toprule
                 Cluster & Method & \multicolumn{3}{c}{Place365}
       \\
          & & IoU & ActCov &DetAcc\\
         \midrule

         1 & Human \cite{LaRosa2023Towards}
         & 0.219 &  0.352 & 0.369 \\
          & Closed \cite{bau2020units} &
        0.215  & 0.341 & 0.368  \\
          & Ours &
        0.212 & 0.327 & 0.376  \\

         \midrule
             2 & Human \cite{LaRosa2023Towards} &
         0.132 &  0.322 & 0.184 \\
                 & Closed \cite{bau2020units} & 
        0.130 & 0.306  &  0.187  \\
& Ours &
         0.130  & 0.302 & 0.188  \\

         \midrule
             3 & Human \cite{LaRosa2023Towards} &
         0.102 & 0.276 & 0.148 \\
         & Closed \cite{bau2020units} &
        0.106   &
        0.272 &  0.155 \\

         & Ours &
         0.130  & 0.302 & 0.188\\
         
         \midrule
             4 & Human \cite{LaRosa2023Towards} &
         0.083  &  0.226 & 0.139\\
                  
                  & Closed \cite{bau2020units} &
         0.090 &
         0.241 & 0.140 \\
        
        & Ours &
         0.090 &   0.235 & 0.148  \\

         \midrule
           5 & Human \cite{LaRosa2023Towards} &
                  0.070  &   0.183 & 0.137 \\
                 
        & Closed \cite{bau2020units} &
         0.065  &  0.213 & 0.109  \\
        
        & Ours &
         0.079   &  
         0.214 & 0.139   \\
        
         \bottomrule
    \end{tabular}
\end{table}

\begin{table}[!t]
          \caption{Avg. scores for compositional explanations computed by the competitors for a model trained on CUB.}
   
    \label{tab:cub}
    \centering
    \begin{tabular}{cllll}
        \toprule
                 Cluster & Method &
        IoU & ActCov &DetAcc\\
         \midrule
1 & Human \cite{LaRosa2023Towards}
       &
     -	 &	-	&	-\\
          & Human \cite{LaRosa2023Towards}\textsubscript{Ade20k} 
       &
     0.248	 &	0.356	&	0.451\\
          & Closed \cite{bau2020units} &
        
0.388 &	0.635	&	0.501 \\

          & Ours &

        0.357 &	0.553	&	0.504 \\

         \midrule
2 & Human \cite{LaRosa2023Towards}
       &
     -	 &	-	&	-\\
              & Human \cite{LaRosa2023Towards}\textsubscript{Ade20k} &

       0.130	 &	0.312	 &	0.185 \\
                 & Closed \cite{bau2020units} &

  0.170	  &	0.505	&	0.214\\
& Ours &
      
        0.173	&	0.463	 &	0.221  \\

         \midrule
         3 & Human \cite{LaRosa2023Towards}
       &
     -	 &	-	&	-\\
              & Human \cite{LaRosa2023Towards} &

      0.085	 &	0.228 &	0.126 \\
         & Closed \cite{bau2020units} &

      0.142	 &	0.453	 &	0.175\\

         & Ours &

        0.147	&	0.432	  &	0.185\\
         
         \midrule
         4 & Human \cite{LaRosa2023Towards}
       &
     -	 &	-	&	-\\
              & Human \cite{LaRosa2023Towards}\textsubscript{Ade20k} &

        0.063	 &	0.167	 &	0.105\\
                  
                  & Closed \cite{bau2020units} &

         0.091	 &	0.571	 &	0.100\\
        
        & Ours &

        0.113	&	0.356	 &	0.147 \\

         \midrule
         5 & Human \cite{LaRosa2023Towards}
       &
     -	 &	-	&	-\\
           & Human \cite{LaRosa2023Towards}\textsubscript{Ade20k} &

               0.052	 &	0.144	 &	0.100 \\
                 
        & Closed \cite{bau2020units} &
        
        0.029	 &	0.674	 &	0.033 \\
        
        & Ours &
        
       0.077	 &	0.188	 &	0.131 \\
        
         \bottomrule
    \end{tabular}
\end{table}

We begin our analysis by comparing quantitatively compositional explanations for 512 neurons in a ResNet18~\cite{He2016} model trained on Place365~\cite{zhou2017places}. 
In this first experiment, we use the validation split of Ade20k~\cite{Zhou2017} as a probing dataset because it has been extensively used in literature to evaluate both compositional explanations and segmentation models and it includes human annotations. We use these annotations as masks for \cite{LaRosa2023Towards} and their labels as a concept set for our framework. The goal of this experiment is to investigate whether there is a \textit{degradation} in explanation quality when transitioning from human-annotated data to model-annotated data. This potential degradation could arise due to imprecision in the segmentation masks returned by the models or errors in the masks' labeling process.  As shown in \cref{tab:ade20k}, our framework achieves comparable or better average scores (with Std. Dev. reported in Appx. A.1
)
than the competitors across all of the activation ranges (i.e., clusters) but the lowest activations (Cluster 1), where scores are slightly worse. However, as noted by \cite{LaRosa2023Towards}, the lowest clusters often include fixed (uninformative) compositional explanations where the algorithm converges when no alignment is observed.  In such cases, the explanations generated by different competitors differ by only one concept within these degenerate explanations (i.e., \cite{LaRosa2023Towards} converges on ``building'' while our framework converges on ``person''), rendering the differences insignificant. Consequently, we consider the results in these settings satisfactory, and \textbf{we do not observe any significant degradation in explanation quality when using model-annotated data to compute compositional explanations}. Although \cite{LaRosa2023Towards} is applicable when the dataset includes human annotations, our framework remains useful in these scenarios for generating explanations at a different granularity and providing a deeper and more flexible interpretation.

\Cref{tab:cub} shows the results for 2048 neurons in a ResNet50 model~\cite{Song2021} trained on CUB~\cite{WahCUB_200_2011} for bird species classification and using the validation split of CUB as a probing dataset. This setting represents the task our framework is targeting: we consider the case where no human-annotated relevant masks are available\footnote{As a result, we do not include the additional data provided by \cite{Farrell2022} in our experiments.}. For our framework, we identify a multi-granularity concept set obtained through refinements and task-specific information (see Appx. D
). Because there are no human-annotated data, \cite{LaRosa2023Towards} could not be applied, and our framework aims to address this limitation. However, one could alternatively attempt to probe the model using a different dataset where annotations are available. To explore this, we consider using Ade20K as a probing dataset for their approach (Human$_{Ade20k}$). While this provides a point of comparison, we argue this strategy is not optimal and should be avoided due to several drawbacks (e.g., hallucinations and concept misalignment).
In this case, \textbf{our framework represents a significantly better choice} than alternatives, particularly in the highest clusters in terms of IoU and DetAcc. A qualitative analysis reveals even more significant differences. For \cite{LaRosa2023Towards}, compositional explanations are often computed over hallucinations of the probed model when parsing objects in Ade20K not available in CUB (i.e., the dataset used to train the probed model), leading to artifact alignments. This issue is evident when inspecting the most aligned concepts in the highest cluster, where we observed hallucinated concepts such as ``pool table'' (IoU=0.328) and ``car'' (IoU=0.22), which are absent and not relevant in CUB. These findings confirm the limitations of the human-based approaches when applied to datasets lacking annotations. Regarding the \textit{Closed} approach~\cite{bau2020units}, it achieves reasonable IoU scores in lower activation ranges because those ranges capture general concepts (e.g., water, sky), which are shared between CUB  and COCO~\cite{Lin2014}, the dataset used to train this approach. However, in the higher clusters, Closed compositional explanations are associated with abnormally high ActCov and low DetAcc, suggesting that they fail to detect the alignment of more specific concepts. Indeed, the resulting explanations (Appx. K) are associated with concepts (e.g., bird or animal) that are too general for the given task and fail to highlight relevant alignment exhibited by the probed model (e.g., species, colors).

\begin{table}[t]
    \caption{Average Alignment, Precision, and Relevance scores attributed by users to compositional explanations computed by the competitors. The superscript\textsuperscript{*} indicates that the results are computed on a different probing dataset.}
    \label{table:userstudy}
    \centering
    \begin{tabular}[b]{lccc}
        \toprule
         Scores & Align & Prec & Relev \\
         \midrule
          \multicolumn{4}{c}{Places365 Probed Model}  \\
           \midrule
        Human \cite{LaRosa2023Towards} & 3.53 & 2.98 & 3.13\\
        Closed \cite{bau2020units}& 3.10 & 2.60 & 3.25 \\
        Our & 3.53 & 3.19 & 3.34\\
         \midrule
          \multicolumn{4}{c}{CUB Probed Model}  \\
          \midrule
        Human \cite{LaRosa2023Towards} & ~3.17\textsuperscript{*}  & ~3.08\textsuperscript{*} & ~1.51\textsuperscript{*} \\
        Closed \cite{bau2020units}& 3.83 & 3.22 & 2.59\\
        Our & 3.32 & 3.27 & 4.30\\
         \bottomrule
    \end{tabular}   
\end{table}

To qualitatively validate our results and changes in \textbf{human interpretability}, we conducted a user study in which 100 participants were asked to rate, on a scale from 1 (none) to 5 (all), how many concepts in the compositional explanations generated by each method were aligned, precise, and relevant, as defined above.
The average scores reported in \cref{table:userstudy} (with std. dev. and p-values reported in Appx. G) suggest that our framework is the only one demonstrating consistency across both datasets, thus confirming its good properties. Indeed,  \cite{LaRosa2023Towards} performs poorly on the relevance score in the CUB dataset, as the compositional explanations are based on a probing dataset that includes concepts not relevant to the task. Conversely, the closed approach~\cite{bau2020units} achieves good scores on CUB, likely because its training data included the concept ``bird'', which is a label that is difficult for non-expert users to penalize (see Appx. G 
for a detailed analysis of the user study and Fig. 1 in the same appendix for an example of this problem). However, it fails to provide the appropriate level of granularity in ADE20K and to identify the alignment of relevant concepts in CUB, highlighting the lack of flexibility of closed vocabulary approaches.

\subsection{Explanations Analysis}
\label{sec:analysis}
\begin{figure}[t]
    \centering
    \includegraphics[scale=0.25]{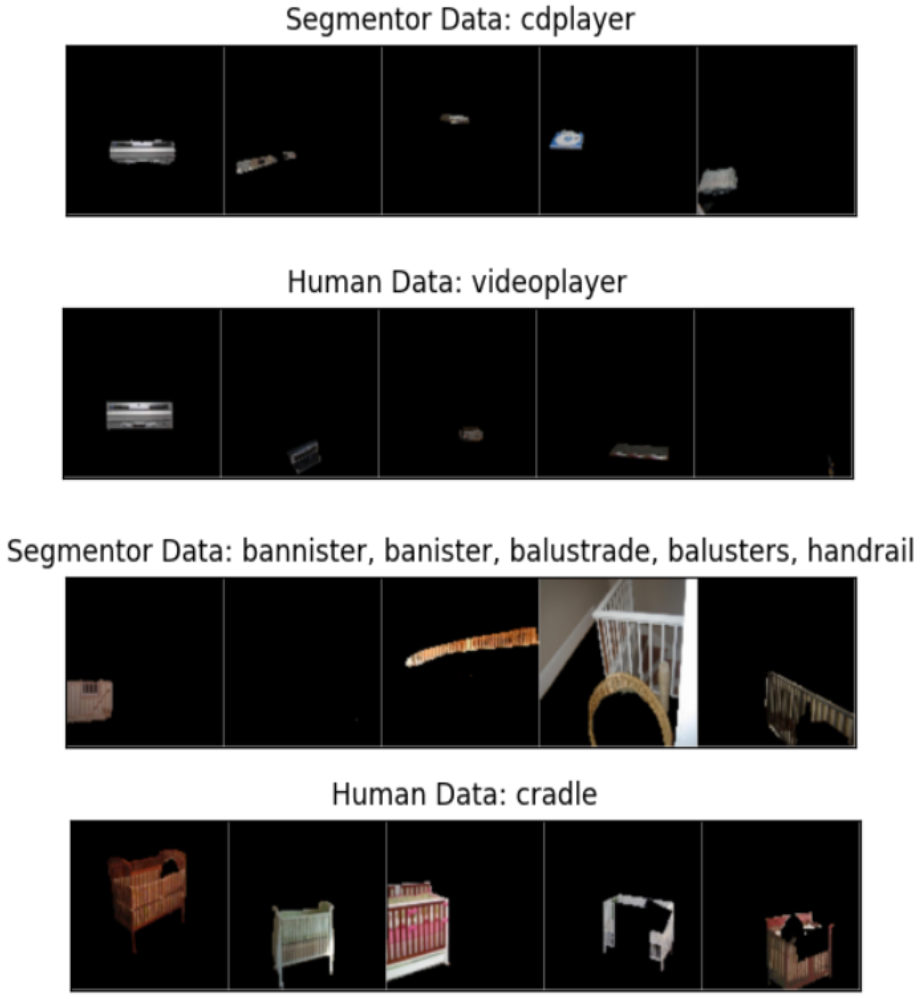}
    \caption{Examples of misalignment between human and model-annotated data due to different granularity in annotations (top) and the lack of concepts capturing patterns (bottom) in the concept set.}
    \label{fig:misalignment}
\end{figure}
\begin{figure}[t]
    \centering
        \begin{subfloat}{
\includegraphics[scale=0.34]{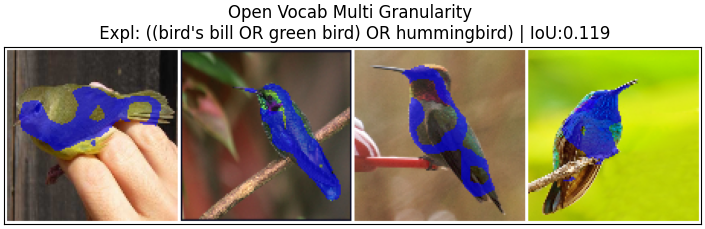}}
    \end{subfloat}
    \begin{subfloat}{
\includegraphics[scale=0.34]{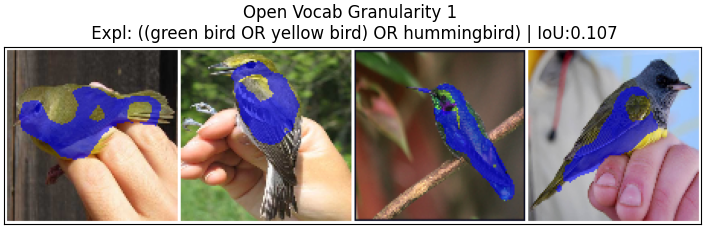}}
    \end{subfloat}
        \begin{subfloat}{
\includegraphics[scale=0.34]{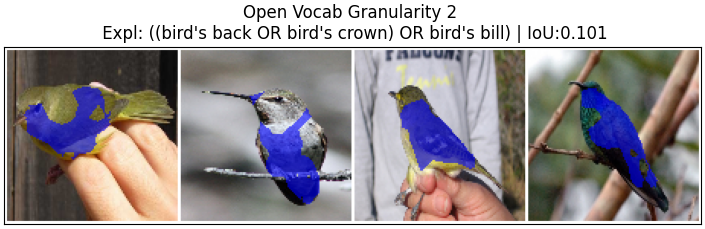}}
    \caption{Explanations associated with neuron \#19 and cluster 4 by our framework using different levels of granularity. In blue are areas of neuron activation within the considered range.}
    \label{fig:granularity}
    \end{subfloat}
\end{figure}

After validating the explanation quality of the proposed framework, this section analyzes the differences between explanations computed using open vocabulary and human-annotated data in a dataset where human annotations are available (Ade20k). 
The first question we aim to address is whether the differences in explanation scores arise from the segmentation masks (e.g., due to segmentation errors) while converging on the same explanation or whether the approaches converge on entirely different explanations. To explore this aspect, we measure the overlap in the explanation's concepts between the two approaches. We find that they share 86\%, 91\%, 82\%, 70\%, and 56\% of the labels across the five clusters, respectively. As discussed in the previous section, differences in the lower clusters are due to the algorithm converging differently on activations that do not align with any concept. More interesting, however, is the case of the highest activations, where almost half of the explanations differ. In this case, we observe that
the differences stem from \textbf{misalignment}. This phenomenon occurs when the two approaches converge on the same (or closely related) concept but assign different labels to it. In some cases, this misalignment can be attributed to hallucinations (e.g., vertical tanks often labeled as arcade machines). However, these cases are easy to identify by visually inspecting the samples that activate the explanations. More subtle and frequent cases of misalignment arise from differences in the concept set and the granularity of segmentations and annotations. For instance, as shown at the top of \cref{fig:misalignment}, a neuron associated with the concept ``cdplayer'' by the first approach is associated with ``videoplayer'' by explanations computed over model-annotated data. Although these two labels are closely related and likely represent the same underlying object (e.g., a generic ``media player''), the difference in annotation and segmentation granularity results in divergent explanations. Differently, at the bottom of the same figure, the two approaches converge on different samples and concepts. However, by visual inspection of these samples, they share highly similar patterns that are not available, as concepts, in the concept set (see \cref{sec:refinements}).

To measure the extent of these two kinds of misalignment, we leverage the semantic knowledge graph of WordNet~\cite{Miller1995} and then measure the extent of co-occurrence between misaligned concepts. Briefly, we map the concept set to nodes in WordNet and iteratively search for a \textit{meaningful} hypernym that generalizes the concepts causing the misalignment. We then remap the dataset's concept annotations to the identified hypernym and regenerate the segmentation maps, repeating the process until no other meaningful hypernym can be found (see Appx. E 
for further details). However, due to the incompleteness of the ontology, some misaligned concepts (e.g., cushion and pillow) cannot be unified through this approach.
Regarding co-occurrence, we categorize misaligned concepts into three groups: hyper-related concepts that co-occur in more than 75\% of the samples activating the explanation, highly related concepts with co-occurrence above 50\%, and concepts with low or no co-occurrence. 
Through this process, we observe that granularity impacts 12\% of the total concepts, with 4\% \textit{unifiable} through the ontology and 8\% hyper-related. The latter includes concepts whose annotations and segmentations are inconsistent or not aligned in granularity (e.g., traffic light vs road or mountain vs hill). Finally, 17\% are highly related and 19\% exhibit low or no co-occurrence. These represent cases where both approaches struggle due to the limitations of the concept set (similarly to \cref{fig:misalignment}). 
While this limitation could potentially be mitigated through refinements, some areas of misalignment (e.g., patterns) need further advancements in semantic segmentation to support concepts that are highly relevant for explainability but remain underexplored in standard semantic segmentation settings. In this direction, \textbf{we identify and discuss these limitations and potential research directions in Appx. C 
}.

\section{Application Scenarios}
\label{sec:applications}
In this section, we show how we can exploit the proposed framework to improve the explanations associated with neurons and improve our understanding of what they recognize.
\subsection{Supporting Custom Granularity}
\label{sec:granularity}
As described in \cref{sec:framework}, our framework supports multiple granularities through the use of concept subsets. These sets allow the algorithm to adjust explanations to the most aligned granularity. However, the framework can also be used to study individual neurons at different granularities, guided by the user. This capability is important because, due to superposition~\cite{elhage2022superposition,o2023disentangling,dreyer2024pure} and the fixed maximum length of explanations, some concepts aligned to the neuron may not be included in the explanation if they are weaker than those selected by the algorithm or do not add enough value to the previously selected concepts. \cref{fig:granularity} shows multi-granularity explanations and two single-granularity explanations for a neuron in the CUB model probed in \cref{sec:quantitative}.  
The first individual granularity represents bird-level concepts (i.e., shapes, colors, and species), while the second one represents birds' parts. Although the explanation that includes all of the granularities achieves the highest score, the analysis of individual granularities provides further insights into the neuron's recognition power. In this example, we can derive that the neuron recognizes species and colored birds as well as specific parts of these birds. \textbf{This analysis offers the user a more complete picture of the concepts learned by neurons}. Notably, this analysis cannot be supported by the \textit{Closed} approach~\cite{bau2020units} because it uses only one concept set and can be only partially supported (from lower to higher granularity) when combining ontologies and human-annotated data. Thus, this flexibility represents an additional advantage of our framework.

\subsection{Improving Explanations via Refinements}
\label{sec:refinements}
\begin{figure}
    \centering
    \includegraphics[scale=0.30]{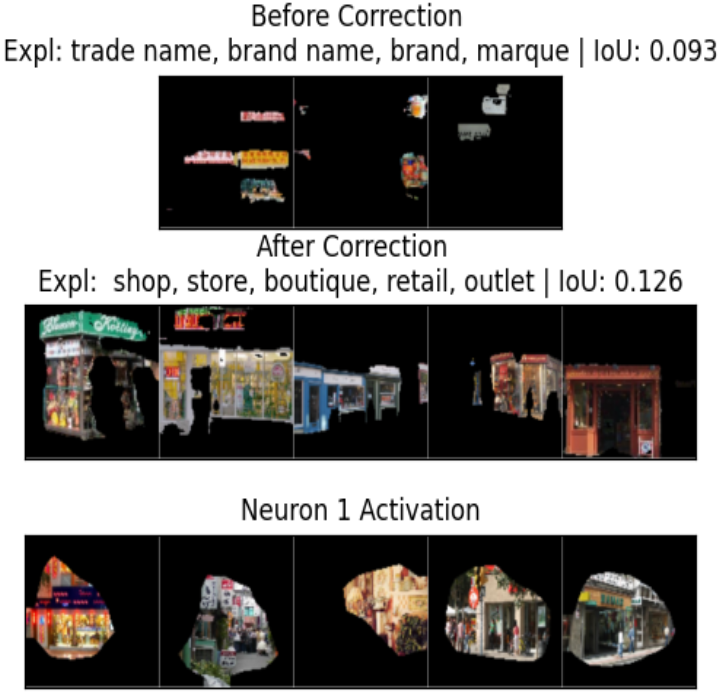}
    \caption{An example of how iterative refinements of the concept set can improve open vocabulary explanations. }
    \label{fig:correction}
\end{figure}
This section showcases how to improve misaligned explanations by correcting the concept set. In particular, the goal is to \textbf{analyze neurons' activations and explanations, identify possible misalignments due to the concept set, and fix them by refining the concept set}. Specifically, given an explanation of length $n$, we isolate the effect of a given concept into the explanation and we visually compare it with the neuron's activations not captured by the non-isolated part of the explanation (see Appx. F 
for the procedure). Here, we focus on the misaligned labels identified in \cref{sec:analysis}.
For example, as shown in \cref{fig:correction}, when examining neuron 1, we observed that this neuron appears to represent concepts such as ``shop'' or ``window shop''. However, the probing dataset (Ade20K) does not include labels for these concepts, causing both \cite{LaRosa2023Towards}'s approach and our method to converge on related concepts (e.g., trader name). To address this problem, we added the missing concepts to the concept set and re-generated the masks for our framework.
It is important to note that in this process, \textit{the user is not correcting the explanations but the concept set}. This means that when the user suggests a concept not aligned with the neuron's activation, 
 the segmentation model will still identify the new concept, but the compositional algorithm will discard it since it would be less aligned to the activation than the previous concepts. This ensures that the neuron explanation is faithful even if the concept set is modified. \Cref{fig:correction} shows that, after the refinement, the framework includes new concepts in the explanations and the updated explanations reach higher IoU scores than before. This means that the updated explanations are better aligned with the neuron activations or, equivalently, that the framework more accurately captures the alignment of the neuron activations. Finally, note that these improvements are not possible when using closed vocabulary segmentation models and require extensive and costly human labor to both annotate and fix the consistency of annotations in the human-based approaches.

\section{Conclusion}
\label{sec:conclusion}
In this paper, we introduced a novel framework to compute open vocabulary compositional explanations, addressing one of the main limitations of compositional explanations: their dependency on human-annotated datasets.
We demonstrated that our framework produces explanations that are comparable to or outperform previous approaches, while offering greater flexibility and broader applicability. We also call for further research in semantic segmentation to better support explainability tasks. Finally, future research directions could explore more advanced relationships between concepts, adapt the framework to different domains, and develop adaptive mechanisms to automatically identify the most suitable concept set for a given task.
\bibliographystyle{abbrvnat}
\bibliography{biblio}

\clearpage

\appendix
\label{appx}
\section{Extended Quantitative Evaluation}
\label{appx:quantitative}
This section complements the quantitative evaluation of our proposed frameworks by providing results computed using additional configurations. Results include both average and standard deviation. It is important to emphasize that the \textbf{explanations and the metrics used to measure their quality are not expected to exhibit low variance}. This variability arises from the overparameterization and the learning process of deep neural networks. Indeed, as observed by \cite{Bau2017} and \cite{Mu2020}, not all neurons within the network are aligned with specific concepts, leading to high variability in the degree of alignment. This effect is especially pronounced for compositional explanations and the metrics computed over pixel-level data, where the alignment between neuron activations and labeled concepts can fluctuate significantly. 

\subsection{Additional Open Vocabulary Segmentation Models}
\label{appx:eval_ovseg}
In this section, we compute explanations for the same settings considered in the main paper by using other segmentation models as the backbone of our proposed framework. \textbf{The goal of this experiment is not to select the best open vocabulary segmentation model} but to assess the general validity of the combination between compositional explanations and research in open vocabulary semantic segmentation. From the extensive range of models available in the literature~\cite{Xu_2023_CVPR2,xu2021,wysoczanska2024clipdino,Jiao2023,Cha_2023_CVPR,Liang_2023_CVPR,rao2021denseclip,Luo2023SegCLIP,Xu_2022_CVPR,shin2022reco,zhou2022maskclip,barsellotti2024training,zhou2022zegclip,xu2023learning,Ren2024,lueddecke22_cvpr,ghiasi2021open,Zou2024,Zou_2023_CVPR,OMGSeg,yu2023fcclip,APE,wang2023hierarchical,Xu_2023_CVPR}, we selected five representative models: CAT-Seg (L)~\cite{Cho2024CatSeg}, MasQCLIP~\cite{xu2023masqclip}, SCAN (VitL)~\cite{Liu_2024_CVPR}, SED (L)~\cite{Xie_2024_CVPR}, and OpenSeed (Swin-T)~\cite{Zhang2023openseed}. These models have been selected based on the following criteria: (i) they are among the most recent ones and published in major conferences, (ii) the pre-trained models are available to the general public, and (ii) the implementation is compatible with the technical settings considered in this paper (i.e., PyTorch 1.3~\cite{torch}, Detectron2~\cite{wu2019detectron2}, MMEngine 1.6.2~\cite{mmcv}, and MMSegmentation 0.27.0~\cite{mmseg2020}),  without requiring major code changes. While these models serve as examples of implementations of the framework, better explanations could potentially be obtained by using models beyond the settings tested in this paper, especially when using models trained on very large corpora~\cite{Zou2024,Li2024}. As weights, we use the pre-trained weights available in the official repositories of selected models.

In \Cref{tab:models_ade20k,tab:models_cub}, we report the results for the selected models when probing the Place365 and the CUB model, respectively. Note that, for all the open-vocabulary models proving the CUB model, we use the concept set identified for Cat-Seg (see \Cref{appx:conceptset}). This implies that the reported results could be further improved by refining the concept sets to better match the specific characteristics of each model. Moreover, the implementation based on \textit{OpenSeed} and the \textit{Closed} approach are both trained on the same dataset (i.e., COCO~\cite{Lin2014}) and share the same trained backbone (i.e., Swin-T~\cite{Liu_2021_ICCV}. This similarity results in similar scores for the highest activation range (Cluster 5), which is typically associated with the recognition of specific and complex objects~\cite{Mu2020,LaRosa2023Towards}. These results suggest a potential dependency of this implementation on the recognition capabilities of the shared backbone, where both models recognize the same mask, and thus they lean toward similar overlap with neuron activations, but they can possibly assign labels at different granularities. Overall, we observe comparable results across all models. No single model is able to outperform all the others in every setting, with each excelling in specific activation ranges and scores. These little differences can be attributed to the specific capabilities of recognizing more general or specific concepts of each segmentation model. Lastly, note that MasQCLIP is the only model that includes the ``background'' concept (by default). This difference explains the differences in the lower clusters of the Ade20K settings, typically influenced by default rules~\cite{LaRosa2023Towards} that include this kind of concept. Therefore, overall, the quality of the proposed framework does not strictly depend on the specific choice of its implementation.

\begin{table*}[!th]
          \caption{Avg. and Std. Dev. scores for explanations associated with a model trained on the Place365 dataset using Ade20K as a probing dataset.}
   
    \label{tab:models_ade20k}
    \centering
    \begin{tabular}{clccc}
        \toprule
        Cluster & Method& IoU & ActCov &DetAcc \\
         \midrule

         1 & Human \cite{LaRosa2023Towards} 
         &  0.219	 \footnotesize{ $\pm$ 	0.015	} &	0.352	 \footnotesize{ $\pm$ 	0.018	} &	0.369	 \footnotesize{ $\pm$ 	0.032	} \\
          & Closed \cite{bau2020units} &
          0.215	 \footnotesize{ $\pm$ 	0.015	} &	0.341	 \footnotesize{ $\pm$ 	0.018	} &	0.368	 \footnotesize{ $\pm$ 	0.032	} \\
                  & Ours\textsubscript{MasQCLIP} &
         0.112	 \footnotesize{ $\pm$ 	0.010	} &	0.137	 \footnotesize{ $\pm$ 	0.013	} &	0.373	 \footnotesize{ $\pm$ 	0.036	} \\
            & Ours\textsubscript{SCAN} &
         0.202	 \footnotesize{ $\pm$ 	0.014	} &	0.302	 \footnotesize{ $\pm$ 	0.014	} &	0.379	 \footnotesize{ $\pm$ 	0.034	} \\
        & Ours\textsubscript{SED} &
          0.206	 \footnotesize{ $\pm$ 	0.014	} &	0.313	 \footnotesize{ $\pm$ 	0.016	} &	0.377	 \footnotesize{ $\pm$ 	0.033	} \\
          & Ours\textsubscript{CAT-Seg} &
         0.212	 \footnotesize{ $\pm$ 	0.014	} &	0.327	 \footnotesize{ $\pm$ 	0.016	} &	0.376	 \footnotesize{ $\pm$ 	0.033	} \\
                 
        & Ours\textsubscript{OpenSeed} &
         0.226	 \footnotesize{ $\pm$ 	0.015	} &	0.372	 \footnotesize{ $\pm$ 	0.021	} &	0.367	 \footnotesize{ $\pm$ 	0.032	} \\
         \midrule
             2 & Human \cite{LaRosa2023Towards} &
          0.132	 \footnotesize{ $\pm$ 	0.021	} &	0.322	 \footnotesize{ $\pm$ 	0.040	} &	0.184	 \footnotesize{ $\pm$ 	0.033	} \\
                 & Closed \cite{bau2020units} & 
         0.130	 \footnotesize{ $\pm$ 	0.019	} &	0.306	 \footnotesize{ $\pm$ 	0.042	} &	0.187	 \footnotesize{ $\pm$ 	0.033	} \\
                 & Ours\textsubscript{MasQCLIP} &
         0.090	 \footnotesize{ $\pm$ 	0.014	} &	0.142	 \footnotesize{ $\pm$ 	0.026	} &	0.200	 \footnotesize{ $\pm$ 	0.033	} \\
            & Ours\textsubscript{SCAN} &
         0.125	 \footnotesize{ $\pm$ 	0.021	} &	0.272	 \footnotesize{ $\pm$ 	0.042	} &	0.190	 \footnotesize{ $\pm$ 	0.035	} \\
        & Ours\textsubscript{SED} &
         0.128	 \footnotesize{ $\pm$ 	0.020	} &	0.285	 \footnotesize{ $\pm$ 	0.040	} &	0.190	 \footnotesize{ $\pm$ 	0.034	} \\
& Ours\textsubscript{CAT-Seg} &
          0.130	 \footnotesize{ $\pm$ 	0.021	} &	0.302	 \footnotesize{ $\pm$ 	0.040	} &	0.188	 \footnotesize{ $\pm$ 	0.033	} \\
 & Ours\textsubscript{OpenSeed} &
0.136	 \footnotesize{ $\pm$ 	0.020	} &	0.340	 \footnotesize{ $\pm$ 	0.046	} &	0.186	 \footnotesize{ $\pm$ 	0.032	} \\
         \midrule
             3 & Human \cite{LaRosa2023Towards} &
          0.102	 \footnotesize{ $\pm$ 	0.031	} &	0.276	 \footnotesize{ $\pm$ 	0.086	} &	0.148	 \footnotesize{ $\pm$ 	0.048	} \\
         & Closed \cite{bau2020units} &
        0.106	 \footnotesize{ $\pm$ 	0.029	} &	0.272	 \footnotesize{ $\pm$ 	0.083	} &	0.155	 \footnotesize{ $\pm$ 	0.045	} \\
                 & Ours\textsubscript{MasQCLIP} &
          0.087	 \footnotesize{ $\pm$ 	0.023	} &	0.157	 \footnotesize{ $\pm$ 	0.049	} &	0.168	 \footnotesize{ $\pm$ 	0.044	} \\
            & Ours\textsubscript{SCAN} &
          0.104	 \footnotesize{ $\pm$ 	0.030	} &	0.244	 \footnotesize{ $\pm$ 	0.079	} &	0.161	 \footnotesize{ $\pm$ 	0.047	} \\
        & Ours\textsubscript{SED} &
          0.105	 \footnotesize{ $\pm$ 	0.030	} &	0.256	 \footnotesize{ $\pm$ 	0.077	} &	0.156	 \footnotesize{ $\pm$ 	0.046	} \\
& Ours\textsubscript{CAT-Seg} &
     0.105	 \footnotesize{ $\pm$ 	0.030	} &	0.266	 \footnotesize{ $\pm$ 	0.081	} &	0.155	 \footnotesize{ $\pm$ 	0.046	} \\
         & Ours\textsubscript{OpenSeed} &
        0.108	 \footnotesize{ $\pm$ 	0.029	} &	0.296	 \footnotesize{ $\pm$ 	0.089	} &	0.152	 \footnotesize{ $\pm$ 	0.044	} \\

         \midrule
             4 & Human \cite{LaRosa2023Towards} &
        0.083	 \footnotesize{ $\pm$ 	0.033	} &	0.226	 \footnotesize{ $\pm$ 	0.122	} &	0.139	 \footnotesize{ $\pm$ 	0.066	} \\
                  
                  & Closed \cite{bau2020units} &
         0.090	 \footnotesize{ $\pm$ 	0.033	} &	0.241	 \footnotesize{ $\pm$ 	0.121	} &	0.140	 \footnotesize{ $\pm$ 	0.056	} \\
              & Ours\textsubscript{MasQCLIP} &
          0.087	 \footnotesize{ $\pm$ 	0.032	} &	0.182	 \footnotesize{ $\pm$ 	0.078	} &	0.154	 \footnotesize{ $\pm$ 	0.055	} \\
            & Ours\textsubscript{SCAN} &
         0.093	 \footnotesize{ $\pm$ 	0.034	} &	0.222	 \footnotesize{ $\pm$ 	0.109	} &	0.154	 \footnotesize{ $\pm$ 	0.060	} \\
        & Ours\textsubscript{SED} &
          0.091	 \footnotesize{ $\pm$ 	0.033	} &	0.228	 \footnotesize{ $\pm$ 	0.114	} &	0.152	 \footnotesize{ $\pm$ 	0.064	} \\
        & Ours\textsubscript{CAT-Seg} &
          0.090	 \footnotesize{ $\pm$ 	0.034	} &	0.235	 \footnotesize{ $\pm$ 	0.118	} &	0.148	 \footnotesize{ $\pm$ 	0.065	} \\
          
          & Ours\textsubscript{OpenSeed} &
        0.088	 \footnotesize{ $\pm$ 	0.032	} &	0.256	 \footnotesize{ $\pm$ 	0.131	} &	0.137	 \footnotesize{ $\pm$ 	0.058	} \\

         \midrule
           5 & Human \cite{LaRosa2023Towards} &
               0.070	 \footnotesize{ $\pm$ 	0.044	} &	0.183	 \footnotesize{ $\pm$ 	0.134	} &	0.137	 \footnotesize{ $\pm$ 	0.094	} \\
                 
        & Closed \cite{bau2020units} &
          0.065	 \footnotesize{ $\pm$ 	0.034	} &	0.213	 \footnotesize{ $\pm$ 	0.140	} &	0.109	 \footnotesize{ $\pm$ 	0.070	} \\

        & Ours\textsubscript{MasQCLIP} &
          0.075	 \footnotesize{ $\pm$ 	0.036	} &	0.214	 \footnotesize{ $\pm$ 	0.126	} &	0.118	 \footnotesize{ $\pm$ 	0.059	} \\
            & Ours\textsubscript{SCAN} &
          0.082	 \footnotesize{ $\pm$ 	0.044	} &	0.220	 \footnotesize{ $\pm$ 	0.132	} &	0.139	 \footnotesize{ $\pm$ 	0.083	} \\
        & Ours\textsubscript{SED} &
          0.081	 \footnotesize{ $\pm$ 	0.044	} &	0.216	 \footnotesize{ $\pm$ 	0.134	} &	0.137	 \footnotesize{ $\pm$ 	0.078	} \\
        & Ours\textsubscript{CAT-Seg} &
         0.079	 \footnotesize{ $\pm$ 	0.044	} &	0.214	 \footnotesize{ $\pm$ 	0.141	} &	0.139	 \footnotesize{ $\pm$ 	0.085	} \\
        
        & Ours\textsubscript{OpenSeed} &
          0.064	 \footnotesize{ $\pm$ 	0.038	} &	0.215	 \footnotesize{ $\pm$ 	0.151	} &	0.110	 \footnotesize{ $\pm$ 	0.079	} \\
         \bottomrule
    \end{tabular}
\end{table*}

\begin{table*}[!th]
          \caption{Avg. and Std. Dev. scores for explanations associated with a model trained on the CUB dataset using CUB as a probing dataset.}
   
    \label{tab:models_cub}
    \centering
    \begin{tabular}{clccc}
        \toprule
        Cluster & Method& IoU & ActCov &DetAcc \\
         \midrule

         1 & Human \cite{LaRosa2023Towards} &
        0.248	 \footnotesize{ $\pm$ 	0.022	} &	0.356	 \footnotesize{ $\pm$ 	0.019	} &	0.451	 \footnotesize{ $\pm$ 	0.057	} \\
                 
        & Closed \cite{bau2020units} &
0.388	 \footnotesize{ $\pm$ 	0.040	} &	0.635	 \footnotesize{ $\pm$ 	0.019	} &	0.501	 \footnotesize{ $\pm$ 	0.061	} \\

        & Ours\textsubscript{MasQCLIP} &
       0.306	 \footnotesize{ $\pm$ 	0.028	} &	0.441	 \footnotesize{ $\pm$ 	0.022	} &	0.502	 \footnotesize{ $\pm$ 	0.063	} \\
            & Ours\textsubscript{SCAN} &
0.439	 \footnotesize{ $\pm$ 	0.045	} &	0.836	 \footnotesize{ $\pm$ 	0.020	} &	0.481	 \footnotesize{ $\pm$ 	0.055	} \\
        & Ours\textsubscript{SED}&
0.405	 \footnotesize{ $\pm$ 	0.040	} &	0.678	 \footnotesize{ $\pm$ 	0.016	} &	0.503	 \footnotesize{ $\pm$ 	0.059	} \\
        & Ours\textsubscript{CAT-Seg} &
0.357	 \footnotesize{ $\pm$ 	0.034	} &	0.553	 \footnotesize{ $\pm$ 	0.019	} &	0.504	 \footnotesize{ $\pm$ 	0.060	} \\
        & Ours\textsubscript{OpenSeed} &
        0.470	 \footnotesize{ $\pm$ 	0.051	} &	0.929	 \footnotesize{ $\pm$ 	0.030	} &	0.488	 \footnotesize{ $\pm$ 	0.059	} \\
         \midrule
             2 & Human \cite{LaRosa2023Towards} &
0.130	 \footnotesize{ $\pm$ 	0.035	} &	0.312	 \footnotesize{ $\pm$ 	0.059	} &	0.185	 \footnotesize{ $\pm$ 	0.057	} \\
                 
        & Closed \cite{bau2020units} &
0.170	 \footnotesize{ $\pm$ 	0.032	} &	0.505	 \footnotesize{ $\pm$ 	0.152	} &	0.214	 \footnotesize{ $\pm$ 	0.041	} \\

        & Ours\textsubscript{MasQCLIP} &
0.161	 \footnotesize{ $\pm$ 	0.024	} &	0.407	 \footnotesize{ $\pm$ 	0.076	} &	0.214	 \footnotesize{ $\pm$ 	0.035	} \\
            & Ours\textsubscript{SCAN} &
0.174	 \footnotesize{ $\pm$ 	0.034	} &	0.563	 \footnotesize{ $\pm$ 	0.188	} &	0.209	 \footnotesize{ $\pm$ 	0.038	} \\
        & Ours\textsubscript{SED}&
0.176	 \footnotesize{ $\pm$ 	0.032	} &	0.522	 \footnotesize{ $\pm$ 	0.138	} &	0.215	 \footnotesize{ $\pm$ 	0.036	} \\
        & Ours\textsubscript{CAT-Seg} &
0.173	 \footnotesize{ $\pm$ 	0.028	} &	0.463	 \footnotesize{ $\pm$ 	0.102	} &	0.221	 \footnotesize{ $\pm$ 	0.038	} \\
        & Ours\textsubscript{OpenSeed} &
         0.179	 \footnotesize{ $\pm$ 	0.033	} &	0.602	 \footnotesize{ $\pm$ 	0.198	} &	0.211	 \footnotesize{ $\pm$ 	0.036	} \\
         \midrule
             3 & Human \cite{LaRosa2023Towards} &
        0.085	 \footnotesize{ $\pm$ 	0.031	} &	0.228	 \footnotesize{ $\pm$ 	0.088	} &	0.126	 \footnotesize{ $\pm$ 	0.046	} \\
                 
        & Closed \cite{bau2020units} &
0.142	 \footnotesize{ $\pm$ 	0.030	} &	0.453	 \footnotesize{ $\pm$ 	0.116	} &	0.175	 \footnotesize{ $\pm$ 	0.039	} \\

        & Ours\textsubscript{MasQCLIP} &
0.136	 \footnotesize{ $\pm$ 	0.027	} &	0.388	 \footnotesize{ $\pm$ 	0.074	} &	0.176	 \footnotesize{ $\pm$ 	0.038	} \\
            & Ours\textsubscript{SCAN} &
0.144	 \footnotesize{ $\pm$ 	0.029	} &	0.422	 \footnotesize{ $\pm$ 	0.101	} &	0.182	 \footnotesize{ $\pm$ 	0.039	} \\
        & Ours\textsubscript{SED}&
0.143	 \footnotesize{ $\pm$ 	0.028	} &	0.425	 \footnotesize{ $\pm$ 	0.089	} &	0.180	 \footnotesize{ $\pm$ 	0.036	} \\
        & Ours\textsubscript{CAT-Seg} &
0.147	 \footnotesize{ $\pm$ 	0.030	} &	0.432	 \footnotesize{ $\pm$ 	0.093	} &	0.185	 \footnotesize{ $\pm$ 	0.038	} \\
        & Ours\textsubscript{OpenSeed} &
         0.141	 \footnotesize{ $\pm$ 	0.027	} &	0.463	 \footnotesize{ $\pm$ 	0.101	} &	0.170	 \footnotesize{ $\pm$ 	0.034	} \\

         \midrule
             4 & Human \cite{LaRosa2023Towards} &
0.063	 \footnotesize{ $\pm$ 	0.030	} &	0.167	 \footnotesize{ $\pm$ 	0.101	} &	0.105	 \footnotesize{ $\pm$ 	0.050	} \\
                 
        & Closed \cite{bau2020units} &
0.091	 \footnotesize{ $\pm$ 	0.027	} &	0.571	 \footnotesize{ $\pm$ 	0.136	} &	0.100	 \footnotesize{ $\pm$ 	0.031	} \\

        & Ours\textsubscript{MasQCLIP} &
0.098	 \footnotesize{ $\pm$ 	0.024	} &	0.336	 \footnotesize{ $\pm$ 	0.105	} &	0.126	 \footnotesize{ $\pm$ 	0.035	} \\
            & Ours\textsubscript{SCAN} &
0.101	 \footnotesize{ $\pm$ 	0.026	} &	0.426	 \footnotesize{ $\pm$ 	0.139	} &	0.123	 \footnotesize{ $\pm$ 	0.037	} \\
        & Ours\textsubscript{SED}&
0.103	 \footnotesize{ $\pm$ 	0.025	} &	0.383	 \footnotesize{ $\pm$ 	0.122	} &	0.129	 \footnotesize{ $\pm$ 	0.038	} \\
        & Ours\textsubscript{CAT-Seg} &
0.113	 \footnotesize{ $\pm$ 	0.027	} &	0.356	 \footnotesize{ $\pm$ 	0.115	} &	0.147	 \footnotesize{ $\pm$ 	0.039	} \\
        & Ours\textsubscript{OpenSeed} &
       0.095	 \footnotesize{ $\pm$ 	0.025	} &	0.413	 \footnotesize{ $\pm$ 	0.089	} &	0.111	 \footnotesize{ $\pm$ 	0.031	} \\

         \midrule
           5 & Human \cite{LaRosa2023Towards} &
0.052	 \footnotesize{ $\pm$ 	0.029	} &	0.144	 \footnotesize{ $\pm$ 	0.124	} &	0.100	 \footnotesize{ $\pm$ 	0.058	} \\
                 
        & Closed \cite{bau2020units} &
0.029	 \footnotesize{ $\pm$ 	0.014	} &	0.674	 \footnotesize{ $\pm$ 	0.195	} &	0.033	 \footnotesize{ $\pm$ 	0.028	} \\

        & Ours\textsubscript{MasQCLIP} &
0.059	 \footnotesize{ $\pm$ 	0.019	} &	0.165	 \footnotesize{ $\pm$ 	0.067	} &	0.095	 \footnotesize{ $\pm$ 	0.044	} \\
            & Ours\textsubscript{SCAN} &
0.060	 \footnotesize{ $\pm$ 	0.021	} &	0.153	 \footnotesize{ $\pm$ 	0.080	} &	0.112	 \footnotesize{ $\pm$ 	0.059	} \\
        & Ours\textsubscript{SED}&
0.068	 \footnotesize{ $\pm$ 	0.023	} &	0.155	 \footnotesize{ $\pm$ 	0.069	} &	0.125	 \footnotesize{ $\pm$ 	0.055	} \\
        & Ours\textsubscript{CAT-Seg} &
0.077	 \footnotesize{ $\pm$ 	0.024	} &	0.188	 \footnotesize{ $\pm$ 	0.072	} &	0.131	 \footnotesize{ $\pm$ 	0.056	} \\
        & Ours\textsubscript{OpenSeed} &
        0.042	 \footnotesize{ $\pm$ 	0.016	} &	0.170	 \footnotesize{ $\pm$ 	0.103	} &	0.060	 \footnotesize{ $\pm$ 	0.039	} \\
         \bottomrule
    \end{tabular}
\end{table*}
\subsection{Additional Probing Datasets}
\label{appx:eval_dataset}
In this section, we report the results obtained by using several datasets as probing datasets for computing compositional explanations. We report the results for all the implementations (\Cref{appx:eval_ovseg}) of our proposed framework other than the human-based and closed vocabulary approaches. These datasets have been chosen because there are publicly available scripts to make them compatible with Detectron2~\cite{wu2019detectron2}, which is the most common framework used for evaluating open vocabulary segmentation models, and they are commonly used to evaluate progress in the image segmentation field or compositional explanations (Ade20k and PASCAL). Specifically, we randomly extract 50 neurons for each probed model and we generate explanations for those neurons using as a probing dataset the validation split of the following datasets: Mapillary Vistas~\cite{Neuhold2017}, Cityscapes~\cite{Cordts2016Cityscapes},  Pascal VOC~\cite{pascal-voc-2012}, PASCAL-Context-459~\cite{mottaghi_cvpr14}, Ade20k in its extended version with 847 classes~\cite{Zhou2017}, and COCO-Stuff~\cite{Caesar_2018_CVPR}. Note that we do not include OpenSeed in the Mapillary Vistas evaluation due to technical limitations\footnote{Out of Memory issues on a GTX 3090 graphic card.}. As a probed model, we use the same model used in Section 4 trained on Place365, since the learned place categories are related to the concepts and segmentation masks included in these datasets\footnote{Note that we do not probe models trained on these datasets, as they are segmentation models specifically trained to classify the same concepts. This undermines the utility of compositional explanations.}. We follow the same settings used in Section 4. Therefore, we use the masks' labels from the dataset as the concept set for our framework without further refining the concept set and without splitting it into concept subsets.

\begin{table*}[!t]
          \caption{Avg. scores for explanations associated with a model trained on the Place365 dataset using Ade20K-Extended (847 classes) as a probing dataset. }
   
    \label{tab:ade20kfull}
    \centering
    \begin{tabular}{clccc}
        \toprule
        Cluster & Method& IoU & ActCov &DetAcc \\
         \midrule

         1 & Human \cite{LaRosa2023Towards} &
          0.218	 \footnotesize{ $\pm$ 	0.016	} &	0.350	 \footnotesize{ $\pm$ 	0.016	} &	0.367	 \footnotesize{ $\pm$ 	0.034	} \\
          & Closed \cite{bau2020units} &
          0.215	 \footnotesize{ $\pm$ 	0.015	} &	0.340	 \footnotesize{ $\pm$ 	0.019	} &	0.369	 \footnotesize{ $\pm$ 	0.033	} \\
        & Ours\textsubscript{MasQCLIP} &
       0.087	 \footnotesize{ $\pm$ 	0.007	} &	0.100	 \footnotesize{ $\pm$ 	0.008	} &	0.393	 \footnotesize{ $\pm$ 	0.033	} \\
            & Ours\textsubscript{SCAN} &
             0.185	 \footnotesize{ $\pm$ 	0.014	} &	0.263	 \footnotesize{ $\pm$ 	0.014	} &	0.383	 \footnotesize{ $\pm$ 	0.038	} \\
        & Ours\textsubscript{SED} &
       0.198	 \footnotesize{ $\pm$ 	0.014	} &	0.293	 \footnotesize{ $\pm$ 	0.013	} &	0.379	 \footnotesize{ $\pm$ 	0.035	} \\
          & Ours\textsubscript{CAT-Seg} &
         0.203	 \footnotesize{ $\pm$ 	0.014	} &	0.304	 \footnotesize{ $\pm$ 	0.013	} &	0.378	 \footnotesize{ $\pm$ 	0.034	} \\
                 
        & Ours\textsubscript{OpenSeed} &
         0.223	 \footnotesize{ $\pm$ 	0.015	} &	0.363	 \footnotesize{ $\pm$ 	0.020	} &	0.368	 \footnotesize{ $\pm$ 	0.032	} \\
         \midrule
             2 & Human \cite{LaRosa2023Towards} &
         0.131	 \footnotesize{ $\pm$ 	0.020	} &	0.320	 \footnotesize{ $\pm$ 	0.039	} &	0.184	 \footnotesize{ $\pm$ 	0.035	} \\
                 & Closed \cite{bau2020units} & 
        0.130	 \footnotesize{ $\pm$ 	0.019	} &	0.308	 \footnotesize{ $\pm$ 	0.039	} &	0.186	 \footnotesize{ $\pm$ 	0.035	} \\
                 & Ours\textsubscript{MasQCLIP} &
        0.076	 \footnotesize{ $\pm$ 	0.015	} &	0.107	 \footnotesize{ $\pm$ 	0.021	} &	0.210	 \footnotesize{ $\pm$ 	0.043	} \\
            & Ours\textsubscript{SCAN} &
             0.114	 \footnotesize{ $\pm$ 	0.020	} &	0.222	 \footnotesize{ $\pm$ 	0.038	} &	0.193	 \footnotesize{ $\pm$ 	0.042	} \\
        & Ours\textsubscript{SED} &
        0.122	 \footnotesize{ $\pm$ 	0.020	} &	0.258	 \footnotesize{ $\pm$ 	0.035	} &	0.191	 \footnotesize{ $\pm$ 	0.039	} \\
& Ours\textsubscript{CAT-Seg} &
         0.124	 \footnotesize{ $\pm$ 	0.020	} &	0.272	 \footnotesize{ $\pm$ 	0.034	} &	0.188	 \footnotesize{ $\pm$ 	0.037	} \\
 & Ours\textsubscript{OpenSeed} &
         0.133	 \footnotesize{ $\pm$ 	0.019	} &	0.333	 \footnotesize{ $\pm$ 	0.040	} &	0.184	 \footnotesize{ $\pm$ 	0.034	} \\
         \midrule
             3 & Human \cite{LaRosa2023Towards} &
         0.101	 \footnotesize{ $\pm$ 	0.029	} &	0.263	 \footnotesize{ $\pm$ 	0.083	} &	0.149	 \footnotesize{ $\pm$ 	0.046	} \\
         & Closed \cite{bau2020units} &
         0.104	 \footnotesize{ $\pm$ 	0.028	} &	0.262	 \footnotesize{ $\pm$ 	0.079	} &	0.156	 \footnotesize{ $\pm$ 	0.047	} \\
        & Ours\textsubscript{MasQCLIP} &
       0.075	 \footnotesize{ $\pm$ 	0.023	} &	0.118	 \footnotesize{ $\pm$ 	0.042	} &	0.176	 \footnotesize{ $\pm$ 	0.049	} \\
            & Ours\textsubscript{SCAN} &
             0.096	 \footnotesize{ $\pm$ 	0.027	} &	0.203	 \footnotesize{ $\pm$ 	0.062	} &	0.163	 \footnotesize{ $\pm$ 	0.047	} \\
        & Ours\textsubscript{SED} &
        0.100	 \footnotesize{ $\pm$ 	0.029	} &	0.230	 \footnotesize{ $\pm$ 	0.066	} &	0.155	 \footnotesize{ $\pm$ 	0.045	} \\
& Ours\textsubscript{CAT-Seg} &
       0.100	 \footnotesize{ $\pm$ 	0.029	} &	0.240	 \footnotesize{ $\pm$ 	0.073	} &	0.155	 \footnotesize{ $\pm$ 	0.047	} \\
         & Ours\textsubscript{OpenSeed} &
         0.104	 \footnotesize{ $\pm$ 	0.028	} &	0.280	 \footnotesize{ $\pm$ 	0.081	} &	0.151	 \footnotesize{ $\pm$ 	0.046	} \\

         \midrule
             4 & Human \cite{LaRosa2023Towards} &
         0.083	 \footnotesize{ $\pm$ 	0.030	} &	0.219	 \footnotesize{ $\pm$ 	0.107	} &	0.142	 \footnotesize{ $\pm$ 	0.073	} \\
                  
                  & Closed \cite{bau2020units} &
         0.089	 \footnotesize{ $\pm$ 	0.030	} &	0.230	 \footnotesize{ $\pm$ 	0.098	} &	0.141	 \footnotesize{ $\pm$ 	0.058	} \\
         & Ours\textsubscript{MasQCLIP} &
       0.079	 \footnotesize{ $\pm$ 	0.027	} &	0.141	 \footnotesize{ $\pm$ 	0.055	} &	0.166	 \footnotesize{ $\pm$ 	0.057	} \\
            & Ours\textsubscript{SCAN} &
             0.086	 \footnotesize{ $\pm$ 	0.029	} &	0.192	 \footnotesize{ $\pm$ 	0.083	} &	0.155	 \footnotesize{ $\pm$ 	0.072	} \\
        & Ours\textsubscript{SED} &
        0.086	 \footnotesize{ $\pm$ 	0.031	} &	0.209	 \footnotesize{ $\pm$ 	0.085	} &	0.146	 \footnotesize{ $\pm$ 	0.071	} \\     
        & Ours\textsubscript{CAT-Seg} &
         0.086	 \footnotesize{ $\pm$ 	0.031	} &	0.211	 \footnotesize{ $\pm$ 	0.091	} &	0.148	 \footnotesize{ $\pm$ 	0.074	} \\
          
          & Ours\textsubscript{OpenSeed} &
       0.088	 \footnotesize{ $\pm$ 	0.030	} &	0.232	 \footnotesize{ $\pm$ 	0.110	} &	0.145	 \footnotesize{ $\pm$ 	0.067	} \\

         \midrule
           5 & Human \cite{LaRosa2023Towards} &
               0.098	 \footnotesize{ $\pm$ 	0.076	} &	0.196	 \footnotesize{ $\pm$ 	0.148	} &	0.242	 \footnotesize{ $\pm$ 	0.171	} \\
                 
        & Closed \cite{bau2020units} &
          0.071	 \footnotesize{ $\pm$ 	0.042	} &	0.221	 \footnotesize{ $\pm$ 	0.145	} &	0.119	 \footnotesize{ $\pm$ 	0.081	} \\
          & Ours\textsubscript{MasQCLIP} &
        0.103	 \footnotesize{ $\pm$ 	0.056	} &	0.186	 \footnotesize{ $\pm$ 	0.100	} &	0.207	 \footnotesize{ $\pm$ 	0.092	} \\
            & Ours\textsubscript{SCAN} &
            0.107	 \footnotesize{ $\pm$ 	0.072	} &	0.195	 \footnotesize{ $\pm$ 	0.127	} &	0.240	 \footnotesize{ $\pm$ 	0.140	} \\
        & Ours\textsubscript{SED} &
       0.106	 \footnotesize{ $\pm$ 	0.071	} &	0.195	 \footnotesize{ $\pm$ 	0.136	} &	0.234	 \footnotesize{ $\pm$ 	0.135	} \\     
        & Ours\textsubscript{CAT-Seg} &
        0.103	 \footnotesize{ $\pm$ 	0.070	} &	0.208	 \footnotesize{ $\pm$ 	0.145	} &	0.236	 \footnotesize{ $\pm$ 	0.139	} \\	
        
        & Ours\textsubscript{OpenSeed} &
         0.079	 \footnotesize{ $\pm$ 	0.065	} &	0.226	 \footnotesize{ $\pm$ 	0.169	} &	0.152	 \footnotesize{ $\pm$ 	0.119	} \\	

         \bottomrule
    \end{tabular}
\end{table*}

\begin{table*}[!t]
          \caption{Avg. and Std. Dev. scores for explanations associated with a model trained on the Place365 dataset using Mapillary Vistas as a probing dataset.}
   
    \label{tab:mapillary}
    \centering

    \begin{tabular}{clccc}
        \toprule
        Cluster & Method& IoU & ActCov &DetAcc \\
         \midrule

         1 & Human \cite{LaRosa2023Towards} &
        0.304	 \footnotesize{ $\pm$ 	0.042	} &	0.652	 \footnotesize{ $\pm$ 	0.040	} &	0.363	 \footnotesize{ $\pm$ 	0.052	} \\
          & Closed \cite{bau2020units} &
          0.310	 \footnotesize{ $\pm$ 	0.043	} &	0.680	 \footnotesize{ $\pm$ 	0.045	} &	0.363	 \footnotesize{ $\pm$ 	0.052	} \\
                  & Ours\textsubscript{MasQCLIP} &
   0.202	 \footnotesize{ $\pm$ 	0.021	} &	0.308	 \footnotesize{ $\pm$ 	0.019	} &	0.374	 \footnotesize{ $\pm$ 	0.056	} \\
            & Ours\textsubscript{SCAN} &
            0.311	 \footnotesize{ $\pm$ 	0.042	} &	0.691	 \footnotesize{ $\pm$ 	0.038	} &	0.362	 \footnotesize{ $\pm$ 	0.051	} \\
        & Ours\textsubscript{SED} &
          0.313	 \footnotesize{ $\pm$ 	0.042	} &	0.714	 \footnotesize{ $\pm$ 	0.051	} &	0.358	 \footnotesize{ $\pm$ 	0.051	} \\
          & Ours\textsubscript{CAT-Seg} &
         0.313	 \footnotesize{ $\pm$ 	0.042	} &	0.704	 \footnotesize{ $\pm$ 	0.038	} &	0.361	 \footnotesize{ $\pm$ 	0.052	} \\

         \midrule
             2 & Human \cite{LaRosa2023Towards} &
         0.161	 \footnotesize{ $\pm$ 	0.035	} &	0.517	 \footnotesize{ $\pm$ 	0.106	} &	0.192	 \footnotesize{ $\pm$ 	0.043	} \\
                 & Closed \cite{bau2020units} & 
         0.166	 \footnotesize{ $\pm$ 	0.037	} &	0.530	 \footnotesize{ $\pm$ 	0.116	} &	0.197	 \footnotesize{ $\pm$ 	0.045	} \\
                           & Ours\textsubscript{MasQCLIP} &
 0.138	 \footnotesize{ $\pm$ 	0.028	} &	0.301	 \footnotesize{ $\pm$ 	0.041	} &	0.205	 \footnotesize{ $\pm$ 	0.050	} \\
            & Ours\textsubscript{SCAN} &
            0.168	 \footnotesize{ $\pm$ 	0.037	} &	0.552	 \footnotesize{ $\pm$ 	0.118	} &	0.196	 \footnotesize{ $\pm$ 	0.043	} \\
        & Ours\textsubscript{SED} &
           0.168	 \footnotesize{ $\pm$ 	0.039	} &	0.569	 \footnotesize{ $\pm$ 	0.119	} &	0.195	 \footnotesize{ $\pm$ 	0.046	} \\
& Ours\textsubscript{CAT-Seg} &
         0.169	 \footnotesize{ $\pm$ 	0.038	} &	0.570	 \footnotesize{ $\pm$ 	0.121	} &	0.195	 \footnotesize{ $\pm$ 	0.044	} \\

         \midrule
             3 & Human \cite{LaRosa2023Towards} &
         0.121	 \footnotesize{ $\pm$ 	0.039	} &	0.409	 \footnotesize{ $\pm$ 	0.109	} &	0.152	 \footnotesize{ $\pm$ 	0.055	} \\
         & Closed \cite{bau2020units} &
        0.124	 \footnotesize{ $\pm$ 	0.042	} &	0.436	 \footnotesize{ $\pm$ 	0.126	} &	0.151	 \footnotesize{ $\pm$ 	0.055	} \\
                           & Ours\textsubscript{MasQCLIP} &
 0.110	 \footnotesize{ $\pm$ 	0.035	} &	0.276	 \footnotesize{ $\pm$ 	0.065	} &	0.159	 \footnotesize{ $\pm$ 	0.055	} \\
            & Ours\textsubscript{SCAN} &
            0.126	 \footnotesize{ $\pm$ 	0.043	} &	0.460	 \footnotesize{ $\pm$ 	0.116	} &	0.151	 \footnotesize{ $\pm$ 	0.054	} \\
        & Ours\textsubscript{SED} &
           0.125	 \footnotesize{ $\pm$ 	0.043	} &	0.476	 \footnotesize{ $\pm$ 	0.130	} &	0.149	 \footnotesize{ $\pm$ 	0.057	} \\
& Ours\textsubscript{CAT-Seg} &
       0.126	 \footnotesize{ $\pm$ 	0.043	} &	0.482	 \footnotesize{ $\pm$ 	0.125	} &	0.149	 \footnotesize{ $\pm$ 	0.056	} \\

         \midrule
             4 & Human \cite{LaRosa2023Towards} &
        0.088	 \footnotesize{ $\pm$ 	0.042	} &	0.323	 \footnotesize{ $\pm$ 	0.139	} &	0.123	 \footnotesize{ $\pm$ 	0.076	} \\
                  
                  & Closed \cite{bau2020units} &
         0.087	 \footnotesize{ $\pm$ 	0.040	} &	0.334	 \footnotesize{ $\pm$ 	0.150	} &	0.117	 \footnotesize{ $\pm$ 	0.065	} \\
                          & Ours\textsubscript{MasQCLIP} &
   0.083	 \footnotesize{ $\pm$ 	0.037	} &	0.241	 \footnotesize{ $\pm$ 	0.100	} &	0.125	 \footnotesize{ $\pm$ 	0.081	} \\
            & Ours\textsubscript{SCAN} &
            0.087	 \footnotesize{ $\pm$ 	0.041	} &	0.370	 \footnotesize{ $\pm$ 	0.155	} &	0.112	 \footnotesize{ $\pm$ 	0.057	} \\
        & Ours\textsubscript{SED} &
           0.086	 \footnotesize{ $\pm$ 	0.041	} &	0.371	 \footnotesize{ $\pm$ 	0.165	} &	0.116	 \footnotesize{ $\pm$ 	0.072	} \\
        & Ours\textsubscript{CAT-Seg} &
         0.087	 \footnotesize{ $\pm$ 	0.042	} &	0.372	 \footnotesize{ $\pm$ 	0.163	} &	0.116	 \footnotesize{ $\pm$ 	0.076	} \\

         \midrule
           5 & Human \cite{LaRosa2023Towards} &
                0.053	 \footnotesize{ $\pm$ 	0.036	} &	0.266	 \footnotesize{ $\pm$ 	0.200	} &	0.080	 \footnotesize{ $\pm$ 	0.068	} \\
                 
        & Closed \cite{bau2020units} &
        0.050	 \footnotesize{ $\pm$ 	0.028	} &	0.254	 \footnotesize{ $\pm$ 	0.207	} &	0.082	 \footnotesize{ $\pm$ 	0.068	} \\
                          & Ours\textsubscript{MasQCLIP} &
   0.056	 \footnotesize{ $\pm$ 	0.038	} &	0.187	 \footnotesize{ $\pm$ 	0.114	} &	0.082	 \footnotesize{ $\pm$ 	0.072	} \\
            & Ours\textsubscript{SCAN} &
            0.052	 \footnotesize{ $\pm$ 	0.033	} &	0.264	 \footnotesize{ $\pm$ 	0.217	} &	0.081	 \footnotesize{ $\pm$ 	0.057	} \\
        & Ours\textsubscript{SED} &
          0.052	 \footnotesize{ $\pm$ 	0.035	} &	0.273	 \footnotesize{ $\pm$ 	0.223	} &	0.077	 \footnotesize{ $\pm$ 	0.057	} \\
        & Ours\textsubscript{CAT-Seg} &
          0.052	 \footnotesize{ $\pm$ 	0.033	} &	0.271	 \footnotesize{ $\pm$ 	0.226	} &	0.078	 \footnotesize{ $\pm$ 	0.051	} \\
        
         \bottomrule
    \end{tabular}

\end{table*}

\begin{table*}[!t]
          \caption{Avg. and Std. Dev. scores for explanations associated with a model trained on the Place365 dataset using Citiscapes as a probing dataset. }
   
    \label{tab:cityscapes}
    \centering

    \begin{tabular}{clccc}
        \toprule
        Cluster & Method& IoU & ActCov &DetAcc \\
         \midrule

         1 & Human \cite{LaRosa2023Towards} &
          0.294	 \footnotesize{ $\pm$ 	0.038	} &	0.650	 \footnotesize{ $\pm$ 	0.055	} &	0.353	 \footnotesize{ $\pm$ 	0.055	} \\
          & Closed \cite{bau2020units} &
          0.309	 \footnotesize{ $\pm$ 	0.042	} &	0.687	 \footnotesize{ $\pm$ 	0.061	} &	0.363	 \footnotesize{ $\pm$ 	0.059	} \\
        & Ours\textsubscript{MasQCLIP} &
        0.306	 \footnotesize{ $\pm$ 	0.045	} &	0.701	 \footnotesize{ $\pm$ 	0.048	} &	0.354	 \footnotesize{ $\pm$ 	0.055	} \\
            & Ours\textsubscript{SCAN} &
            0.316	 \footnotesize{ $\pm$ 	0.044	} &	0.753	 \footnotesize{ $\pm$ 	0.067	} &	0.355	 \footnotesize{ $\pm$ 	0.055	} \\
        & Ours\textsubscript{SED} &
       0.310	 \footnotesize{ $\pm$ 	0.042	} &	0.724	 \footnotesize{ $\pm$ 	0.068	} &	0.355	 \footnotesize{ $\pm$ 	0.057	} \\
          & Ours\textsubscript{CAT-Seg} &
        0.314	 \footnotesize{ $\pm$ 	0.043	} &	0.729	 \footnotesize{ $\pm$ 	0.066	} &	0.359	 \footnotesize{ $\pm$ 	0.057	} \\
                 
        & Ours\textsubscript{OpenSeed} &
        0.309	 \footnotesize{ $\pm$ 	0.041	} &	0.711	 \footnotesize{ $\pm$ 	0.069	} &	0.357	 \footnotesize{ $\pm$ 	0.058	} \\
         \midrule
             2 & Human \cite{LaRosa2023Towards} &
          0.178	 \footnotesize{ $\pm$ 	0.044	} &	0.580	 \footnotesize{ $\pm$ 	0.097	} &	0.206	 \footnotesize{ $\pm$ 	0.052	} \\
                 & Closed \cite{bau2020units} & 
        0.183	 \footnotesize{ $\pm$ 	0.046	} &	0.620	 \footnotesize{ $\pm$ 	0.100	} &	0.208	 \footnotesize{ $\pm$ 	0.053	} \\
                 & Ours\textsubscript{MasQCLIP} &
        0.177	 \footnotesize{ $\pm$ 	0.048	} &	0.639	 \footnotesize{ $\pm$ 	0.097	} &	0.197	 \footnotesize{ $\pm$ 	0.054	} \\
            & Ours\textsubscript{SCAN} &
             0.186	 \footnotesize{ $\pm$ 	0.046	} &	0.655	 \footnotesize{ $\pm$ 	0.117	} &	0.207	 \footnotesize{ $\pm$ 	0.051	} \\
        & Ours\textsubscript{SED} &
       0.184	 \footnotesize{ $\pm$ 	0.048	} &	0.655	 \footnotesize{ $\pm$ 	0.106	} &	0.205	 \footnotesize{ $\pm$ 	0.054	} \\
& Ours\textsubscript{CAT-Seg} &
          0.185	 \footnotesize{ $\pm$ 	0.047	} &	0.649	 \footnotesize{ $\pm$ 	0.115	} &	0.207	 \footnotesize{ $\pm$ 	0.053	} \\
 & Ours\textsubscript{OpenSeed} &
         0.183	 \footnotesize{ $\pm$ 	0.047	} &	0.650	 \footnotesize{ $\pm$ 	0.103	} &	0.204	 \footnotesize{ $\pm$ 	0.053	} \\
         \midrule
             3 & Human \cite{LaRosa2023Towards} &
          0.131	 \footnotesize{ $\pm$ 	0.045	} &	0.500	 \footnotesize{ $\pm$ 	0.099	} &	0.154	 \footnotesize{ $\pm$ 	0.056	} \\
         & Closed \cite{bau2020units} &
         0.130	 \footnotesize{ $\pm$ 	0.044	} &	0.538	 \footnotesize{ $\pm$ 	0.112	} &	0.149	 \footnotesize{ $\pm$ 	0.054	} \\
        & Ours\textsubscript{MasQCLIP} &
        0.120	 \footnotesize{ $\pm$ 	0.042	} &	0.463	 \footnotesize{ $\pm$ 	0.149	} &	0.142	 \footnotesize{ $\pm$ 	0.049	} \\
            & Ours\textsubscript{SCAN} &
            0.131	 \footnotesize{ $\pm$ 	0.044	} &	0.563	 \footnotesize{ $\pm$ 	0.115	} &	0.149	 \footnotesize{ $\pm$ 	0.054	} \\
        & Ours\textsubscript{SED} &
        0.130	 \footnotesize{ $\pm$ 	0.045	} &	0.571	 \footnotesize{ $\pm$ 	0.138	} &	0.148	 \footnotesize{ $\pm$ 	0.055	} \\
& Ours\textsubscript{CAT-Seg} &
        0.131	 \footnotesize{ $\pm$ 	0.045	} &	0.565	 \footnotesize{ $\pm$ 	0.122	} &	0.149	 \footnotesize{ $\pm$ 	0.054	} \\
         & Ours\textsubscript{OpenSeed} &
          0.130	 \footnotesize{ $\pm$ 	0.045	} &	0.558	 \footnotesize{ $\pm$ 	0.130	} &	0.148	 \footnotesize{ $\pm$ 	0.055	} \\

         \midrule
             4 & Human \cite{LaRosa2023Towards} &
         0.091	 \footnotesize{ $\pm$ 	0.047	} &	0.412	 \footnotesize{ $\pm$ 	0.188	} &	0.114	 \footnotesize{ $\pm$ 	0.067	} \\
                  
                  & Closed \cite{bau2020units} &
         0.088	 \footnotesize{ $\pm$ 	0.042	} &	0.391	 \footnotesize{ $\pm$ 	0.201	} &	0.109	 \footnotesize{ $\pm$ 	0.052	} \\
         & Ours\textsubscript{MasQCLIP} &
        0.082	 \footnotesize{ $\pm$ 	0.036	} &	0.342	 \footnotesize{ $\pm$ 	0.173	} &	0.107	 \footnotesize{ $\pm$ 	0.049	} \\
            & Ours\textsubscript{SCAN} &
            0.088	 \footnotesize{ $\pm$ 	0.043	} &	0.417	 \footnotesize{ $\pm$ 	0.210	} &	0.108	 \footnotesize{ $\pm$ 	0.053	} \\
        & Ours\textsubscript{SED} &
        0.087	 \footnotesize{ $\pm$ 	0.042	} &	0.447	 \footnotesize{ $\pm$ 	0.208	} &	0.105	 \footnotesize{ $\pm$ 	0.051	} \\       
        & Ours\textsubscript{CAT-Seg} &
          0.088	 \footnotesize{ $\pm$ 	0.042	} &	0.432	 \footnotesize{ $\pm$ 	0.200	} &	0.106	 \footnotesize{ $\pm$ 	0.050	} \\
          
          & Ours\textsubscript{OpenSeed} &
        0.086	 \footnotesize{ $\pm$ 	0.042	} &	0.434	 \footnotesize{ $\pm$ 	0.211	} &	0.104	 \footnotesize{ $\pm$ 	0.051	} \\

         \midrule
           5 & Human \cite{LaRosa2023Towards} &
               0.050	 \footnotesize{ $\pm$ 	0.038	} &	0.308	 \footnotesize{ $\pm$ 	0.246	} &	0.068	 \footnotesize{ $\pm$ 	0.057	} \\
                 
        & Closed \cite{bau2020units} &
         0.048	 \footnotesize{ $\pm$ 	0.029	} &	0.277	 \footnotesize{ $\pm$ 	0.239	} &	0.068	 \footnotesize{ $\pm$ 	0.045	} \\
          & Ours\textsubscript{MasQCLIP} &
       0.045	 \footnotesize{ $\pm$ 	0.028	} &	0.290	 \footnotesize{ $\pm$ 	0.188	} &	0.057	 \footnotesize{ $\pm$ 	0.037	} \\
            & Ours\textsubscript{SCAN} &
            0.045	 \footnotesize{ $\pm$ 	0.031	} &	0.333	 \footnotesize{ $\pm$ 	0.271	} &	0.057	 \footnotesize{ $\pm$ 	0.043	} \\
        & Ours\textsubscript{SED} &
       0.044	 \footnotesize{ $\pm$ 	0.029	} &	0.352	 \footnotesize{ $\pm$ 	0.287	} &	0.058	 \footnotesize{ $\pm$ 	0.044	} \\     
        & Ours\textsubscript{CAT-Seg} &
          0.045	 \footnotesize{ $\pm$ 	0.028	} &	0.342	 \footnotesize{ $\pm$ 	0.280	} &	0.060	 \footnotesize{ $\pm$ 	0.043	} \\
        
        & Ours\textsubscript{OpenSeed} &
          0.043	 \footnotesize{ $\pm$ 	0.029	} &	0.358	 \footnotesize{ $\pm$ 	0.283	} &	0.055	 \footnotesize{ $\pm$ 	0.042	} \\
         \bottomrule
    \end{tabular}

\end{table*}

\begin{table*}[!t]
\caption{Avg. and Std. Dev. scores for explanations associated with a model trained on the Place365 dataset using Pascal-Context with 459 labels as a probing dataset. }
   
    \label{tab:pc-459}
    \centering
          
    \begin{tabular}{clccc}
        \toprule
        Cluster & Method& IoU & ActCov &DetAcc \\
         \midrule

         1 & Human \cite{LaRosa2023Towards} &
          0.177	 \footnotesize{ $\pm$ 	0.011	} &	0.247	 \footnotesize{ $\pm$ 	0.012	} &	0.386	 \footnotesize{ $\pm$ 	0.035	} \\
          & Closed \cite{bau2020units} &
          0.188	 \footnotesize{ $\pm$ 	0.012	} &	0.271	 \footnotesize{ $\pm$ 	0.014	} &	0.383	 \footnotesize{ $\pm$ 	0.036	} \\
        & Ours\textsubscript{MasQCLIP} &
       0.179	 \footnotesize{ $\pm$ 	0.012	} &	0.255	 \footnotesize{ $\pm$ 	0.010	} &	0.376	 \footnotesize{ $\pm$ 	0.037	} \\
            & Ours\textsubscript{SCAN} &
             0.177	 \footnotesize{ $\pm$ 	0.012	} &	0.249	 \footnotesize{ $\pm$ 	0.011	} &	0.381	 \footnotesize{ $\pm$ 	0.038	} \\
        & Ours\textsubscript{SED} &
        0.182	 \footnotesize{ $\pm$ 	0.012	} &	0.259	 \footnotesize{ $\pm$ 	0.011	} &	0.381	 \footnotesize{ $\pm$ 	0.038	} \\
          & Ours\textsubscript{CAT-Seg} &
        0.184	 \footnotesize{ $\pm$ 	0.012	} &	0.264	 \footnotesize{ $\pm$ 	0.013	} &	0.380	 \footnotesize{ $\pm$ 	0.038	} \\
                 
        & Ours\textsubscript{OpenSeed} &
         0.193	 \footnotesize{ $\pm$ 	0.012	} &	0.280	 \footnotesize{ $\pm$ 	0.014	} &	0.383	 \footnotesize{ $\pm$ 	0.035	} \\
         \midrule
             2 & Human \cite{LaRosa2023Towards} &
          0.118	 \footnotesize{ $\pm$ 	0.011	} &	0.233	 \footnotesize{ $\pm$ 	0.022	} &	0.194	 \footnotesize{ $\pm$ 	0.024	} \\
                 & Closed \cite{bau2020units} & 
       0.119	 \footnotesize{ $\pm$ 	0.013	} &	0.245	 \footnotesize{ $\pm$ 	0.029	} &	0.190	 \footnotesize{ $\pm$ 	0.027	} \\
                 & Ours\textsubscript{MasQCLIP} &
        0.101	 \footnotesize{ $\pm$ 	0.013	} &	0.220	 \footnotesize{ $\pm$ 	0.024	} &	0.158	 \footnotesize{ $\pm$ 	0.024	} \\
            & Ours\textsubscript{SCAN} &
            0.112	 \footnotesize{ $\pm$ 	0.013	} &	0.217	 \footnotesize{ $\pm$ 	0.025	} &	0.191	 \footnotesize{ $\pm$ 	0.029	} \\
        & Ours\textsubscript{SED} &
        0.115	 \footnotesize{ $\pm$ 	0.013	} &	0.228	 \footnotesize{ $\pm$ 	0.027	} &	0.191	 \footnotesize{ $\pm$ 	0.028	} \\
& Ours\textsubscript{CAT-Seg} &
          0.117	 \footnotesize{ $\pm$ 	0.013	} &	0.236	 \footnotesize{ $\pm$ 	0.028	} &	0.192	 \footnotesize{ $\pm$ 	0.028	} \\
 & Ours\textsubscript{OpenSeed} &
         0.121	 \footnotesize{ $\pm$ 	0.013	} &	0.253	 \footnotesize{ $\pm$ 	0.031	} &	0.192	 \footnotesize{ $\pm$ 	0.028	} \\
         \midrule
             3 & Human \cite{LaRosa2023Towards} &
          0.106	 \footnotesize{ $\pm$ 	0.018	} &	0.220	 \footnotesize{ $\pm$ 	0.049	} &	0.180	 \footnotesize{ $\pm$ 	0.047	} \\
         & Closed \cite{bau2020units} &
      0.105	 \footnotesize{ $\pm$ 	0.020	} &	0.216	 \footnotesize{ $\pm$ 	0.055	} &	0.182	 \footnotesize{ $\pm$ 	0.048	} \\
        & Ours\textsubscript{MasQCLIP} &
        0.086	 \footnotesize{ $\pm$ 	0.019	} &	0.158	 \footnotesize{ $\pm$ 	0.043	} &	0.177	 \footnotesize{ $\pm$ 	0.059	} \\
            & Ours\textsubscript{SCAN} &
           0.103	 \footnotesize{ $\pm$ 	0.019	} &	0.204	 \footnotesize{ $\pm$ 	0.051	} &	0.185	 \footnotesize{ $\pm$ 	0.048	} \\
        & Ours\textsubscript{SED} &
        0.104	 \footnotesize{ $\pm$ 	0.020	} &	0.212	 \footnotesize{ $\pm$ 	0.051	} &	0.181	 \footnotesize{ $\pm$ 	0.046	} \\
& Ours\textsubscript{CAT-Seg} &
        0.105	 \footnotesize{ $\pm$ 	0.020	} &	0.215	 \footnotesize{ $\pm$ 	0.051	} &	0.181	 \footnotesize{ $\pm$ 	0.045	} \\
         & Ours\textsubscript{OpenSeed} &
         0.106	 \footnotesize{ $\pm$ 	0.020	} &	0.220	 \footnotesize{ $\pm$ 	0.055	} &	0.181	 \footnotesize{ $\pm$ 	0.045	} \\

         \midrule
             4 & Human \cite{LaRosa2023Towards} &
        0.112	 \footnotesize{ $\pm$ 	0.055	} &	0.251	 \footnotesize{ $\pm$ 	0.088	} &	0.175	 \footnotesize{ $\pm$ 	0.087	} \\
                  
                  & Closed \cite{bau2020units} &
                   0.113	 \footnotesize{ $\pm$ 	0.054	} &	0.250	 \footnotesize{ $\pm$ 	0.095	} &	0.177	 \footnotesize{ $\pm$ 	0.084	} \\
         & Ours\textsubscript{MasQCLIP} &
         0.100	 \footnotesize{ $\pm$ 	0.051	} &	0.189	 \footnotesize{ $\pm$ 	0.086	} &	0.175	 \footnotesize{ $\pm$ 	0.083	} \\
        & Ours\textsubscript{SCAN} &  
        0.112	 \footnotesize{ $\pm$ 	0.055	} &	0.241	 \footnotesize{ $\pm$ 	0.097	} &	0.177	 \footnotesize{ $\pm$ 	0.084	} \\
            & Ours\textsubscript{SED} &
            0.113	 \footnotesize{ $\pm$ 	0.056	} &	0.246	 \footnotesize{ $\pm$ 	0.097	} &	0.178	 \footnotesize{ $\pm$ 	0.088	} \\
     
        & Ours\textsubscript{CAT-Seg} &
          0.113	 \footnotesize{ $\pm$ 	0.056	} &	0.250	 \footnotesize{ $\pm$ 	0.096	} &	0.177	 \footnotesize{ $\pm$ 	0.087	} \\
          
          & Ours\textsubscript{OpenSeed} &
        0.113	 \footnotesize{ $\pm$ 	0.055	} &	0.254	 \footnotesize{ $\pm$ 	0.096	} &	0.177	 \footnotesize{ $\pm$ 	0.088	} \\

         \midrule
           5 & Human \cite{LaRosa2023Towards} &
                0.077	 \footnotesize{ $\pm$ 	0.055	} &	0.280	 \footnotesize{ $\pm$ 	0.203	} &	0.103	 \footnotesize{ $\pm$ 	0.063	} \\
                 
        & Closed \cite{bau2020units} &
       0.077	 \footnotesize{ $\pm$ 	0.055	} &	0.301	 \footnotesize{ $\pm$ 	0.195	} &	0.102	 \footnotesize{ $\pm$ 	0.071	} \\

          & Ours\textsubscript{MasQCLIP} &
                0.078	 \footnotesize{ $\pm$ 	0.052	} &	0.233	 \footnotesize{ $\pm$ 	0.191	} &	0.120	 \footnotesize{ $\pm$ 	0.065	} \\ 
            & Ours\textsubscript{SCAN} &
             0.079	 \footnotesize{ $\pm$ 	0.055	} &	0.286	 \footnotesize{ $\pm$ 	0.198	} &	0.105	 \footnotesize{ $\pm$ 	0.068	} \\
        & Ours\textsubscript{SED} &
       0.080	 \footnotesize{ $\pm$ 	0.056	} &	0.279	 \footnotesize{ $\pm$ 	0.206	} &	0.110	 \footnotesize{ $\pm$ 	0.069	} \\    
        & Ours\textsubscript{CAT-Seg} &
          0.079	 \footnotesize{ $\pm$ 	0.056	} &	0.286	 \footnotesize{ $\pm$ 	0.205	} &	0.109	 \footnotesize{ $\pm$ 	0.071	} \\
        
        & Ours\textsubscript{OpenSeed} &
          0.078	 \footnotesize{ $\pm$ 	0.056	} &	0.290	 \footnotesize{ $\pm$ 	0.209	} &	0.108	 \footnotesize{ $\pm$ 	0.074	} \\
         \bottomrule
    \end{tabular}
\end{table*}

\begin{table*}[!t]
\caption{Avg. and Std. Dev. scores for explanations associated with a model trained on the Place365 dataset using COCO-Stuff as a probing dataset. }
   
    \label{tab:coco-stuff}
    \centering
          
    \begin{tabular}{clccc}
        \toprule
        Cluster & Method& IoU & ActCov &DetAcc \\
         \midrule

         1 & Human \cite{LaRosa2023Towards} &
          0.152	 \footnotesize{ $\pm$ 	0.008	} &	0.205	 \footnotesize{ $\pm$ 	0.009	} &	0.374	 \footnotesize{ $\pm$ 	0.032	} \\
          & Closed \cite{bau2020units} &
       0.187	 \footnotesize{ $\pm$ 	0.010	} &	0.270	 \footnotesize{ $\pm$ 	0.009	} &	0.377	 \footnotesize{ $\pm$ 	0.032	} \\
        & Ours\textsubscript{MasQCLIP} &
       0.107	 \footnotesize{ $\pm$ 	0.005	} &	0.129	 \footnotesize{ $\pm$ 	0.005	} &	0.384	 \footnotesize{ $\pm$ 	0.032	} \\
            & Ours\textsubscript{SCAN} &
            0.145	 \footnotesize{ $\pm$ 	0.008	} &	0.191	 \footnotesize{ $\pm$ 	0.009	} &	0.376	 \footnotesize{ $\pm$ 	0.031	} \\
        & Ours\textsubscript{SED} &
       0.165	 \footnotesize{ $\pm$ 	0.009	} &	0.228	 \footnotesize{ $\pm$ 	0.010	} &	0.373	 \footnotesize{ $\pm$ 	0.032	} \\
          & Ours\textsubscript{CAT-Seg} &
         0.164	 \footnotesize{ $\pm$ 	0.009	} &	0.228	 \footnotesize{ $\pm$ 	0.009	} &	0.372	 \footnotesize{ $\pm$ 	0.032	} \\
                 
        & Ours\textsubscript{OpenSeed} &
         0.187	 \footnotesize{ $\pm$ 	0.010	} &	0.270	 \footnotesize{ $\pm$ 	0.010	} &	0.378	 \footnotesize{ $\pm$ 	0.032	} \\
         \midrule
            
             2 & Human \cite{LaRosa2023Towards} &
          0.104	 \footnotesize{ $\pm$ 	0.010	} &	0.189	 \footnotesize{ $\pm$ 	0.020	} &	0.188	 \footnotesize{ $\pm$ 	0.021	} \\
                 & Closed \cite{bau2020units} & 
        0.117	 \footnotesize{ $\pm$ 	0.012	} &	0.243	 \footnotesize{ $\pm$ 	0.029	} &	0.186	 \footnotesize{ $\pm$ 	0.021	} \\
                 & Ours\textsubscript{MasQCLIP} &
       0.078	 \footnotesize{ $\pm$ 	0.008	} &	0.119	 \footnotesize{ $\pm$ 	0.011	} &	0.185	 \footnotesize{ $\pm$ 	0.027	} \\
            & Ours\textsubscript{SCAN} &
             0.103	 \footnotesize{ $\pm$ 	0.010	} &	0.185	 \footnotesize{ $\pm$ 	0.017	} &	0.188	 \footnotesize{ $\pm$ 	0.021	} \\
        & Ours\textsubscript{SED} &
        0.111	 \footnotesize{ $\pm$ 	0.011	} &	0.215	 \footnotesize{ $\pm$ 	0.021	} &	0.189	 \footnotesize{ $\pm$ 	0.022	} \\
& Ours\textsubscript{CAT-Seg} &
         0.111	 \footnotesize{ $\pm$ 	0.011	} &	0.213	 \footnotesize{ $\pm$ 	0.022	} &	0.189	 \footnotesize{ $\pm$ 	0.021	} \\
 & Ours\textsubscript{OpenSeed} &
         0.117	 \footnotesize{ $\pm$ 	0.012	} &	0.242	 \footnotesize{ $\pm$ 	0.029	} &	0.186	 \footnotesize{ $\pm$ 	0.022	} \\
         \midrule
         
             3 & Human \cite{LaRosa2023Towards} &
         0.090	 \footnotesize{ $\pm$ 	0.019	} &	0.182	 \footnotesize{ $\pm$ 	0.046	} &	0.160	 \footnotesize{ $\pm$ 	0.045	} \\
         & Closed \cite{bau2020units} &
         0.098	 \footnotesize{ $\pm$ 	0.022	} &	0.220	 \footnotesize{ $\pm$ 	0.058	} &	0.158	 \footnotesize{ $\pm$ 	0.045	} \\
        & Ours\textsubscript{MasQCLIP} &
        0.074	 \footnotesize{ $\pm$ 	0.016	} &	0.119	 \footnotesize{ $\pm$ 	0.023	} &	0.169	 \footnotesize{ $\pm$ 	0.045	} \\
            & Ours\textsubscript{SCAN} &
            0.090	 \footnotesize{ $\pm$ 	0.018	} &	0.180	 \footnotesize{ $\pm$ 	0.047	} &	0.164	 \footnotesize{ $\pm$ 	0.047	} \\
        & Ours\textsubscript{SED} &
       0.095	 \footnotesize{ $\pm$ 	0.020	} &	0.203	 \footnotesize{ $\pm$ 	0.052	} &	0.161	 \footnotesize{ $\pm$ 	0.046	} \\
& Ours\textsubscript{CAT-Seg} &
        0.095	 \footnotesize{ $\pm$ 	0.020	} &	0.201	 \footnotesize{ $\pm$ 	0.052	} &	0.162	 \footnotesize{ $\pm$ 	0.047	} \\
         & Ours\textsubscript{OpenSeed} &
          0.097	 \footnotesize{ $\pm$ 	0.021	} &	0.217	 \footnotesize{ $\pm$ 	0.056	} &	0.159	 \footnotesize{ $\pm$ 	0.046	} \\

         \midrule
             4 & Human \cite{LaRosa2023Towards} &
         0.089	 \footnotesize{ $\pm$ 	0.037	} &	0.185	 \footnotesize{ $\pm$ 	0.084	} &	0.166	 \footnotesize{ $\pm$ 	0.070	} \\
                  
                  & Closed \cite{bau2020units} &
          0.094	 \footnotesize{ $\pm$ 	0.038	} &	0.213	 \footnotesize{ $\pm$ 	0.087	} &	0.160	 \footnotesize{ $\pm$ 	0.069	} \\
         & Ours\textsubscript{MasQCLIP} &
       0.086	 \footnotesize{ $\pm$ 	0.035	} &	0.150	 \footnotesize{ $\pm$ 	0.065	} &	0.174	 \footnotesize{ $\pm$ 	0.065	} \\
            & Ours\textsubscript{SCAN} &
             0.093	 \footnotesize{ $\pm$ 	0.038	} &	0.188	 \footnotesize{ $\pm$ 	0.082	} &	0.172	 \footnotesize{ $\pm$ 	0.071	} \\
        & Ours\textsubscript{SED} &
       0.093	 \footnotesize{ $\pm$ 	0.038	} &	0.203	 \footnotesize{ $\pm$ 	0.084	} &	0.164	 \footnotesize{ $\pm$ 	0.073	} \\       
        & Ours\textsubscript{CAT-Seg} &
        0.093	 \footnotesize{ $\pm$ 	0.038	} &	0.201	 \footnotesize{ $\pm$ 	0.085	} &	0.165	 \footnotesize{ $\pm$ 	0.071	} \\
          
          & Ours\textsubscript{OpenSeed} &
       0.094	 \footnotesize{ $\pm$ 	0.038	} &	0.211	 \footnotesize{ $\pm$ 	0.089	} &	0.161	 \footnotesize{ $\pm$ 	0.070	} \\

         \midrule
           5 & Human \cite{LaRosa2023Towards} &
              0.078	 \footnotesize{ $\pm$ 	0.048	} &	0.197	 \footnotesize{ $\pm$ 	0.135	} &	0.127	 \footnotesize{ $\pm$ 	0.071	} \\
                 
        & Closed \cite{bau2020units} &
          0.078	 \footnotesize{ $\pm$ 	0.047	} &	0.225	 \footnotesize{ $\pm$ 	0.148	} &	0.120	 \footnotesize{ $\pm$ 	0.068	} \\
          & Ours\textsubscript{MasQCLIP} &
       0.078	 \footnotesize{ $\pm$ 	0.040	} &	0.200	 \footnotesize{ $\pm$ 	0.111	} &	0.119	 \footnotesize{ $\pm$ 	0.058	} \\
            & Ours\textsubscript{SCAN} &
            0.080	 \footnotesize{ $\pm$ 	0.046	} &	0.220	 \footnotesize{ $\pm$ 	0.132	} &	0.123	 \footnotesize{ $\pm$ 	0.068	} \\
        & Ours\textsubscript{SED} &
       0.080	 \footnotesize{ $\pm$ 	0.047	} &	0.209	 \footnotesize{ $\pm$ 	0.129	} &	0.125	 \footnotesize{ $\pm$ 	0.068	} \\     
        & Ours\textsubscript{CAT-Seg} &
         0.080	 \footnotesize{ $\pm$ 	0.047	} &	0.221	 \footnotesize{ $\pm$ 	0.141	} &	0.123	 \footnotesize{ $\pm$ 	0.068	} \\
        
        & Ours\textsubscript{OpenSeed} &
        0.079	 \footnotesize{ $\pm$ 	0.048	} &	0.221	 \footnotesize{ $\pm$ 	0.142	} &	0.122	 \footnotesize{ $\pm$ 	0.071	} \\
         \bottomrule
    \end{tabular}
\end{table*}

\begin{table*}[!t]
\caption{Avg. scores for explanations associated with a model trained on the Place365 dataset using VOC2012 as a probing dataset. }
    \label{tab:voc2012}
    \centering
          
    \begin{tabular}{clccc}
        \toprule
        Cluster & Method& IoU & ActCov &DetAcc \\
         \midrule

         1 & Human \cite{LaRosa2023Towards} &
         0.077	 \footnotesize{ $\pm$ 	0.007	} &	0.090	 \footnotesize{ $\pm$ 	0.008	} &	0.362	 \footnotesize{ $\pm$ 	0.040	} \\
          & Closed \cite{bau2020units} &
         0.193	 \footnotesize{ $\pm$ 	0.014	} &	0.280	 \footnotesize{ $\pm$ 	0.014	} &	0.385	 \footnotesize{ $\pm$ 	0.039	} \\
        & Ours\textsubscript{MasQCLIP} &
        0.269	 \footnotesize{ $\pm$ 	0.022	} &	0.499	 \footnotesize{ $\pm$ 	0.015	} &	0.369	 \footnotesize{ $\pm$ 	0.038	} \\
            & Ours\textsubscript{SCAN} &
             0.209	 \footnotesize{ $\pm$ 	0.014	} &	0.324	 \footnotesize{ $\pm$ 	0.015	} &	0.371	 \footnotesize{ $\pm$ 	0.036	} \\
        & Ours\textsubscript{SED} &
       0.238	 \footnotesize{ $\pm$ 	0.018	} &	0.407	 \footnotesize{ $\pm$ 	0.015	} &	0.366	 \footnotesize{ $\pm$ 	0.038	} \\
          & Ours\textsubscript{CAT-Seg} &
         0.189	 \footnotesize{ $\pm$ 	0.011	} &	0.277	 \footnotesize{ $\pm$ 	0.009	} &	0.377	 \footnotesize{ $\pm$ 	0.036	} \\
                 
        & Ours\textsubscript{OpenSeed} &
        0.276	 \footnotesize{ $\pm$ 	0.022	} &	0.522	 \footnotesize{ $\pm$ 	0.016	} &	0.370	 \footnotesize{ $\pm$ 	0.036	} \\
         \midrule
             2 & Human \cite{LaRosa2023Towards} &
          0.074	 \footnotesize{ $\pm$ 	0.010	} &	0.107	 \footnotesize{ $\pm$ 	0.014	} &	0.199	 \footnotesize{ $\pm$ 	0.029	} \\
                 & Closed \cite{bau2020units} & 
         0.115	 \footnotesize{ $\pm$ 	0.014	} &	0.232	 \footnotesize{ $\pm$ 	0.034	} &	0.190	 \footnotesize{ $\pm$ 	0.030	} \\
                 & Ours\textsubscript{MasQCLIP} &
       0.132	 \footnotesize{ $\pm$ 	0.015	} &	0.438	 \footnotesize{ $\pm$ 	0.034	} &	0.160	 \footnotesize{ $\pm$ 	0.020	} \\
            & Ours\textsubscript{SCAN} &
            0.132	 \footnotesize{ $\pm$ 	0.014	} &	0.296	 \footnotesize{ $\pm$ 	0.038	} &	0.195	 \footnotesize{ $\pm$ 	0.028	} \\
        & Ours\textsubscript{SED} &
       0.134	 \footnotesize{ $\pm$ 	0.014	} &	0.335	 \footnotesize{ $\pm$ 	0.051	} &	0.185	 \footnotesize{ $\pm$ 	0.023	} \\
& Ours\textsubscript{CAT-Seg} &
          0.131	 \footnotesize{ $\pm$ 	0.014	} &	0.282	 \footnotesize{ $\pm$ 	0.033	} &	0.201	 \footnotesize{ $\pm$ 	0.033	} \\
 & Ours\textsubscript{OpenSeed} &
        0.137	 \footnotesize{ $\pm$ 	0.016	} &	0.479	 \footnotesize{ $\pm$ 	0.042	} &	0.162	 \footnotesize{ $\pm$ 	0.021	} \\
         \midrule
             3 & Human \cite{LaRosa2023Towards} &
         0.082	 \footnotesize{ $\pm$ 	0.018	} &	0.134	 \footnotesize{ $\pm$ 	0.031	} &	0.184	 \footnotesize{ $\pm$ 	0.051	} \\
         & Closed \cite{bau2020units} &
        0.105	 \footnotesize{ $\pm$ 	0.021	} &	0.201	 \footnotesize{ $\pm$ 	0.051	} &	0.191	 \footnotesize{ $\pm$ 	0.046	} \\
        & Ours\textsubscript{MasQCLIP} &
        0.107	 \footnotesize{ $\pm$ 	0.020	} &	0.272	 \footnotesize{ $\pm$ 	0.100	} &	0.166	 \footnotesize{ $\pm$ 	0.047	} \\
            & Ours\textsubscript{SCAN} &
            0.117	 \footnotesize{ $\pm$ 	0.021	} &	0.290	 \footnotesize{ $\pm$ 	0.060	} &	0.169	 \footnotesize{ $\pm$ 	0.037	} \\
        & Ours\textsubscript{SED} &
      0.115	 \footnotesize{ $\pm$ 	0.021	} &	0.294	 \footnotesize{ $\pm$ 	0.066	} &	0.164	 \footnotesize{ $\pm$ 	0.037	} \\
& Ours\textsubscript{CAT-Seg} &
        0.117	 \footnotesize{ $\pm$ 	0.022	} &	0.305	 \footnotesize{ $\pm$ 	0.051	} &	0.163	 \footnotesize{ $\pm$ 	0.038	} \\
         & Ours\textsubscript{OpenSeed} &
          0.110	 \footnotesize{ $\pm$ 	0.020	} &	0.283	 \footnotesize{ $\pm$ 	0.100	} &	0.164	 \footnotesize{ $\pm$ 	0.040	} \\

         \midrule
             4 & Human \cite{LaRosa2023Towards} &
       0.098	 \footnotesize{ $\pm$ 	0.050	} &	0.183	 \footnotesize{ $\pm$ 	0.076	} &	0.186	 \footnotesize{ $\pm$ 	0.095	} \\
                  
                  & Closed \cite{bau2020units} &
          0.117	 \footnotesize{ $\pm$ 	0.051	} &	0.248	 \footnotesize{ $\pm$ 	0.089	} &	0.188	 \footnotesize{ $\pm$ 	0.082	} \\
         & Ours\textsubscript{MasQCLIP} &
       0.104	 \footnotesize{ $\pm$ 	0.049	} &	0.272	 \footnotesize{ $\pm$ 	0.104	} &	0.146	 \footnotesize{ $\pm$ 	0.069	} \\
            & Ours\textsubscript{SCAN} &
           0.103	 \footnotesize{ $\pm$ 	0.042	} &	0.327	 \footnotesize{ $\pm$ 	0.103	} &	0.136	 \footnotesize{ $\pm$ 	0.062	} \\
        & Ours\textsubscript{SED} &
       0.100	 \footnotesize{ $\pm$ 	0.041	} &	0.327	 \footnotesize{ $\pm$ 	0.103	} &	0.128	 \footnotesize{ $\pm$ 	0.054	} \\      
        & Ours\textsubscript{CAT-Seg} &
         0.099	 \footnotesize{ $\pm$ 	0.043	} &	0.322	 \footnotesize{ $\pm$ 	0.108	} &	0.130	 \footnotesize{ $\pm$ 	0.059	} \\
          
          & Ours\textsubscript{OpenSeed} &
        0.101	 \footnotesize{ $\pm$ 	0.044	} &	0.288	 \footnotesize{ $\pm$ 	0.102	} &	0.138	 \footnotesize{ $\pm$ 	0.065	} \\

         \midrule
           5 & Human \cite{LaRosa2023Towards} &
              0.077	 \footnotesize{ $\pm$ 	0.062	} &	0.247	 \footnotesize{ $\pm$ 	0.173	} &	0.106	 \footnotesize{ $\pm$ 	0.078	} \\
                 
        & Closed \cite{bau2020units} &
         0.084	 \footnotesize{ $\pm$ 	0.053	} &	0.296	 \footnotesize{ $\pm$ 	0.177	} &	0.114	 \footnotesize{ $\pm$ 	0.068	} \\
          & Ours\textsubscript{MasQCLIP} &
       0.058	 \footnotesize{ $\pm$ 	0.040	} &	0.322	 \footnotesize{ $\pm$ 	0.212	} &	0.067	 \footnotesize{ $\pm$ 	0.045	} \\
            & Ours\textsubscript{SCAN} &
        0.055	 \footnotesize{ $\pm$ 	0.039	} &	0.343	 \footnotesize{ $\pm$ 	0.202	} &	0.064	 \footnotesize{ $\pm$ 	0.046	} \\
        & Ours\textsubscript{SED} &
       0.051	 \footnotesize{ $\pm$ 	0.032	} &	0.370	 \footnotesize{ $\pm$ 	0.210	} &	0.058	 \footnotesize{ $\pm$ 	0.038	} \\     
        & Ours\textsubscript{CAT-Seg} &
         0.051	 \footnotesize{ $\pm$ 	0.030	} &	0.357	 \footnotesize{ $\pm$ 	0.219	} &	0.058	 \footnotesize{ $\pm$ 	0.034	} \\
        
        & Ours\textsubscript{OpenSeed} &
         0.057	 \footnotesize{ $\pm$ 	0.037	} &	0.282	 \footnotesize{ $\pm$ 	0.199	} &	0.070	 \footnotesize{ $\pm$ 	0.044	} \\
         \bottomrule
    \end{tabular}
\end{table*}

\Cref{tab:ade20kfull,tab:mapillary,tab:cityscapes,tab:coco-stuff,tab:pc-459,tab:voc2012} compare the competitors and the framework's implementations using the IoU, Activation coverage and Detection Accuracy metrics.
Similarly to \Cref{appx:eval_ovseg}, we observe comparable results across all models and datasets, confirming the generality of the good performance of our framework.

\subsection{Additional Probed Models}
\label{appx:eval_probed}

\begin{table*}[!t]
          \caption{Avg. and Std. Dev. scores for explanations associated with a DenseNet161 model trained on the Place365 dataset using Ade20K as a probing dataset. }
   
    \label{tab:densenet}
    \centering

    \begin{tabular}{clccc}
        \toprule
        Cluster & Method& IoU & ActCov &DetAcc \\
         \midrule

         1 & Human \cite{LaRosa2023Towards} &
          0.128	 \footnotesize{ $\pm$ 	0.048	} &	0.296	 \footnotesize{ $\pm$ 	0.120	} &	0.198	 \footnotesize{ $\pm$ 	0.072	} \\
          & Closed \cite{bau2020units} &
         0.134	 \footnotesize{ $\pm$ 	0.049	} &	0.304	 \footnotesize{ $\pm$ 	0.119	} &	0.205	 \footnotesize{ $\pm$ 	0.067	} \\
          & Ours &
         0.133	 \footnotesize{ $\pm$ 	0.047	} &	0.294	 \footnotesize{ $\pm$ 	0.119	} &	0.208	 \footnotesize{ $\pm$ 	0.068	} \\
         \midrule
             2 & Human \cite{LaRosa2023Towards} &
          0.209	 \footnotesize{ $\pm$ 	0.041	} &	0.345	 \footnotesize{ $\pm$ 	0.040	} &	0.361	 \footnotesize{ $\pm$ 	0.108	} \\
                 & Closed \cite{bau2020units} & 
        0.205	 \footnotesize{ $\pm$ 	0.039	} &	0.334	 \footnotesize{ $\pm$ 	0.047	} &	0.360	 \footnotesize{ $\pm$ 	0.100	} \\
& Ours &
         0.204	 \footnotesize{ $\pm$ 	0.039	} &	0.325	 \footnotesize{ $\pm$ 	0.045	} &	0.369	 \footnotesize{ $\pm$ 	0.107	} \\
         \midrule
             3 & Human \cite{LaRosa2023Towards} &
         0.207	 \footnotesize{ $\pm$ 	0.026	} &	0.345	 \footnotesize{ $\pm$ 	0.024	} &	0.344	 \footnotesize{ $\pm$ 	0.061	} \\
         & Closed \cite{bau2020units} &
        0.204	 \footnotesize{ $\pm$ 	0.026	} &	0.336	 \footnotesize{ $\pm$ 	0.023	} &	0.346	 \footnotesize{ $\pm$ 	0.062	} \\
& Ours &
       0.201	 \footnotesize{ $\pm$ 	0.026	} &	0.322	 \footnotesize{ $\pm$ 	0.025	} &	0.353	 \footnotesize{ $\pm$ 	0.061	} \\

         \midrule
             4 & Human \cite{LaRosa2023Towards} &
        0.177	 \footnotesize{ $\pm$ 	0.058	} &	0.325	 \footnotesize{ $\pm$ 	0.080	} &	0.290	 \footnotesize{ $\pm$ 	0.102	} \\
                  
                  & Closed \cite{bau2020units} &
          0.175	 \footnotesize{ $\pm$ 	0.056	} &	0.320	 \footnotesize{ $\pm$ 	0.075	} &	0.286	 \footnotesize{ $\pm$ 	0.099	} \\
        
        & Ours &
       0.178	 \footnotesize{ $\pm$ 	0.057	} &	0.315	 \footnotesize{ $\pm$ 	0.072	} &	0.299	 \footnotesize{ $\pm$ 	0.104	} \\

         \midrule
           5 & Human \cite{LaRosa2023Towards} &
             0.103	 \footnotesize{ $\pm$ 	0.047	} &	0.274	 \footnotesize{ $\pm$ 	0.118	} &	0.158	 \footnotesize{ $\pm$ 	0.086	} \\
                 
        & Closed \cite{bau2020units} &
      0.108	 \footnotesize{ $\pm$ 	0.048	} &	0.287	 \footnotesize{ $\pm$ 	0.119	} &	0.158	 \footnotesize{ $\pm$ 	0.075	} \\
        
        & Ours &
         0.108	 \footnotesize{ $\pm$ 	0.048	} &	0.286	 \footnotesize{ $\pm$ 	0.116	} &	0.160	 \footnotesize{ $\pm$ 	0.082	} \\
        
         \bottomrule
    \end{tabular}
\end{table*}
\begin{table*}[!t]
         \caption{Avg. and Std. Dev. scores for explanations associated with an AlexNet model trained on the Place365 dataset using Ade20K as a probing dataset. }
   
    \label{tab:alexnet}
    \centering
 
    \begin{tabular}{clccc}
        \toprule
        Cluster & Method& IoU & ActCov &DetAcc \\
         \midrule

         1 & Human \cite{LaRosa2023Towards} &
          0.192	 \footnotesize{ $\pm$ 	0.024	} &	0.333	 \footnotesize{ $\pm$ 	0.024	} &	0.314	 \footnotesize{ $\pm$ 	0.053	} \\
          & Closed \cite{bau2020units} &
          0.188	 \footnotesize{ $\pm$ 	0.023	} &	0.322	 \footnotesize{ $\pm$ 	0.029	} &	0.314	 \footnotesize{ $\pm$ 	0.050	} \\
          & Ours &
        0.184	 \footnotesize{ $\pm$ 	0.022	} &	0.309	 \footnotesize{ $\pm$ 	0.020	} &	0.317	 \footnotesize{ $\pm$ 	0.054	} \\
         \midrule
             2 & Human \cite{LaRosa2023Towards} &
         0.115	 \footnotesize{ $\pm$ 	0.026	} &	0.300	 \footnotesize{ $\pm$ 	0.078	} &	0.161	 \footnotesize{ $\pm$ 	0.035	} \\
                 & Closed \cite{bau2020units} & 
        0.117	 \footnotesize{ $\pm$ 	0.025	} &	0.292	 \footnotesize{ $\pm$ 	0.065	} &	0.167	 \footnotesize{ $\pm$ 	0.038	} \\
& Ours &
         0.117	 \footnotesize{ $\pm$ 	0.025	} &	0.287	 \footnotesize{ $\pm$ 	0.075	} &	0.169	 \footnotesize{ $\pm$ 	0.035	} \\
         \midrule
             3 & Human \cite{LaRosa2023Towards} &
         0.097	 \footnotesize{ $\pm$ 	0.028	} &	0.262	 \footnotesize{ $\pm$ 	0.100	} &	0.142	 \footnotesize{ $\pm$ 	0.043	} \\
         & Closed \cite{bau2020units} &
0.101	 \footnotesize{ $\pm$ 	0.029	} &	0.277	 \footnotesize{ $\pm$ 	0.091	} &	0.143	 \footnotesize{ $\pm$ 	0.042	} \\
& Ours &
       0.102	 \footnotesize{ $\pm$ 	0.029	} &	0.272	 \footnotesize{ $\pm$ 	0.099	} &	0.145	 \footnotesize{ $\pm$ 	0.040	} \\

         \midrule
             4 & Human \cite{LaRosa2023Towards} &
         0.079	 \footnotesize{ $\pm$ 	0.028	} &	0.233	 \footnotesize{ $\pm$ 	0.126	} &	0.120	 \footnotesize{ $\pm$ 	0.039	} \\
                  
                  & Closed \cite{bau2020units} &
          0.082	 \footnotesize{ $\pm$ 	0.028	} &	0.265	 \footnotesize{ $\pm$ 	0.134	} &	0.117	 \footnotesize{ $\pm$ 	0.038	} \\
        
        & Ours &
         0.082	 \footnotesize{ $\pm$ 	0.028	} &	0.251	 \footnotesize{ $\pm$ 	0.127	} &	0.121	 \footnotesize{ $\pm$ 	0.039	} \\

         \midrule
           5 & Human \cite{LaRosa2023Towards} &
               0.055	 \footnotesize{ $\pm$ 	0.028	} &	0.226	 \footnotesize{ $\pm$ 	0.191	} &	0.093	 \footnotesize{ $\pm$ 	0.064	} \\
                 
        & Closed \cite{bau2020units} &
         0.054	 \footnotesize{ $\pm$ 	0.023	} &	0.254	 \footnotesize{ $\pm$ 	0.186	} &	0.080	 \footnotesize{ $\pm$ 	0.049	} \\
        
        & Ours &
        0.059	 \footnotesize{ $\pm$ 	0.026	} &	0.245	 \footnotesize{ $\pm$ 	0.183	} &	0.094	 \footnotesize{ $\pm$ 	0.057	} \\
        
         \bottomrule
    \end{tabular}
\end{table*}
In this section, we report the results explaining different probed models. Following ~\cite{Mu2020,LaRosa2023Towards}, we compute explanations scores for {DenseNet161~\cite{Huang_2017_CVPR} and AlexNet~\cite{Krizhevsky2012} pre-trained on the Place365 dataset~\cite{zhou2017places}. We report the results for our framework using the same configuration as in the main text, \cite{LaRosa2023Towards}, and the closed approach~\cite{bau2020units}.
Specifically, we randomly extract 50 neurons for each probed model and we generate explanations for those neurons using as a probing dataset the validation split of Ade20K~\cite{Zhou2017}. \Cref{tab:densenet,tab:alexnet} confirm the comparable performance of the framework with respect to the competitors, making the insights independent of the probed model in use.

\subsection{Metrics Details and Additional Metrics}
\label{appx:eval_metrics}
This section provides the formalization of Detection Accuracy and Activation Coverage and introduces and compares the competitors using two additional metrics: Sample Coverage and Explanation Coverage~\cite{LaRosa2023Towards}. We chose these metrics because they have been used by previous literature to study cluster-level explanations~\cite{LaRosa2023Towards} and allow us to perform a pixel-level comparison of the different segmentation masks produced by different competitors.

We use the same notation introduced in Section 3. However, because Sample Coverage and Explanation Coverage are computed per sample, we need to introduce an additional notation. Namely, we use $\mathbb{M}^x$ to indicate the set of binarized segmentation masks associated with the sample $x$ and $\mathbb{A}^x$ to indicate the set of binarized activations associated with the sample $x$.

\paragraph{Detection Accuracy} quantifies the percentage of label annotations recognized within the activation range. A high value indicates that most of the label's masks are detected by the neuron using the given activation range.
\begin{equation}
    DetAcc(L, \mathbb{A}, \mathbb{M} ) = \frac{|\mathbb{A} \cap \theta(\mathbb{M}, L)|}{ |\theta(\mathbb{M}, L)|}
\end{equation}

\paragraph{Activation Coverage} measures the percentage of neuron activations within the annotated label regions. A high value indicates that the label ``dominates'' large parts of the activation range (i.e., there is a strong mapping). 
\begin{equation}
    ActCov(L,  \mathbb{A}, \mathbb{M}) =\frac{|\mathbb{A} \cap \theta(\mathbb{M}, L)|}{ |\mathbb{A}|}
\end{equation}

\paragraph{Samples Coverage} calculates the ratio of samples in the probing dataset that are captured by the explanation and where the neuron activation falls within the activation and the total number of samples satisfying the explanation
\begin{equation}
\begin{split}
        SampleCov(L,& \mathbb{A}, \mathbb{M} ,  \mathfrak{D} ) = \\
         &\frac{|\{x \in \mathfrak{D}: |\mathbb{A}^x \cap \theta(\mathbb{M}^x, L)|>0\}|}{|\{x \in \mathfrak{D}: |\theta(\mathbb{M}^x, L)|>0\}|}
\end{split}  
\end{equation}

\paragraph{Explanation Coverage} calculates the ratio of samples in the probing dataset that are captured by the explanation and where the neuron activation falls within the activation range and the total number of samples where the neuron activation falls within the activation range.
\begin{equation}
\begin{split}
        ExplCov(L,& \mathbb{A}, \mathbb{M}, \mathfrak{D} )   = \\
   & \frac{|\{x \in \mathfrak{D}: |\mathbb{A}^x \cap \theta(\mathbb{M}^x, L)|>0\}|}{|\{x \in \mathfrak{D}: |\mathbb{A}^x|>0\}|}
\end{split}
\end{equation}
}
\subsubsection{Results}
As shown in \Cref{tab:additional_metrics}, the results for the additional metrics are similar to the ones reported in the main text for the other metrics. Thus, considering the large standard deviation of these metrics, the results can be considered comparable. 
\begin{table}[!t]
          \caption{Avg. and Std. Dev Sample Coverage and Explanation Coverage for explanations associated with a model trained on the Place365 dataset using Ade20K as a probing dataset.}
   
    \label{tab:additional_metrics}
    \centering

    \begin{tabular}{clll}
        \toprule
        Cluster & Method& SampleCov & ExplCov \\
         \midrule

         1 & Human \cite{LaRosa2023Towards} &
         0.911	 \footnotesize{ $\pm$ 	0.029	} &	0.873	 \footnotesize{ $\pm$ 	0.059	}\\
          & Closed \cite{bau2020units} &
          0.904	 \footnotesize{ $\pm$ 	0.028	} &	0.872	 \footnotesize{ $\pm$ 	0.078	}\\
          & Ours &
         0.899	 \footnotesize{ $\pm$ 	0.030	} &	0.855	 \footnotesize{ $\pm$ 	0.072	} \\
                 
         \midrule
             2 & Human \cite{LaRosa2023Towards} &
          0.766	 \footnotesize{ $\pm$ 	0.065	} &	0.693	 \footnotesize{ $\pm$ 	0.127	} \\
                 & Closed \cite{bau2020units} & 
        0.743	 \footnotesize{ $\pm$ 	0.057	} &	0.690	 \footnotesize{ $\pm$ 	0.136	} \\
& Ours &
         0.752	 \footnotesize{ $\pm$ 	0.072	} &	0.667	 \footnotesize{ $\pm$ 	0.126	} \\

         \midrule
             3 & Human \cite{LaRosa2023Towards} &
         0.559	 \footnotesize{ $\pm$ 	0.103	} &	0.538	 \footnotesize{ $\pm$ 	0.144	} \\
         & Closed \cite{bau2020units} &
         0.537	 \footnotesize{ $\pm$ 	0.094	} &	0.540	 \footnotesize{ $\pm$ 	0.122	} \\
& Ours &
       0.549	 \footnotesize{ $\pm$ 	0.103	} &	0.522	 \footnotesize{ $\pm$ 	0.128	} \\

         \midrule
             4 & Human \cite{LaRosa2023Towards} &
        0.380	 \footnotesize{ $\pm$ 	0.129	} &	0.411	 \footnotesize{ $\pm$ 	0.186	} \\
                  
                  & Closed \cite{bau2020units} &
         0.342	 \footnotesize{ $\pm$ 	0.112	} &	0.441	 \footnotesize{ $\pm$ 	0.173	} \\
        
        & Ours &
       0.369	 \footnotesize{ $\pm$ 	0.122	} &	0.417	 \footnotesize{ $\pm$ 	0.179	} \\

         \midrule
           5 & Human \cite{LaRosa2023Towards} &
              0.246	 \footnotesize{ $\pm$ 	0.151	} &	0.285	 \footnotesize{ $\pm$ 	0.202	} \\
                 
        & Closed \cite{bau2020units} &
          0.174	 \footnotesize{ $\pm$ 	0.101	} &	0.343	 \footnotesize{ $\pm$ 	0.211	} \\
        
        & Ours &
         0.212	 \footnotesize{ $\pm$ 	0.121	} &	0.311	 \footnotesize{ $\pm$ 	0.200	} \\

         \bottomrule
    \end{tabular}
\end{table}

\section{Clustered Compositional Explanations Algorithm}
\label{appx:mmesh}
Let $\mathbb{A}$ be a binary activation matrix, $\mathbb{C}$ be a set of concepts, $\mathbb{M}$ be a set of binary segmentation masks, one for each concept, and $\mathfrak{L}^n$ be the set of all possible logical connections of arity at maximum $n$ between concepts in the concept set $\mathbb{C}$. These quantities are computed as described in Section 3 of the main text. The goal of compositional explanation algorithms is to find the label $L \in \mathfrak{L}^n$ whose mask maximally overlaps with the neuron binary activations $\mathbb{A}$. Formally, these algorithms find the solution for the following objective: 

\begin{equation}
    \operatorname*{arg\,max}_{L \in \mathfrak{L}^n} IoU(L, \mathbb{A}, \mathbb{M})
\end{equation}
$IoU$ is defined as:
\begin{equation}
    IoU(L, \mathbb{A}, \mathbb{M}) = \frac{|\mathbb{A} \cap \theta(\mathbb{M}, L)|}{|\mathbb{A} \cup \theta(\mathbb{M},L)|}
\end{equation}
and $\theta(\mathbb{M}, L)$ is a function that returns the logical combination of the masks in $\mathbb{M}$ of the concepts involved in the label $L$.

Exhaustive search over $\mathfrak{L}^n$ is computationally infeasible in most of the settings commonly considered in literature. To address this problem, \cite{Mu2020} propose to use beam search in place of exhaustive search. This algorithm has been extended by \cite{LaRosa2023Towards} to speed up the computation of explanation using a beam search guided by the \emph{Min-Max Extension per Sample Heuristic} (\textbf{MMESH}). 

While we refer the reader to \cite{Mu2020} and \cite{LaRosa2023Towards} for full details of the algorithm and its procedures, we briefly outline its main steps and components below.

\begin{algorithm2e}[t]
\caption{\textbf{Beam Search Guided by MMESH}}
\KwIn{$\mathbb{C}$, $\mathbb{M}$, $\mathbb{A}$, \DataSty{MMESHInfo}, \DataSty{b}, \DataSty{length}}
\KwOut{\DataSty{BestLabel},\DataSty{BestIoU}}

$\DataSty{Beam} \gets \text{empty list}$

$\DataSty{UpdatedInfo} \gets \DataSty{MMESHInfo}$

\For{$c_{k,i}$ \textbf{in} $\mathbb{C}$}{

    $\DataSty{Iou} \gets \FuncSty{compute\_iou}(c_{k,i}, \mathbb{M}, \mathbb{A})$
    
    $\DataSty{Beam}\FuncSty{.add}(label=c_{k,i}, iou=\ArgSty{Iou})$
    
}

\FuncSty{sort}(\ArgSty{Beam}) \quad \CommentSty{\# Sort by IoU}

\CommentSty{\# Select the best b candidates}

$\DataSty{Beam} \gets \DataSty{Beam[:b]}$ 

$\DataSty{MinIoU} \gets \FuncSty{find\_min}(\ArgSty{Beam}) $

\For{ 2 \textbf{to} \DataSty{length}}{
    $\DataSty{SearchSpace} \gets \FuncSty{expand\_beam}(\ArgSty{Beam}, \mathbb{C})$
    
    \DataSty{Estimations} $\gets$ \FuncSty{estimate\_iou}(\ArgSty{SearchSpace}, \ArgSty{MMESHInfo})
    
    \FuncSty{sort}(\ArgSty{Estimations})
    
    \For{ $L$, \DataSty{EstIoU} \textbf{in} \DataSty{Estimations}}{
        \If{\DataSty{EstIoU} $<$ \DataSty{MinIoU}}{
        
        \CommentSty{\# All the other labels cannot be added to the beam}

        \textbf{break} 

        }

        $\DataSty{Iou} \gets  \FuncSty{compute\_iou}($L$, \mathbb{M}, \mathbb{A})$
        
        \DataSty{Beam}\FuncSty{.add}((label=$L$, iou=\ArgSty{Iou}))
        
    }

    \FuncSty{sort}(\ArgSty{Beam})

    \CommentSty{\# Select the best b candidates}

    $\DataSty{Beam} \gets \DataSty{Beam[:b]}$

    \CommentSty{\# Compute and update info}

    $\DataSty{MinIoU} \gets \FuncSty{find\_min}(\ArgSty{Beam})$ 
             
    $\DataSty{MMESHInfo} \gets \FuncSty{update\_info}(\ArgSty{MMESHInfo}, \ArgSty{Beam})$
    
}

$\DataSty{BestLabel, BestIoU} \gets \FuncSty{max}(\ArgSty{Beam})$

\Return{BestLabel, BestIoU}
\label{algo:mmesh}

\end{algorithm2e}

The pseudocode is shown in Algorithm \ref{algo:mmesh}. At each step $i$, the algorithm maintains a beam of $b$ candidate explanations, selected based on the highest IoU scores from the previous step. From this beam, it generates a search space by combining the beam labels with the concepts in the concept set $\mathbb{C}$. The combinations are based on the propositional logic operators AND, OR, and AND NOT. For each candidate in this search space, the algorithm estimates the IoU using precomputed heuristic information. The candidates are then sorted based on these estimated scores. At this point, the algorithm computes the IoU for the candidates associated with an estimate IoU greater than the current beam minimum, and the $b$ candidates with the highest IoU are retained as the beam for the next step $i+1$. This process is repeated until the maximum allowed explanation length is reached. Finally, the algorithm returns the explanation that achieved the highest IoU across all steps.

\paragraph{Estimating IoU}
For each sample and each concept, MMESH computes both the bounding boxes and the inscribed rectangles within the concept regions. This geometric information is then combined with concept sizes to estimate the IoU of a given label $L$.

In formulas:
\begin{equation}
\begin{split}
    \widehat{IoU}(L, \mathbb{A},  &\mathbb{M}, \mathfrak{D} )  = \frac{\widehat{I}}{\widehat{U}} =   \frac{\sum_{x \in \mathfrak{D}}\widehat{I^x}}{\sum_{x \in \mathfrak{D}}\widehat{U^x}} =     \\
    & = \frac{\widehat{I_x}}{\sum_{x \in \mathfrak{D}} |\mathbb{A}| + \sum_{x \in \mathfrak{D}}|\widehat{\theta(\mathbb{M}^x, L)}| - \widehat{I_x}}
\end{split} 
\label{eq:iouextended}
\end{equation}

The specific computation of the estimate intersection $\widehat{I}^x$ and the estimated label mask $\widehat{\theta(\mathbb{M}^x, L)}$ depends on the logical operator connecting the left side $(L_{\leftarrow})$ and right side $L_{\rightarrow})$ of the label. In all cases, $\widehat{I}^x$ is an overestimation of the actual intersection and $\widehat{\theta(\mathbb{M}^x, L)}$ is an underestimation of the actual label mask. These conservative estimations ensure that the algorithm finds the optimal solution within the beam. Specifically:

\paragraph{OR}
\begin{equation}
    \widehat{I}^x = min(|IMS(x, L_{\leftarrow})|+ |IMS(x, L_{\rightarrow})|, |M(x)| ) 
\end{equation}
\begin{equation}
\begin{split}
    \widehat{\theta(\mathbb{M}^x, L)} = max(&|\theta(M^x,L_{\leftarrow})|, \\ &|\theta(M^x,L_{\rightarrow})|, \\
    & \widehat{\theta(M^x,L_{\leftarrow} \cup L_{\rightarrow}})) 
\end{split}
\end{equation}
\paragraph{AND}
\begin{equation}
\widehat{I}^x = min(|IMS(x, L_{\leftarrow})|, |IMS(x, L_{\rightarrow})|)  
\end{equation}
\begin{equation}
\widehat{\theta(\mathbb{M}^x, L)} =  max(MinOver(L), I_x)
\end{equation}
\paragraph{AND NOT}
\begin{equation}
\widehat{I}^x = min(|IMS(x, L_{\leftarrow})|, |\mathbb{M}^x| - |IMS(x, L_{\rightarrow})|)
\end{equation}
\begin{equation}
\widehat{\theta(\mathbb{M}^x, L)} = max(|\theta(\mathbb{A}^x, L_{\leftarrow})| - MaxOver(L), I_x)
\end{equation}
where: $\mathbb{M}^x$ and $\mathbb{A}^x$ are defined as in \Cref{appx:eval_metrics}, $IMS(x, L)$ denotes the intersection size between the label mask $\theta(M^x, L)$ and the neuron binary activation  $\mathbb{A}^x $ computed a generic activation range $(\tau_1,\tau_2)$;  $MaxOver(L)$  is a function that returns the maximum possible overlap between the bounding boxes associated with the left and right sides of $L$ in the sample $x$; and $MinOver(L)$ is a function that returns the minimum possible overlap between the inscribed rectangles sassociated with the left and right sides of $L$ in the sample $x$. 

For a complete derivation of these estimations and proofs, we refer the reader to \cite{LaRosa2023Towards}.

\section{Limitations}
\label{appx:limitations}
While, as shown in the previous sections, the framework is flexible and competitive across several settings, we identified several limitations that can serve as a base for future research on both open vocabulary semantic segmentation and explainability. 

\paragraph{Number of Concepts.} The number of concepts that can be tested is constrained by the available memory. Ideally, we would like to evaluate every possible concept in a vocabulary (e.g., the most common 10,000 words in English). However, in practice, the output of segmentation models is a matrix $s_x \in R^{|C_i|, h, w}$, where the first dimension represents the logits (or output probabilities) of all the concepts in the given concept (sub)set.

Although, as explained in Section 3, the first dimension can later be reduced by considering only the maximum value as the model's prediction, this matrix still needs to be loaded into memory, even if only temporarily. Consequently, the maximum number of concepts that can be used in explanations is limited and influenced by the available memory on the workstation and the resolution of the segmentation masks.

\paragraph{Completeness of the Concept Subset} One of the limitations of the current framework is its sensitivity to the completeness of each concept subset. Since the open vocabulary segmentation model is ``forced'' to assign at least one concept to every pixel, the concept subset must be as complete as possible to account for all the possible concepts in the input. When an input element cannot be described by using the concepts in the concept subset, that element leads to hallucinations by the segmentation model. Such hallucinations impact the explanation quality of the wrongly assigned concept, potentially triggering a cascade effect. While this issue can be mitigated by including generic concepts (i.e., ``background'', ``thing'', or ``other'') into the concept subset, their effectiveness depends on the training recipe used to pre-train the backbone models (e.g., whether a background class was included in the training). To address this limitation, future work could explore adaptive mechanisms to filter out unreliable masks, possibly arising from hallucinations, thereby reducing such sensitivity.

\paragraph{Sensitivity to the Concept Subset}
The selection of concepts within a generic concept subset can also affect both the quality of the computed explanations and the performance of the framework itself. While it is desirable to have multiple granularities across different concept subsets, including multiple granularities within a single subset could potentially cause inconsistency in explanations. For example, if both the ``\textit{animal}'' and ``\textit{cat}'' concepts are included in the same subset, the model is forced to choose between them when segmenting a cat, even though both could be considered correct. In these cases, the choice will depend entirely on the biases learned from the training dataset and labels used to train the segmentation model or the multi-modal model. To mitigate this issue, we recommend separating concepts with different levels of granularity into different concept subsets, ensuring that two concepts within the same subset cannot be used to describe the same element. We leave for future research the development of an algorithm that can navigate and mitigate this sensitivity.

\paragraph{Dependence on Prompt Templates} One limitation associated with research in open vocabulary semantic segmentation is its reliance on prompt templates. Most of the analyzed models fine-tune the multi-modal backbone using fixed prompt templates (e.g., ``\textit{a photo of a \{\}}''). These prompts are typically designed for the semantic segmentation task, which focuses on objects and tangible elements (e.g., sky, tree). Once the model has been fine-tuned, the number of templates is fixed, and replacing some of them can lead to out-of-distribution issues. This lack of flexibility reduces the models’ effectiveness in recognizing abstract concepts (e.g., patterns) due to the resulting unnatural descriptions and the impossibility of introducing new prompts. The only mitigation could involve additional fine-tuning of the multi-modal model for the explainability task. We call for further research in this direction to make these models more adaptable during inference and to support greater variability in prompt templates during fine-tuning, especially to account for downstream tasks such as explainability.
\paragraph{Refinements' Cascade Effect} 
While this is not strictly a limitation of the framework, we want to emphasize and make the reader aware of the potential cascade effect when applying refinements to the concept set. As explained in the main text, users can modify the concept set after analyzing explanations to retrieve potentially improved explanations based on the refined set. However, when making such refinements, it is \textbf{important to re-generate the masks} for the subsets where the new concepts are introduced. Indeed, adding a concept to the subset changes the output size of the segmentation model and, consequently, its output distribution. Therefore, this adjustment can alter predictions, particularly the most uncertain ones, for all concepts in the concept subset, even those unrelated to the newly added concepts.
At the explanation level, the experiments reported in the main text did not reveal significant changes in explanations. The only difference we observe is in the selection of concepts used to exclude portions of the dataset (e.g., Cat AND NOT Car). These concepts are used by the compositional explanation algorithm to exclude edge cases of neuron behavior. In this case, multiple choices led to similar outcomes, explaining the differences.
However, we expect that if the newly added concepts substantially improve the coverage of the concept subset or better align its granularity with the recognition capabilities of the backbone model, this could potentially result in more significant shifts in explanations, which should be monitored.

\section{Concept Set for CUB}
\label{appx:conceptset}
This section describes the concept set used for the experiments on the CUB dataset in the main text and discusses alternatives and challenges in the selection process for concept sets.

The concept set was chosen based on the availability of a list of relevant concepts for the task, specifically the categories used in the dataset's annotations. Note that we do not use the annotations themselves; the only relevant information is the list of concepts. This list includes bird species, as well as combinations of colors, shapes, and patterns associated with bird parts. After iterative refinements, the resulting concept set is divided into the following subsets:
\begin{enumerate}
    \item Bird species (e.g., \textit{black footed albatross})
    \item Element colors (e.g., bird colors like  \textit{blue bird} and background colors)
    \item Bird shapes (e.g., \textit{long-legged bird})
    \item Parts (e.g., \textit{bird's wing})
    \item Colored parts (e.g., \textit{blue bird's wing})
    \item Part shape (e.g., \textit{curved bird's bill})
    \item Part patterns (e.g., \textit{solid bird's breast})
\end{enumerate}
These subsets are further divided into three levels of granularity: the first includes bird species, bird shapes, and colors; the second includes parts; and the third includes all remaining subsets. We also include the set of concepts annotated in the Ade20k dataset as an additional subset. This decision allows the detection of neurons that capture individual background elements (e.g., water), potentially exploited by biases in the network, as well as neurons that generally recognize birds (recognized as ``animals'') without specialization. To mitigate hallucinations, we also added the generic concepts ``\textit{background}'' and ''\textit{other}'' to each subset to provide the segmentation model with default choices. Masks generated for these generic concepts are excluded from the explanation generation. The full list of concepts will be released as supplemental material and included in the official repository upon acceptance.

It is worth noting that the specific concept set obtained after iterations of our refinements is not, in general, the optimal one and potentially better sets could be found for specific implementations of the framework. Other than identifying the limitations discussed in \Cref{appx:limitations}, throughout the refinement process, we also observed a \textbf{relation between the specificity of the concepts and the completeness of the concept subset}. In this context, we noted that greater specificity in the concept subset helps the segmentation model to reduce hallucinations when the concept subset is either highly specific or weakly complete (i.e., the set is completed by the concepts ``background'' and ``other'' whose effectiveness depends on the specific backbone model). For example, adding the middle term ``\textit{bird’s}'' to the concept subset of parts empirically improved segmentation masks. A similar effect could be achieved by merging the Ade20k set with single-granularity concept subsets. However, sharing concepts across multiple subsets causes inconsistencies in mask generation (i.e., one can have different masks for the same concept across two different subsets) and, consequently, in the explanation process.
We leave the development of a framework capable of addressing and managing repeated concepts across multiple concept subsets as a direction for future work.

\section{Leveraging WordNet to Analyze the Misalignment}
\label{appx:misalignement}
This section describes the multi-step process we use to analyze the misalignment between explanations computed over human and open vocabulary-segmentation by leveraging the semantic knowledge graph of WordNet~\cite{Miller1995}.
The process consists of the following steps:
\paragraph{Step 1: Mapping the Concept Set to Nodes in WordNet} While this step can be performed manually, we utilize information from Ade20k to implement it in a semi-automatic manner. Specifically, each class in the dataset is associated with a list of synonyms retrieved from WordNet. We leverage this list, when available, to locate the corresponding node in the WordNet graph. Given a concept and its list of synonyms, we select the node whose lemmas have the maximum overlap with the list of synonyms. For concepts without available synonyms, we extract the most common node in WordNet associated with that concept (i.e., the first result returned by a WordNet query). Finally, we manually inspect the generated mappings and refine the associations for the following concepts: \textit{water}, \textit{cushion}, \textit{van}, \textit{plate}, and \textit{radiator}.
\paragraph{Step 2: Extracting the Explanation Differences}
This step focuses on identifying differences between explanations (e.g., produced by two different methods). In the main text, we search for concepts identified by approaches that rely on human-annotated data but are absent in the explanations generated by our framework. This step needs to deal with two tasks: identifying differing concepts and accounting for the logical meaning induced by logical operations. 

When both explanations share the same concept subset, identifying differing concepts is straightforward. However, when the explanations are derived from different concept (sub)sets, we rely on the synonyms of each concept to identify equivalences. In this case, two concepts are considered equivalent if they share at least one synonym. For dealing with the logical meaning, we consider two explanations equivalent if they satisfy logical equivalences (e.g., A OR B is equivalent to B OR A). In computing such equivalences we ignore the negative side of explanations, such as the concept C in the explanation ((A OR B) AND NOT C). Indeed, as explained in \Cref{appx:limitations} (i.e., in the paragraph discussing the cascade effect), explanations can differ in their negative components while still achieving the same overlap with the neuron's activations.
\paragraph{Step 3: Identifying a Meaningful Ancestor} Once a missing concept has been identified in Step 2, we use the mapping generated in Step 1 to identify its corresponding node in the graph. In this step, we search for a meaningful common ancestor by tracing the path of hypernyms from the concept's node to the root of the tree. This is done by examining the hypernym relations between the identified concept and any concept in the explanation retrieved by the alternative method. Although any two nodes in the tree always share at least one common ancestor (i.e., the root node), ancestors located high in the hierarchy are often too abstract to provide meaningful insight. To address this, we consider an ancestor ``found'' only if it is not one of the highest-level nodes in the tree. Specifically, we exclude the following general nodes: 
\textit{equipment},
\textit{substance},
\textit{tracheophyte},
\textit{piece of furniture},
\textit{furnishing},
\textit{barrier},
\textit{art},
\textit{surface},
\textit{vessel},
\textit{container},
\textit{covering},
\textit{device},
\textit{way},
\textit{path},
\textit{craft},
\textit{transport},
\textit{conveyance},
\textit{natural object},
\textit{object},
\textit{attribute},
\textit{form},
\textit{relation},
\textit{impediment},
\textit{structure},
\textit{entity},
\textit{matter},
\textit{creation},
\textit{grouping},
\textit{artefact},
\textit{physical entity},
\textit{whole},
\textit{means},
\textit{abstraction},
\textit{measure},
\textit{being},
\textit{language unit},
\textit{consumer goods},
\textit{durable goods},
\textit{animate thing},
\textit{causal agency},
\textit{part}.

\paragraph{Step 4: Remapping the Concept Set and Generating new Masks}
At the end of Step 3, the process generates a mapping between the missed concepts and their corresponding generalizations identified in the tree. Using this mapping, we revisit all the concepts in the concept subset, map them to their identified generalizations (if applicable), and generate a new concept set based on this updated mapping. Once the new concept set is defined, we generate updated segmentation masks using it.

This iterative process (from Step 2 to Step 4) continues until no further generalization can be identified. In our case, we repeated this process three times, reducing the total number of concepts in Ade20K from 150 to 101. 

\section{Isolating Concept's Impact on Explanations}
\label{appx:isolating}

This section describes the procedure to isolate and evaluate the effect of a concept in both an explanation and the corresponding neuron's activations. Specifically, given an explanation of length $n$ and a concept $c_i$ included in the explanation, we compute: (1) the samples where the full explanation holds, (2) the samples where $c_i$ is present, and (3) the samples where the neuron is active within the considered activation range. Then, we compute the intersection of these three sets and we randomly visualize $m$ samples, highlighting the masks associated with $c_i$. These visualized samples represent instances where the concept is present, the neuron is active, and the concept actively contributes to the explanation. 

To analyze the unexplained portion of the neuron's behavior, we consider sub-explanations 
$SE$ of the original explanation, where $SE$ has a length $s < n$. We then extract the samples where the neuron is active, but the sub-explanation $SE$ does not hold. This set represents the portion of the neuron's activations not explained by the sub-explanation. In our case, we use as a sub-explanation the literals shared between two different approaches. As in the previous case, we randomly visualize $m$ samples from this set, highlighting the masks produced by binarizing the activations within the specified activation range. These two visualizations are compared against each other to identify a potential misalignment between parts of the explanation and the neuron's behavior. To ease the analysis and comparison of the visualizations, we prioritize samples associated with larger masks when selecting the samples for visualization.
\section{User Study Details}
\label{appx:userstudy}

\begin{table}

                         \caption{Std. Dev for Alignment, Precision, and Relevance scores attributed by 100 participants to explanations computed by all the competitors. The superscript\textsuperscript{*} indicates that the results are computed on a different probing dataset.}
    \label{table:std_devuserstudy}
    \centering

    \begin{tabular}[t]{lccc}
        \toprule
         Scores & Align & Prec & Relev \\
         \midrule
          \multicolumn{4}{c}{Places365 Probed Model} \\
           \midrule
        Human \cite{LaRosa2023Towards} & 1.12 & 1.22 & 1.22\\
        Closed \cite{bau2020units} & 1.42 & 1.41 & 1.24 \\
        Our & 1.05 & 1.27 & 1.30\\
         \midrule
          \multicolumn{4}{c}{CUB Probed Model} \\
          \midrule
        Human \cite{LaRosa2023Towards} & ~1.31\textsuperscript{*}  & ~1.39\textsuperscript{*} & ~0.72\textsuperscript{*} \\
        Closed \cite{bau2020units} & 1.22 & 1.47 & 1.09\\
        Our & 1.37 & 1.40 & 0.92\\
         \bottomrule
    \end{tabular}   
\end{table}
As described in the main paper, we conducted a user study to qualitatively assess the performance of our framework. Designing such a study presents several challenges. Indeed, neuron-level explanations are intended for researchers or developers involved in building or analyzing the model. Therefore, the complexity of the logical formulas and the need for a deep understanding of the model’s training task represent critical factors and current limitations of this type of explanation. These characteristics constrain the pool of suitable participants and necessitate careful consideration when designing the study instructions.

\paragraph{Setup}
To recruit participants, we used the Prolific platform\footnote{\url{https://www.prolific.com/}}. Eligibility criteria required participants to be AI taskers (i.e., a special group of Prolific participants with proven skills in completing AI evaluation and training tasks\footnote{https://participant-help.prolific.com/en/article/5baf0c}). Additionally, participants were required to have completed over 100 prior submissions with an approval rate above 90\%, and not be affected by color vision deficiency. The former requirements are common in the platform and ensure familiarity with the platform itself. The latter requirement was necessary due to the use of color-based concepts in the explanations and the importance of color features for distinguishing bird species in the CUB dataset. The survey is hosted and has been created using the Qualtrics platform\footnote{https://www.qualtrics.com/}.

The median completion time for the survey was $\sim$30 minutes and all participants were compensated at a rate above the minimum wage in the country of the data collector (full details will be disclosed upon acceptance to preserve anonymity). We recruited a total of 100 participants. Three responses were excluded from the analysis due to their completion times (less than 10 minutes), which were significantly shorter than the median and indicated potential low-quality evaluation.

The full question pool consisted of 120 questions (i.e., 60 questions per model, 20 questions per score and 20 per method). Each model question refers to a different neuron; thus, we consider explanations associated with 60 different neurons. Neurons have not been cherry-picked and they correspond to the neurons at the indices 0-60 for their respective models. All the explanations are associated with the highest cluster and competitors do not share neurons. Each participant was presented with a randomized subset of 30 questions, comprising 10 questions for each score. Due to this randomization, the 10 questions assigned per score could include questions related to one, two, or all competitor methods (see the ``Alternative Design Choices'' paragraph below for a discussion of this design).

\paragraph{Scores} Participants were asked to rate how many concepts in the explanations generated by each method were aligned, precise, and relevant on a scale from 1 (none) to 5 (all). Given a randomly sampled set of activation masks produced by a neuron within a specific activation range, a concept is considered aligned if it appears in at least a subset of the activated masks; precise if its level of granularity matches that of the concepts included in the activation masks; and relevant if it is perceived as discriminative for the given task. Note that, since the visualization is extracted \textbf{independently} from the specific explanation, masks can be noisy and include more or different concepts from the ones included in the explanations due to superposition~\cite{elhage2022superposition,o2023disentangling}. The full set of instructions can be found in the supplemental material. Here, we briefly report the definition of the scores given to the participants:
\begin{itemize}
    \item The \textbf{alignment score} measures the alignment between the label and what is shown in the unmasked regions of a collection of images. A high alignment score means that most (or all) of the concepts included in the label are also present in the unmasked regions of the images. A low alignment score indicates that few (or none) of the label's concepts appear in the unmasked regions of the images.
    \item The \textbf{precision score} measures the difference between the granularity of concepts included in a label and the granularity of the same concepts visualized in the unmasked regions of a set of images. A high precision score indicates that the label accurately reflects the level of detail (granularity) shown in the unmasked regions of the images. A low precision score means that the label is either more general or more specific compared to what is shown. 
    \item The \textbf{relevance score} is used to rate how relevant a concept is for the task (i.e., how informative it is about what the model has learned). A concept is highly relevant if it is discriminative for the task (i.e., it provides useful information to distinguish between different classes or categories of objects). It is correlated to the task if it may frequently appear in the data related to the task but does not help differentiate between classes. A concept is considered low in relevance if it is neither discriminative nor highly correlated. 

\end{itemize}

\begin{figure}[t]
    \centering
    \begin{subfloat}{
\includegraphics[width=0.9\linewidth]{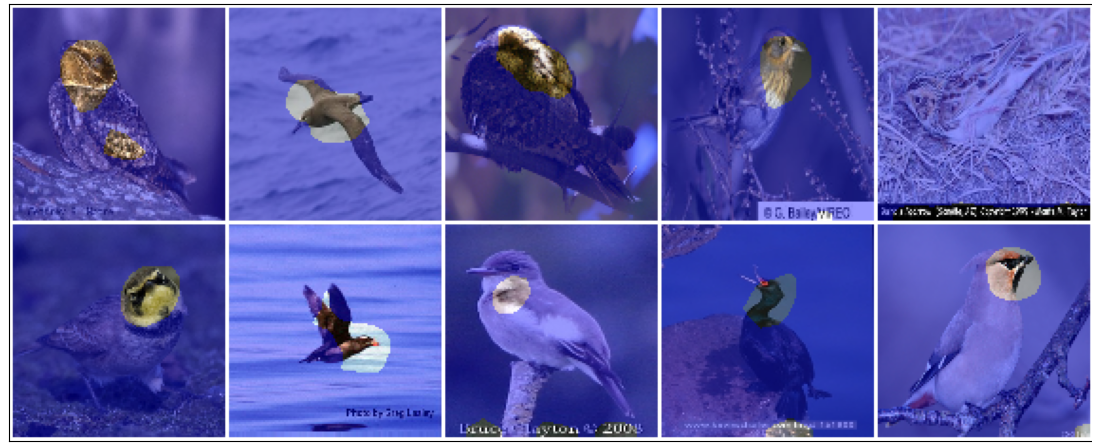}}
    \end{subfloat}
    \begin{subfloat}{
\includegraphics[width=0.9\linewidth]{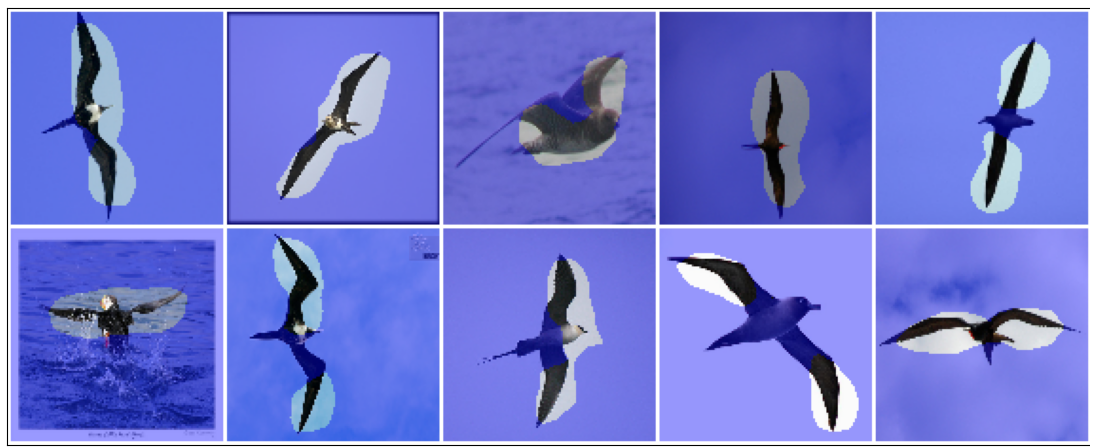}}
    \end{subfloat}
        
    \caption{Examples of highly scored ``bird'' explanations for precision despite most of the images refer to bird parts (head (top) and wings (bottom)). Participants were asked to evaluate only the \textbf{unmasked regions} of the images. }
    \label{fig:userprecision}
\end{figure}
\paragraph{Results}
The average scores are reported in the main paper, while the standard deviations are provided in \Cref{table:std_devuserstudy}. As discussed in the main paper, the user study confirms that our framework performs consistently across both datasets. In this section, we provide a more detailed analysis of these results.

We begin with the model trained on the Places365 dataset. As expected, the explanations generated by both \cite{LaRosa2023Towards} and our framework achieve comparable scores, with no statistically significant difference between them. This similarity is reasonable given that both methods use ADE20K as a concept set, which contains concepts that are known to closely align with the semantic space learned by the Places365 model~\cite{Bau2017} and the differences in concept granularity affect only a small subset of the explanations, as discussed in Section 4.3.
In contrast, the Closed approach~\cite{bau2020units} receives the lowest granularity score. While some concepts are shared between COCO (i.e., the dataset used to train the segmentation model underlying the Closed approach) and Places365, many relevant concepts are either missing or represented at a different level of granularity, resulting in being considered either too broad or too specific. The statistical significance of this difference is supported by P-values $< 0.001$, obtained using a two-tailed t-test comparing the Closed approach's precision scores to those of both \cite{LaRosa2023Towards} and our framework.

For the model trained on the CUB dataset, \cite{LaRosa2023Towards} probes the model using the Ade20K dataset. While its precision score remains comparable to that observed when applied to the Places365 model, its alignment score significantly drops (P-value $< 0.01$ using a two-tailed t-test), likely due to noise introduced by the model's hallucinations over this dataset. We hypothesize that this misalignment could become even worse when the probing dataset differs substantially in terms of visual features from those used to pre-train and fine-tune the probed model. Moreover, this approach achieves the lowest score in terms of relevance (P-value $< 0.01$, two-tailed t-test, compared to our framework’s score for the same model). This drop is related to the fact that, in the vast majority of explanations, the selected concepts are not semantically related to the task of bird species recognition and thus are scored low by users. 

Regarding the Closed approach, it received unexpectedly similar precision scores to our framework. This result is due to the inclusion of the concept ``bird'' in every explanation, which aligns with the fact that all images in the CUB dataset show birds. Since this behavior represents a degenerate case, where a generic concept is trivially included in all explanations, it should ideally be penalized in a meaningful evaluation. After analyzing the participants' responses, we hypothesize two main reasons for this outcome. First, at the individual instance level, it is difficult for inexperienced users to penalize the use of a concept like ``bird'' in a bird dataset, even when activation masks highlight only parts of the bird (\Cref{fig:userprecision}). This is especially true when, due to the study design and the randomness of the sampling process, users are never exposed to explanations associated with a more fine-grained granularity (i.e., they receive questions related to only the competitors' approaches). Second, Closed explanations often contain only one concept (``bird'' in this case). This creates a perceived scenario in which users must decide between two extremes: either the explanation perfectly fits or does not fit at all with the granularity shown within the images. In such cases, participants may be reluctant to assign the lowest score to a concept that is slightly more general than the visualization.

In conclusion, our framework represents the preferred approach overall. Indeed, when applied to explain the CUB model, its explanations are ranked as the most relevant explanations (P-value $< 0.001$ using a two-tailed t-test with respect to both the baselines) by a significant margin while also achieving high scores in both alignment and precision and avoiding the degenerate behaviors observed in the other two approaches. When applied to the Places365 model, it is ranked comparably to the best approach (human) in all the scores.

\paragraph{Alternative Design Choices and Limitations of User Studies} The instructions and the user study design we used in this paper are the product of several iterations aimed at reducing bias and improving the evaluation quality. Specifically, the resulting design is the one that penalizes the competitors the least. Below, we briefly discuss some alternative designs discarded because the resulting user study would have been too hard to understand for an inexperienced user, would have biased the evaluation, or would have penalized a competitor too much.
\begin{itemize}
    \item \textbf{Let the participants rank different explanations for the same neuron.} One of the first designs we considered was to ask users to rank explanations produced by different methods for the same neurons. This approach would have allowed us to directly identify which explanation is preferred on average. However, this design would have unfairly penalized \cite{LaRosa2023Towards} in the questions related to the CUB model. In that case,  \cite{LaRosa2023Towards} generates explanations based on a different dataset (Ade20K). As a result, if we had shown random activation masks from the CUB dataset (on which the model is trained), most, if not all, of the explanations from the human-based approach would appear misaligned, imprecise, and irrelevant, as they were computed using different concepts and a different dataset. We considered the alternative of showing two sets of images per question, one for the dataset used to generate the explanations and one for the dataset used to train the model, but this would likely have confused participants, as the two sets of images would refer to entirely different concepts and the survey setup would have been different between CUB and Place365 models. To resolve this issue, we adopted a design in which participants evaluate each explanation independently, without seeing competing explanations. While this approach prevents us from directly extracting rankings, we can still extract insights through indirect comparisons. More importantly, this design keeps the structure of the survey consistent and easier for participants to follow and understand.
    \item \textbf{Let the participants score the grade of alignment/precision/relevance instead of the number of concepts.}   
    An alternative design is to ask participants to assign a numeric score to each explanation, reflecting their perception of its overall alignment, precision, or relevance. While this setup could be more suitable for expert participants (see discussion below regarding participant pools), we believe it is not appropriate for non-researcher participants. One of the main challenges with this design is achieving a consistent interpretation of the scoring scale across participants. Although the instructions could include example ratings, for non-experts the required level of detail would likely be so extensive that it could bias the evaluation process and compromise the statistical significance of the results. Instead, we ask participants to rank the number of concepts they perceive as aligned, precise, or relevant in each explanation. This approach simplifies the task for non-expert users and avoids the need for detailed examples or guidelines that might influence their judgments. The trade-off of this design, however, is that methods producing shorter explanations, particularly those with only one concept (i.e., the Closed approach applied to the CUB model), may gain an unintended advantage. In such cases, participants are often forced to choose between the two extremes of the ranking scale (either all or none of the concepts are aligned, precise, or relevant) and we observed a tendency to favor the positive extreme in these situations, as we discussed in the previous paragraph.

    \item \textbf{Different level of details for instructions.}
    We iterated several times on the level of detail provided in the instructions and tested them with different types of users. While there is no one-size-fits-all solution, the current version of the instructions is perceived differently by different users. Based on early feedback, we found that some researchers working in the same area as this paper might consider the instructions overly detailed or guided. However, given the very limited number of experts worldwide in this specific field, the likelihood of such users being recruited through a crowdsourcing platform is extremely low and can be considered negligible. In contrast, most participants are individuals with some familiarity with the AI domain, but who likely lack deep knowledge of explainability or the specific tasks discussed in this work. According to some participants’ feedback, they would have preferred even more detailed instructions and a more guided process, as they often struggled to evaluate the explanations due to several challenges (e.g., image and mask noise and resolution, edge cases). Regardless, we intentionally chose not to provide additional guidance to avoid introducing bias into the evaluation process. We believe that the ``uncertainty'' experienced by some users is an intentional and even desirable aspect, as there are no definitive right or wrong answers (i.e., ground truth) in the context of these types of explanations.    
    \item \textbf{Different participants pool.} Given the expertise required to understand logical formulas and the deep familiarity with the underlying tasks needed to evaluate such explanations, one possible option would have been to select participants for the survey exclusively based on these two criteria. However, this approach would have resulted in a very limited participant pool, making it difficult to obtain statistically significant results. Moreover, identifying and recruiting such participants would have required considerable time and effort, effectively ruling out the use of crowdsourcing platforms. For the same reasons, enlarging the participant pool would have meant reducing the quality of the evaluations, as many potential participants might lack knowledge of what constitutes an AI task or even a basic understanding of AI itself. This would increase both the time required to comprehend the instructions and the survey, and introduce noise into the user study, making it more challenging to extract meaningful insights.
\end{itemize}

In conclusion, given the challenges described in this section regarding the design of user studies for this type of explanation, we argue that quantitative metrics, such as those used in Section 4, should remain the main tool for evaluating these methods. However, user studies can still provide insights into aspects that are difficult to capture quantitatively (e.g., relevance). As we have discussed, designing unbiased and fair surveys for these neuron explanations, without compromising the evaluation quality or statistical significance, presents several challenges. We therefore call for further research to lower the expertise needed to interpret logical explanations and to address the need for deep domain knowledge of the training dataset and tasks to evaluate them. These limitations currently restrict the usefulness of these explanations to researchers or developers who are directly involved in training the models to be explained.

\section{Broader Impact Statement}
\label{appx:broader}
The opacity of the learning process in deep neural networks remains a major barrier to their adoption in domains where understanding the rationale behind model decisions is essential for trust and accountability. In this paper, we address one of the limitations highlighted in the broader impact statement of \cite{Mu2020}, namely the reliance on annotated datasets, which \emph{``may be expensive to collect and may be biased in the kinds of features they contain (or omit)''}~\cite{Mu2020}. We argue that the explanations generated by our framework can positively contribute to the broader impact of explainability methods by expanding the range of use cases and potential users.

Although the contributions of this work are experimental and not deployed in downstream applications, we recognize potential sources of negative societal impact if the explanation process is not properly verified or is maliciously manipulated. Specifically, incorporating pre-trained open-vocabulary segmentation models into the explanation pipeline may introduce biases embedded in the segmentation process. However, detecting and mitigating such bias is as challenging in model-generated segmentations as it is in human-annotated datasets.

A more concrete vulnerability lies in the segmentation masks themselves: an adversarial actor could subtly alter the output of the segmentation model in ways that are not immediately noticeable to users but significantly distort the resulting explanations. Furthermore, as discussed in \Cref{appx:userstudy}, this work does not address the challenge related to the technical expertise required to implement and interpret these explanations. Both these limitations can be mitigated in future research exploring adversarial settings and improving the usability of compositional explanations.

\section{Reproducibility}
\label{appx:reproducibility}
To ensure full reproducibility, we will release the complete codebase and all scripts required to reproduce the results presented in this paper upon acceptance. In the meantime, this section serves as a brief summary and documentation of the experimental setup used by our framework, along with the resources required.
\subsection{Dataset, Models, and Explanations}
In this section, we provide the repository, dataset, explanations, and model information, versions, their corresponding licenses, download links, and a brief description of the modifications required to ensure compatibility with our framework.
\paragraph{Datasets}
\begin{itemize}
    \item Mapillary Vistas~\cite{Neuhold2017} v. 1.2
    \begin{itemize}
        \item Accessible at: \url{https://www.mapillary.com/dataset/vistas}
        \item License: CC BY-NC-SA and subject to Mapillary Terms of Use\footnote{https://www.mapillary.com/terms}
    \end{itemize}
    \item Cityscapes~\cite{Cordts2016Cityscapes}  
    \begin{itemize}
        \item Accessible at: \url{https://www.cityscapes-dataset.com/}
        \item License: MIT license and custom terms of use\footnote{https://www.cityscapes-dataset.com/license/}
    \end{itemize}  
    \item Pascal VOC~\cite{pascal-voc-2012}
        \begin{itemize}
        \item Accessible at: \url{http://host.robots.ox.ac.uk/pascal/VOC/}
        \item License: flickr terms of use\footnote{\label{filkr}https://www.flickr.com/help/terms}
    \end{itemize}
     \item PASCAL-Context-459~\cite{mottaghi_cvpr14}
         \begin{itemize}
        \item Accessible at: \url{https://cs.stanford.edu/~roozbeh/pascal-context/}
        \item License:  flickr terms of use\footref{filkr}
    \end{itemize}
      \item Ade20k~\cite{Zhou2017} 
               \begin{itemize}
        \item Accessible at: \url{https://ade20k.csail.mit.edu/}
        \item License: MIT
    \end{itemize}
       \item COCO-Stuff~\cite{Caesar_2018_CVPR}
        \begin{itemize}
        \item Accessible at: \url{https://cocodataset.org/}
        \item License: CC-BY 4.0 and flickr terms of use\footref{filkr}
    \end{itemize}
           \item CUB~\cite{WahCUB_200_2011}
        \begin{itemize}
        \item Accessible at: \url{https://www.vision.caltech.edu/datasets/cub_200_2011/}
        \item License: CCO
    \end{itemize}
    
\end{itemize}

To make the datasets compatible with Detectron2~\cite{wu2019detectron2}, we follow the instructions reported in the following repositories:
\begin{itemize}
    \item \url{https://github.com/cvlab-kaist/CAT-Seg/tree/main} for Ade20k (150 classes and its extended version), Pascal VOC, Pascal-Context, and COCO-Stuff
    \item \url{https://github.com/facebookresearch/MaskFormer/tree/main} for Cityscapes and Mapillary Vistas
\end{itemize}
\paragraph{Models}

\begin{itemize}
    \item CAT-Seg~\cite{Cho2024CatSeg}
    \begin{itemize}
        \item Accessible at: \url{https://github.com/cvlab-kaist/CAT-Seg}
        \item License: MIT
        \item Version: Large (L)
    \end{itemize}
        \item MasQCLIP~\cite{xu2023masqclip}
    \begin{itemize}
        \item Accessible at: \url{https://github.com/mlpc-ucsd/MasQCLIP}
        \item License: CC BY-NC 4.0
        \item Version: Cross-Dataset
    \end{itemize}
        \item SCAN~\cite{Liu_2024_CVPR}
    \begin{itemize}
        \item Accessible at: \url{https://github.com/yongliu20/SCAN}
        \item License: CC BY-NC 4.0
        \item Version: SCAN-VitL
    \end{itemize}
        \item SED~\cite{Xie_2024_CVPR}
    \begin{itemize}
        \item Accessible at: \url{https://github.com/xb534/SED}
        \item License: Apache 2.0
        \item Version: SED (L)
    \end{itemize}
        \item OpenSeed~\cite{Zhang2023openseed}
    \begin{itemize}
        \item Accessible at: \url{https://github.com/IDEA-Research/OpenSeeD}
        \item License:  Apache 2.0
        \item Version: COCO o365 SwinT
    \end{itemize}
            \item Mask2former~\cite{Zhang2023openseed}
    \begin{itemize}
        \item Accessible at: \url{https://github.com/facebookresearch/Mask2Former}
        \item License:  CC BY-NC 4.0
        \item Version: COCO Panoptic SwinT
    \end{itemize}
\end{itemize}
We slightly modified the implementation of all these models to provide a unified interface compatible with the capabilities of our framework. Importantly, these modifications do not affect the pre-trained weights and do not require retraining the segmentation models. Specifically, we extended the models with an interface that allows arbitrary concepts to be added, removed, or specified on the fly. This replaces the default interface, which relies on dataset-specific classes supported by Detectron2. When necessary, we preserve model-specific dataset customizations by loading concepts from JSON files provided by the original authors. For all the models, we use the default parameters suggested and tested by the original authors.
\paragraph{Explanations} 
Our framework generates explanations through a heuristic search guided by the MMESH heuristic~\cite{LaRosa2023Towards}. The implementation of the heuristic is based on the one provided by the original authors, available at \url{https://github.com/KRLGroup/Clustered-Compositional-Explanations}, while the search procedure is inspired by the compositional explanation repository at \url{https://github.com/jayelm/compexp}. In our experiments, we fix the number of clusters to 5 and set the explanation length to 3, following the setup proposed in \cite{Mu2020}. The beam branching factor is also set to 5. For building the logical forms of explanations, we employ the AND, OR, and AND NOT operators, as specified in \cite{Mu2020}.
\paragraph{Repository} As specified in \Cref{appx:eval_ovseg}, the technical settings considered in this paper and the ones needed \textbf{to replicate its results} include the following libraries: PyTorch 1.3~\cite{torch}, Detectron2~\cite{wu2019detectron2}, MMEngine 1.6.2~\cite{mmcv}, and MMSegmentation 0.27.0~\cite{mmseg2020}).
Note, however, that our framework generally supports custom datasets and models. The core implementation is compatible with any PyTorch version $> 1.3$ and does not rely on functionalities specific to MMEngine or MMSegmentation. The only \textbf{general requirement} is that the open vocabulary segmentation model must be adapted to the common interface expected by our framework for parsing datasets and that the dataset loading function is made compatible with our implementation. A complete guide on how to integrate custom models and datasets will be provided in the repository upon acceptance.

\subsection{Resources}
\label{appx:resources}
The computational resources required by our framework are determined by the choice of open vocabulary backbone and the configuration used for compositional explanations. In this context, our framework does not need additional resources beyond the ones required by the individual segmentation models and the compositional explanation process. However, it does increase the time needed to compute concept masks when generating masks for multiple concept subsets. Indeed, in these cases, the framework requires parsing the dataset multiple times. For instance, in the case of CUB, the framework employs 8 concept subsets, resulting in 8 times the time needed by the Closed approach, which parses the dataset only once. However, note that the time required to compute concept masks is generally much lower than the time needed to compute explanations for a layer, particularly when dealing with wide layers.

All experiments were conducted on a workstation equipped with an NVIDIA RTX 3090 GPU, 8 CPU cores, and 64 GB of RAM. The runtimes reported below are based on this setup. However, the implementation of the framework supports lighter configurations at the expense of increased computational time. In this case, the minimum requirements are either 12 GB of VRAM and RAM (for GPU-based execution) or 24 GB of RAM (for CPU-only execution).

The phases of our framework can be identified as: the generation of masks and the generation of explanations.

The time required to generate and store the masks in Compressed Row Format depends on (i) the selected segmentation model, (ii) the dataset, and (iii) the number of concept subsets used in our framework. As a rough estimate, for all the datasets but CUB, \cite{LaRosa2023Towards} takes $\sim$4–10 minutes to process and convert the segmentation masks into the compressed format; the closed approach~\cite{bau2020units} takes $\sim$6–15 minutes; and the time required by our framework depends on the backbone segmentation model. Among the tested open vocabulary backbones, MasqCLIP, OpenSeed, and SED are the fastest, taking $\sim$8-12 minutes; Cat-Seg takes $\sim$15-20 minutes; and SCAN requires $\sim$25-30 minutes. When processing the CUB dataset, the computation time increases due to its larger size and higher image resolution. In this case, the closed approach takes $\sim$12 minutes, while for our framework implementations Cat-Seg takes $\sim$25 minutes per concept subset, resulting in a total of $\sim$5 hours; MasqCLIP, OpenSeed, and SED take between 7 and 10 minutes per set, totaling $\sim$1 hour; SCAN takes $\sim$1 hour per set, resulting in a total of $\sim$8 hours.

The memory requirements of our framework depend on (i) the selected model, (ii) the total number of concepts, and (iii) the number of concepts within each concept subset. Regarding the latter, as discussed in \Cref{appx:limitations}, the number of concepts in a concept subset impacts the size of the output generated by the open vocabulary segmentation models and, consequently, the memory required to store these outputs, even temporarily. In practice, different implementations require between 8 and 16 GB of RAM or VRAM to store the segmentation masks in memory.

Finally, regarding the time needed to compute the explanations, it depends on the number of concepts that overlap with the considered activation range, due to the heuristic employed by our framework. Including a larger and more relevant set of concepts generally increases the number of overlapping concepts, slightly raising the computational time per neuron. On average, all competitor methods take less than 2 minutes per neuron for all models except the CUB model. For CUB, explanation computation is approximately twice as slow, requiring 4–5 minutes per neuron. Additionally, the total computation time depends on the number of neurons analyzed. For instance, the ResNet18 Places365 model contains 512 neurons in the last layer, while the CUB model contains 2048 neurons. As a result, analyzing the full layer in the CUB model takes approximately four times longer than in the Places365 model.

These per-neuron timings can be used to estimate the total time needed to replicate the experiments reported in the paper. For example, assuming the workstation described in this section, reproducing the results in \Cref{tab:models_ade20k} would take approximately 2–3 days, while reproducing the experiments in \Cref{tab:models_cub} would take about 8 days per open vocabulary segmentation model. Note that these runtimes can be significantly reduced by running the code on GPU clusters and parallelizing the analysis of models or neurons.

\section{Additional Preliminary Experiments}
\label{appx:preliminary}
Before conducting the full set of experiments reported in the paper, we performed preliminary tests to evaluate different configurations of the segmentation models. Since the primary goal of this work is not to identify the optimal backbone for our framework, we did not explore this direction further. However, these initial findings may serve as a useful starting point for future users or researchers.
\begin{itemize}
    \item We do not observe a significant difference in the generated explanations when using the standard dataset classes as concepts compared to the customizations provided by the original authors for open vocabulary segmentation models. However, we adopted the original customizations to ensure fairness and consistency with the authors’ intended use. While we hypothesize that these customizations might have a minor effect on specific or rare segmentation masks, their impact appears to be uniformly distributed across concepts and, therefore, does not meaningfully affect the resulting explanations;
    \item We conducted preliminary experiments aimed at improving the template prompts used to generate textual descriptions more tailored to explainability purposes. However, we observed that modifying the templates employed by open-vocabulary segmentation models affects negatively both the resulting segmentation masks and the explanations. Since these models are fine-tuned using specific templates, we hypothesize that even slight changes can lead to out-of-distribution behavior, resulting in potentially unreliable outputs. As noted in \Cref{appx:limitations}, it is currently not possible to increase the number of templates in a trained model. We leave the investigation of this limitation to future work on open-vocabulary segmentation.
    
\end{itemize}

\section{Visual Comparison between Closed and Open Vocabulary Explanations}
\label{appx:visualcomparison}
\begin{figure*}
    \centering
        \begin{subfloat}{
\includegraphics[width=0.7\linewidth]{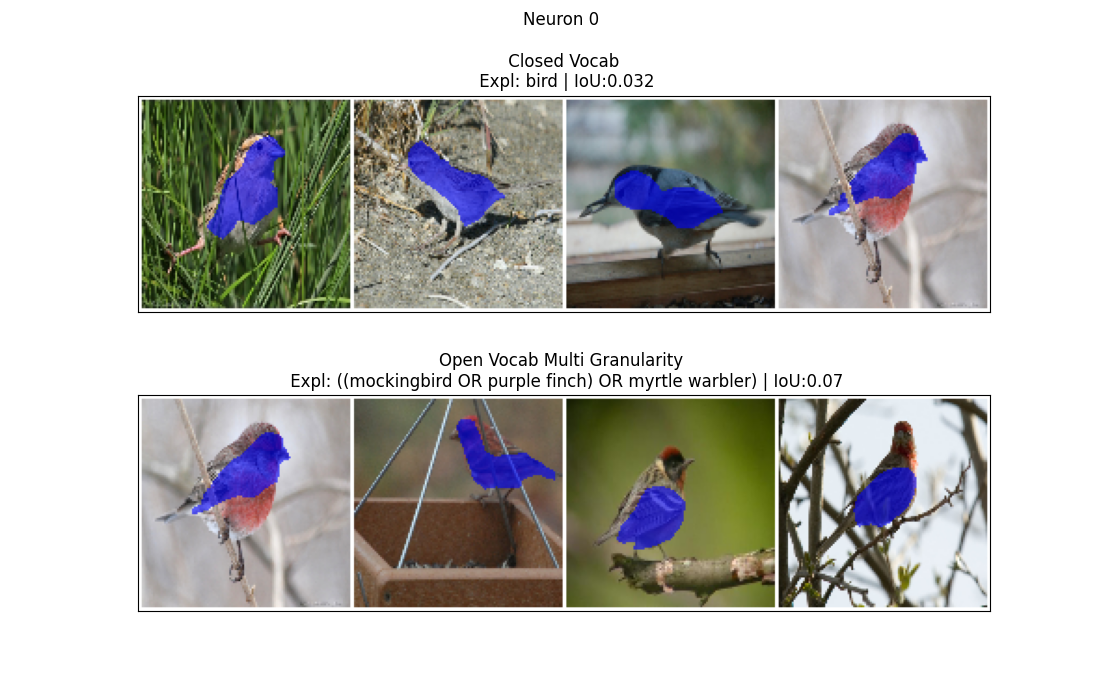}}
    \end{subfloat}
    \begin{subfloat}{
\includegraphics[width=0.7\linewidth]{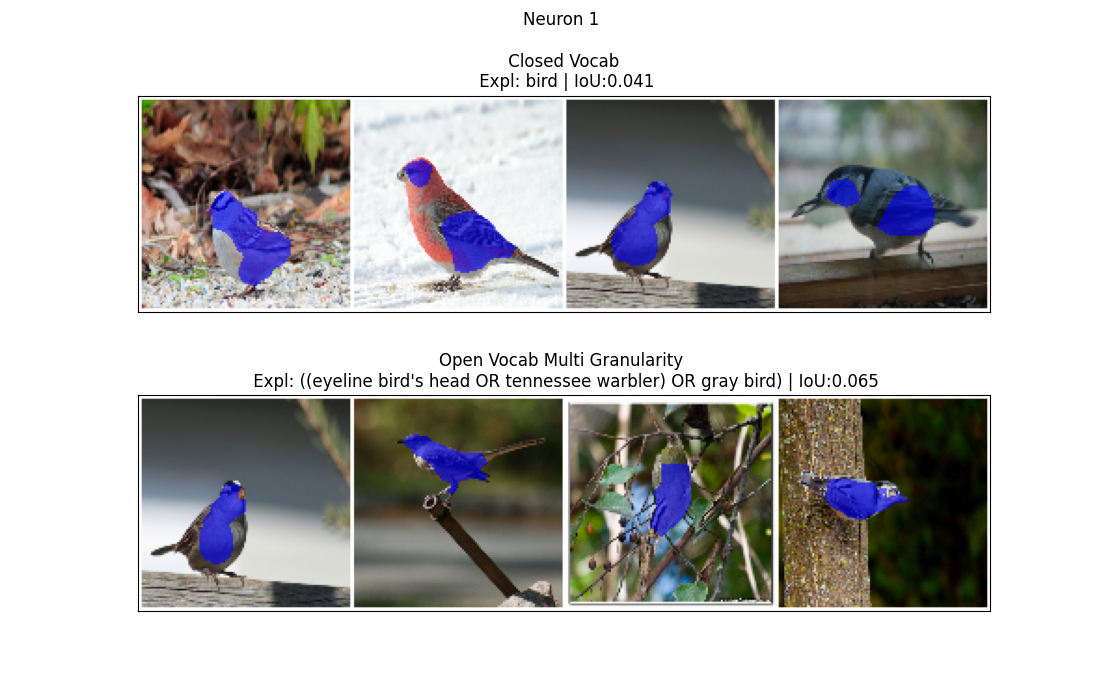}}
    \end{subfloat}
    \begin{subfloat}{
\includegraphics[width=0.7\linewidth]{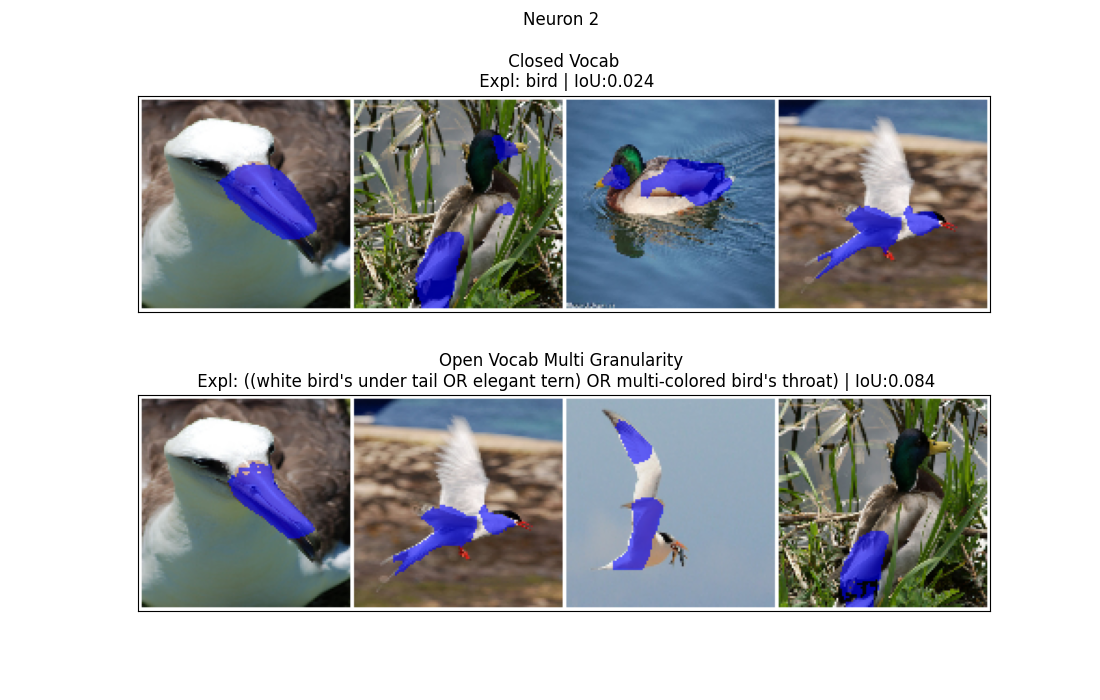}}
    \end{subfloat}
        
    \caption{Explanations associated with Cluster 5 of neurons from 0 to 2 by the \textit{Closed} approach~\cite{bau2020units} and our framework. In blue are areas of neuron activation within the considered range.}
    \label{fig:neuron0-2}
\end{figure*}

\begin{figure*}
    \centering
    \begin{subfloat}{
\includegraphics[width=0.7\linewidth]{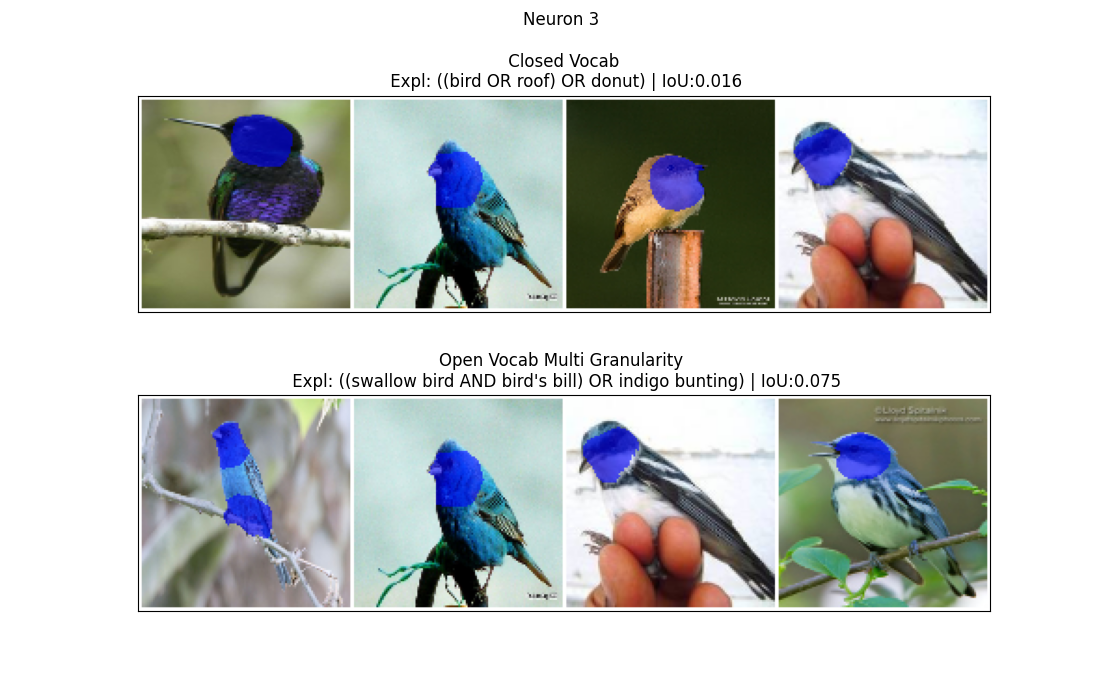}}
    \end{subfloat}
            \begin{subfloat}{
\includegraphics[width=0.7\linewidth]{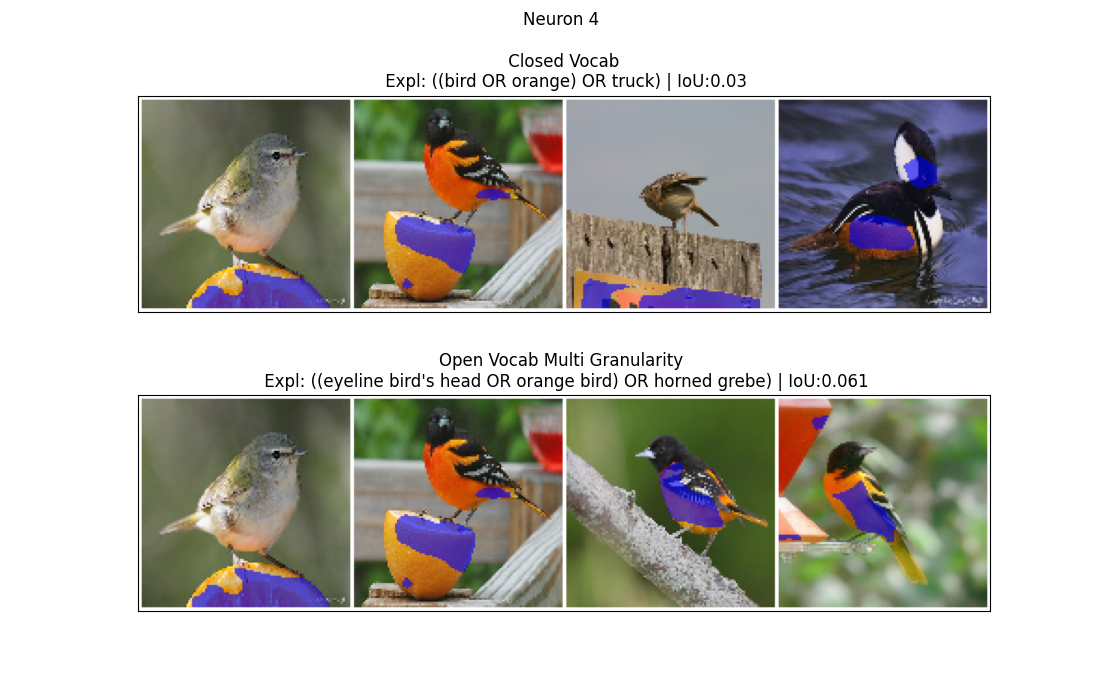}}
    \end{subfloat}
        \begin{subfloat}{
\includegraphics[width=0.7\linewidth]{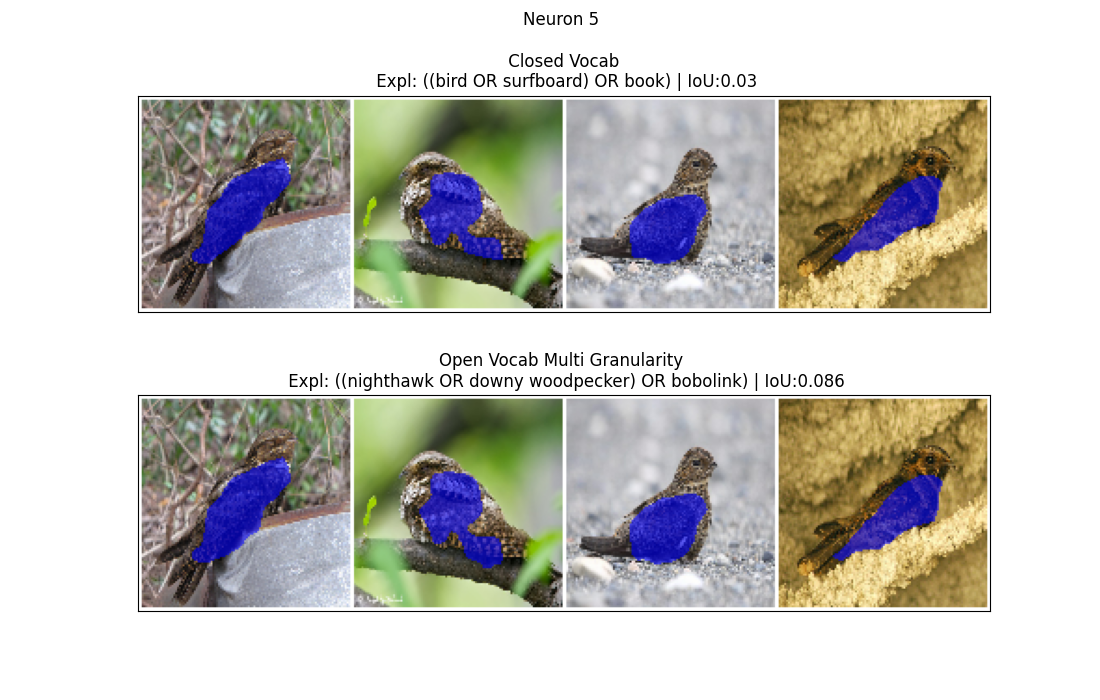}}
    \end{subfloat}

    \caption{Explanations associated with Cluster 5 of neurons from 3 to 5 by the \textit{Closed} approach~\cite{bau2020units} and our framework. In blue are areas of neuron activation within the considered range.}
    \label{fig:neuron3-5}
\end{figure*}

\begin{figure*}
    \centering
            \begin{subfloat}{
\includegraphics[width=0.7\linewidth]{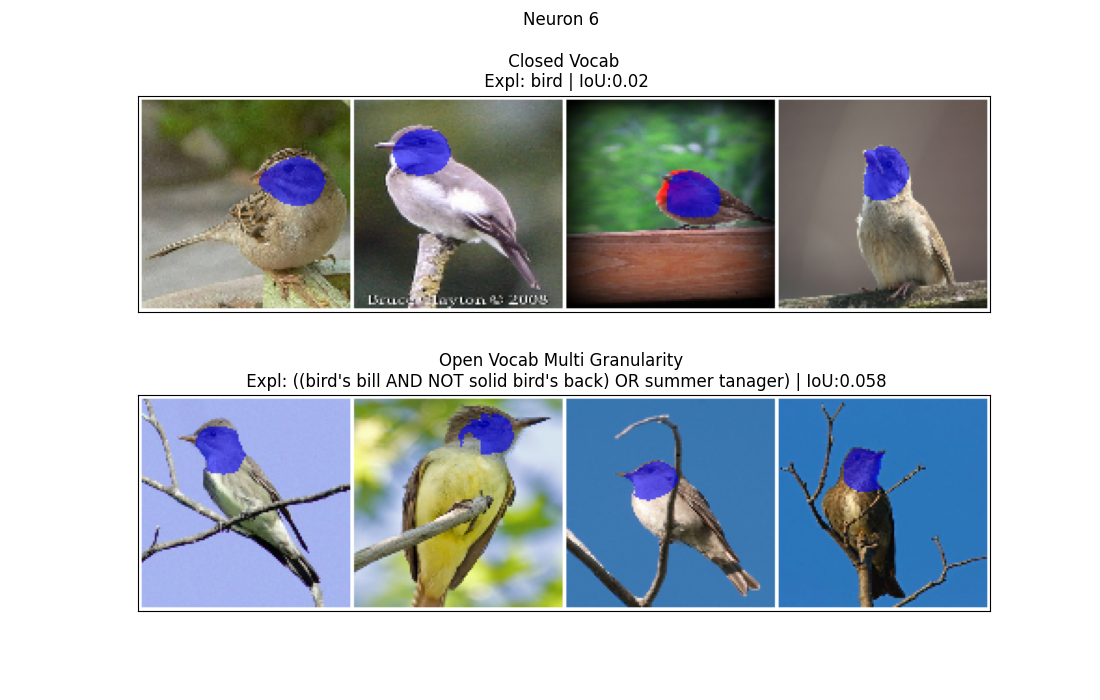}}
    \end{subfloat}
    \begin{subfloat}{
\includegraphics[width=0.7\linewidth]{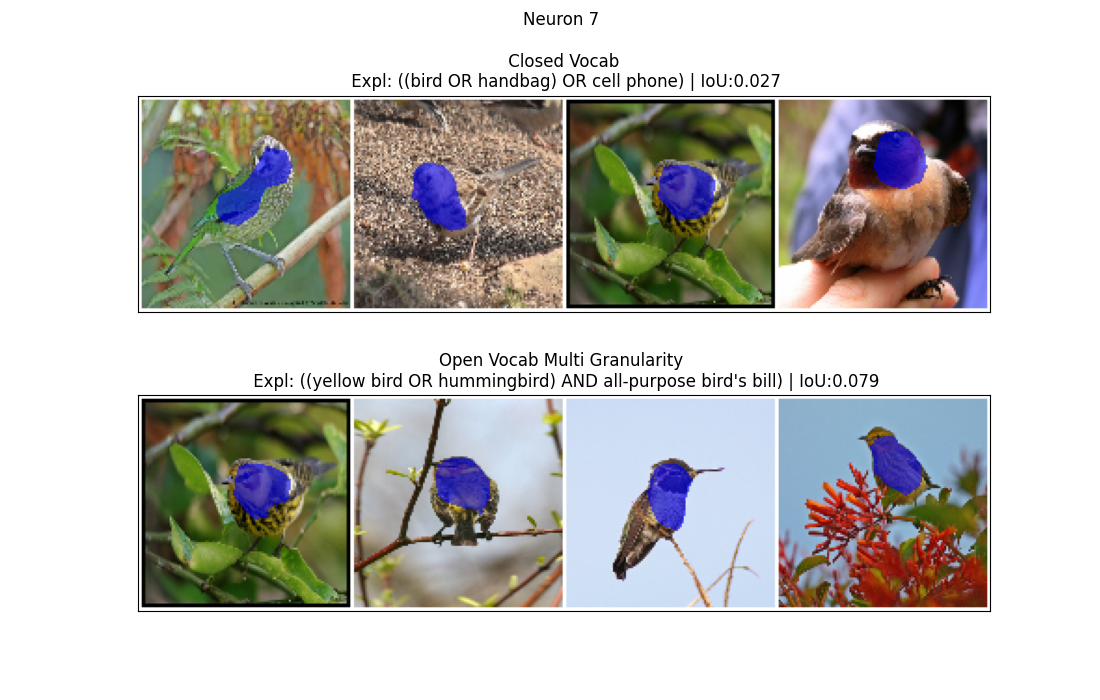}}
    \end{subfloat}
        \begin{subfloat}{
\includegraphics[width=0.7\linewidth]{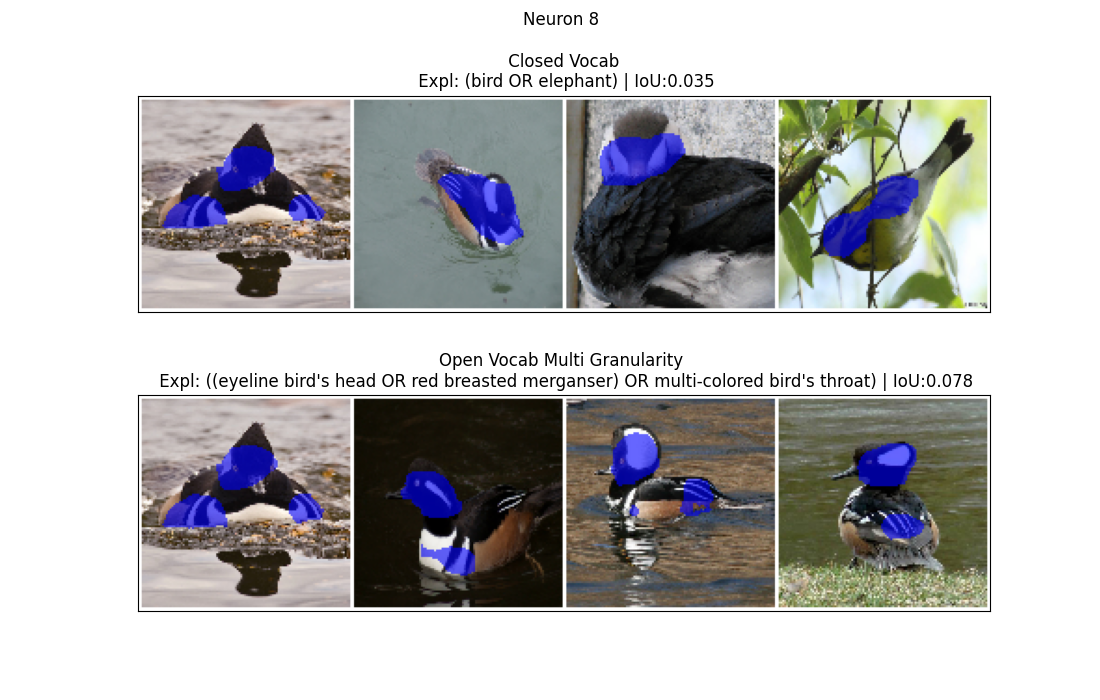}}
    \end{subfloat}

    \caption{Explanations associated with Cluster 5 of neurons from 6 to 8 by the \textit{Closed} approach~\cite{bau2020units} and our framework. In blue are areas of neuron activation within the considered range.}
    \label{fig:neuron6-8}
\end{figure*}

\begin{figure*}
    \centering
            \begin{subfloat}{
\includegraphics[width=0.7\linewidth]{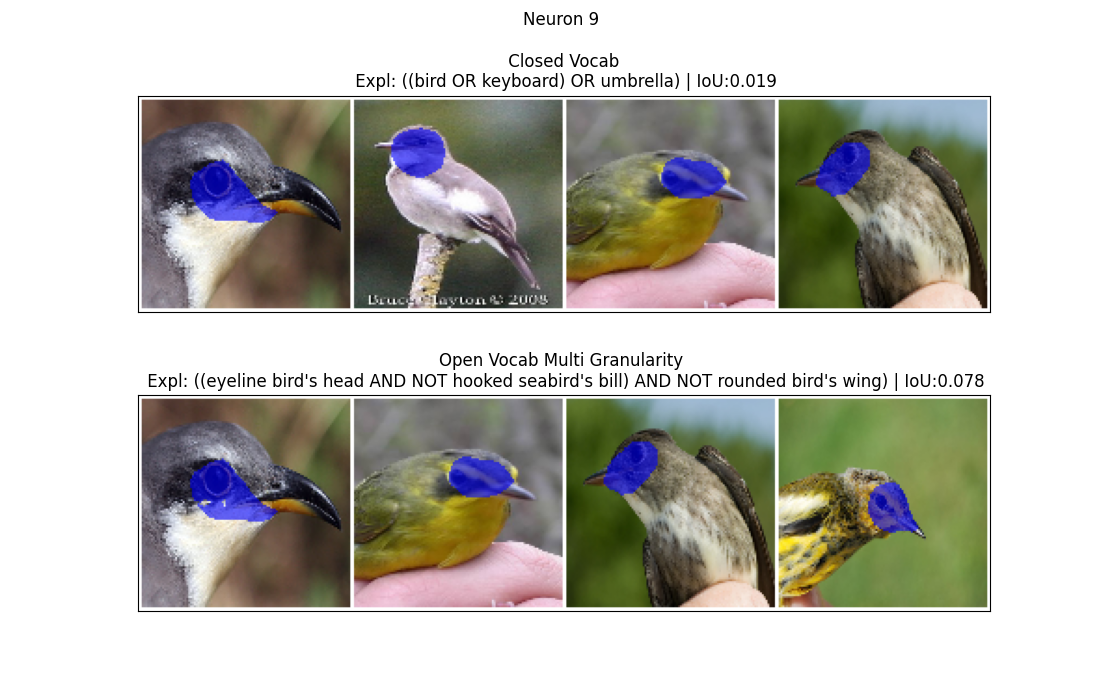}}
    \end{subfloat}
        \begin{subfloat}{
\includegraphics[width=0.7\linewidth]{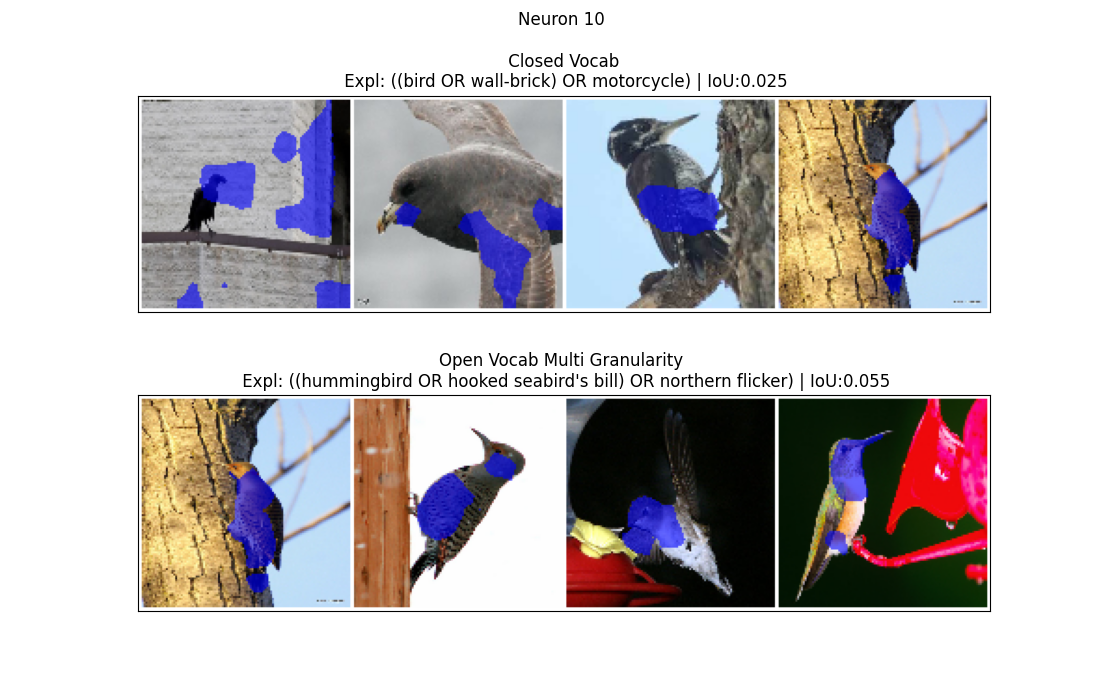}}
    \end{subfloat}
    \begin{subfloat}{
\includegraphics[width=0.7\linewidth]{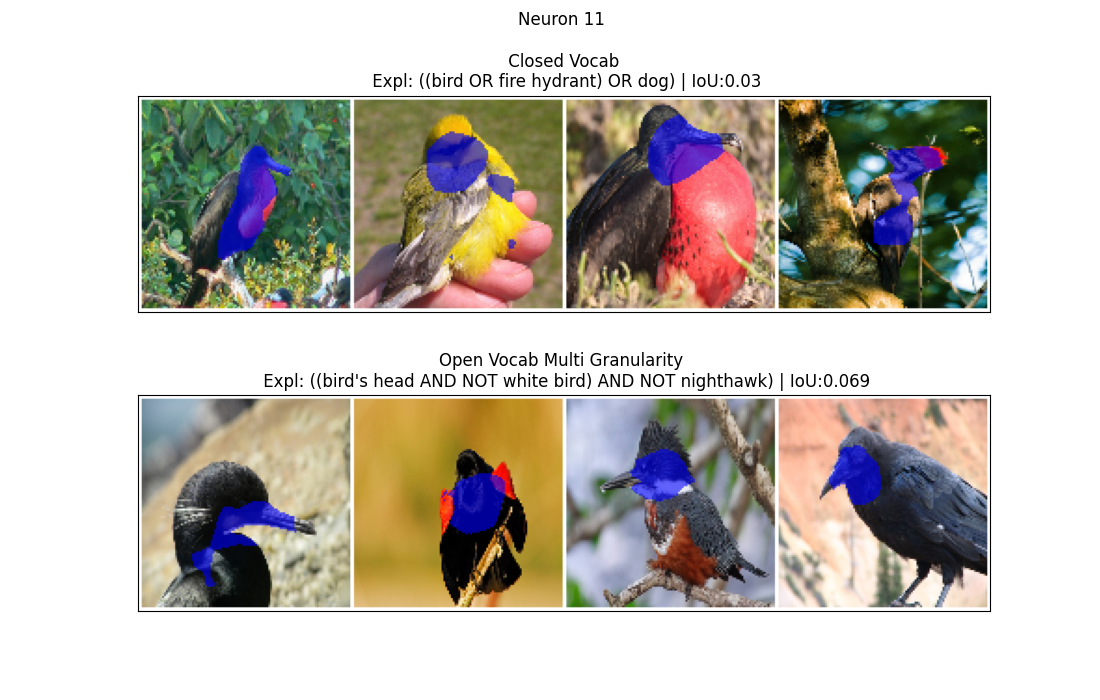}}
    \end{subfloat}

    \caption{Explanations associated with Cluster 5 of neurons from 9 to 11 by the \textit{Closed} approach~\cite{bau2020units} and our framework. In blue are areas of neuron activation within the considered range.}
    \label{fig:neuron9-11}
\end{figure*}

\begin{figure*}
    \centering
        \begin{subfloat}{
\includegraphics[width=0.7\linewidth]{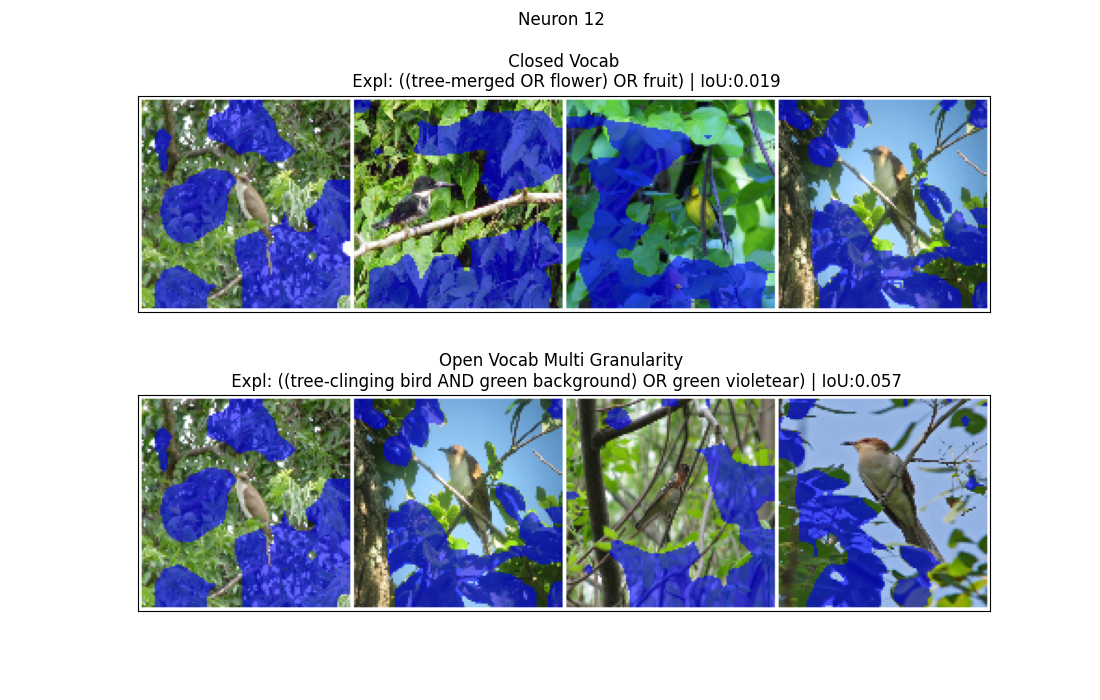}}
    \end{subfloat}
            \begin{subfloat}{
\includegraphics[width=0.7\linewidth]{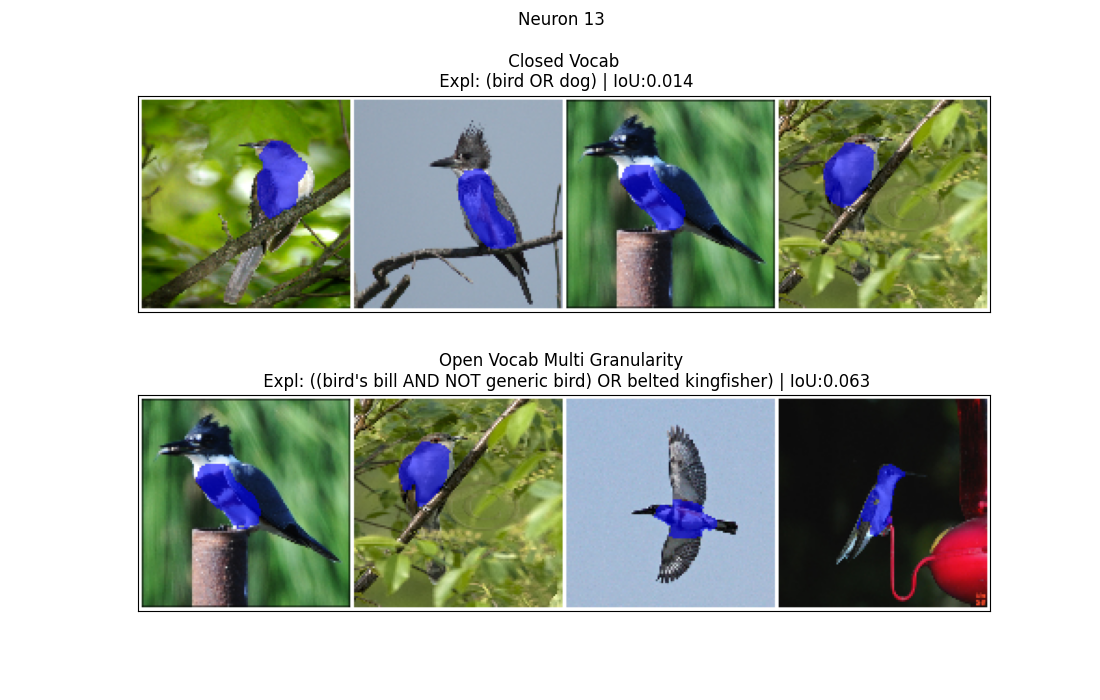}}
    \end{subfloat}
        \begin{subfloat}{
\includegraphics[width=0.7\linewidth]{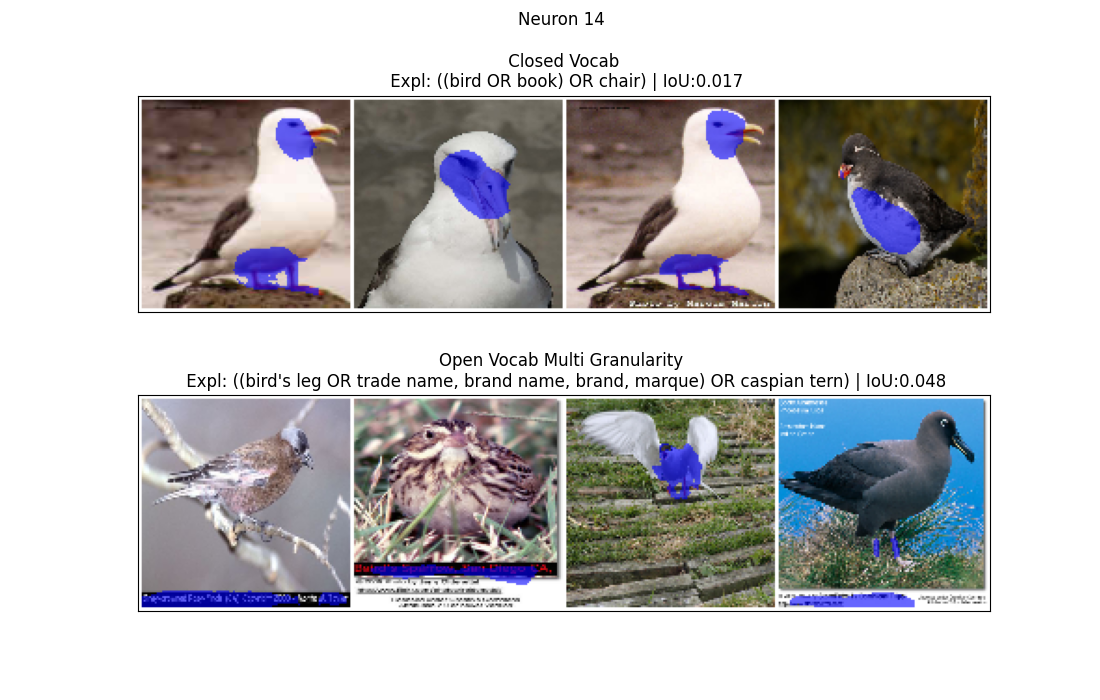}}
    \end{subfloat}

    \caption{Explanations associated with Cluster 5 of neurons from 12 to 14 by the \textit{Closed} approach~\cite{bau2020units} and our framework. In blue are areas of neuron activation within the considered range.}
    \label{fig:neuron12-14}
\end{figure*}

\begin{figure*}
    \centering
            \begin{subfloat}{
\includegraphics[width=0.7\linewidth]{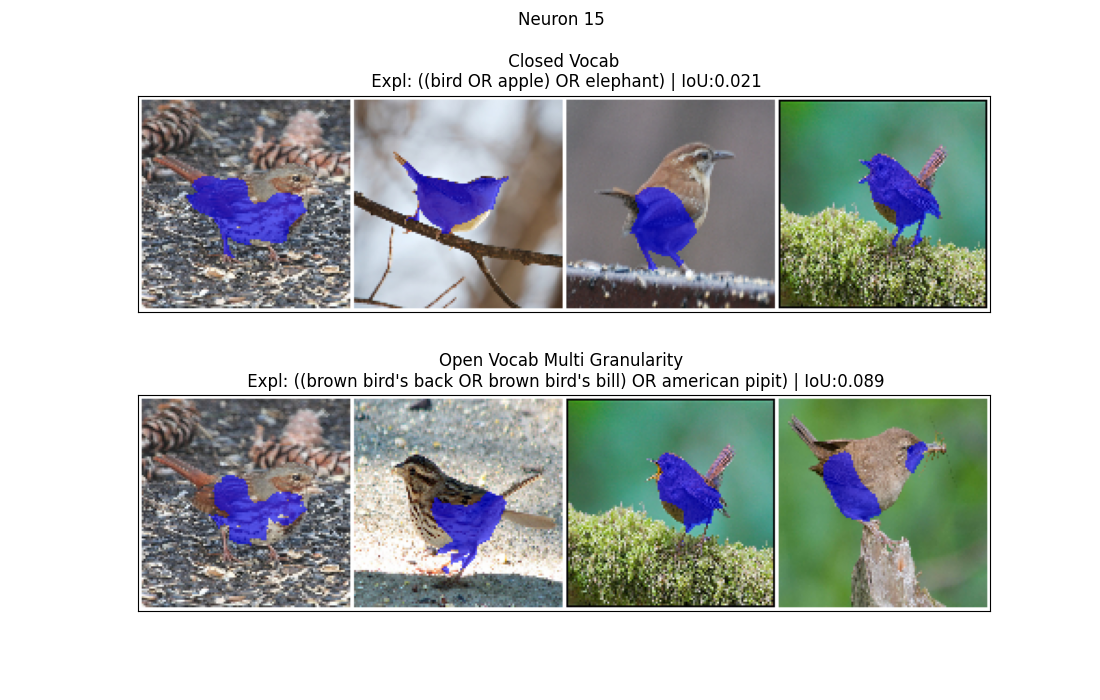}}
    \end{subfloat}
    \begin{subfloat}{
\includegraphics[width=0.7\linewidth]{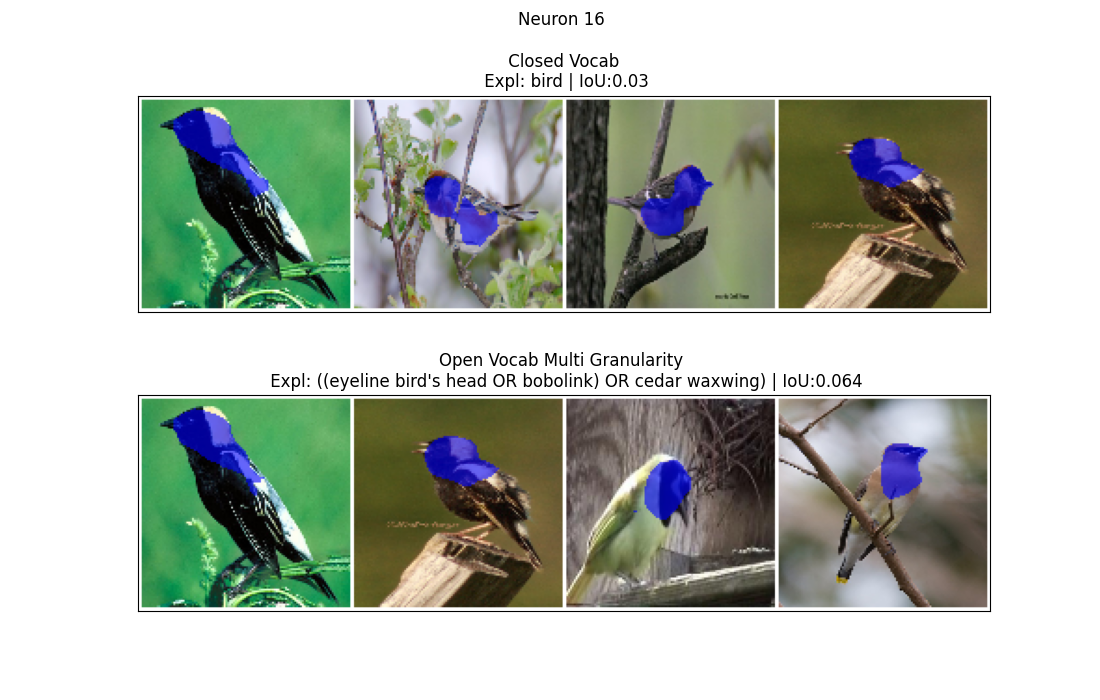}}
    \end{subfloat}
    \begin{subfloat}{
\includegraphics[width=0.7\linewidth]{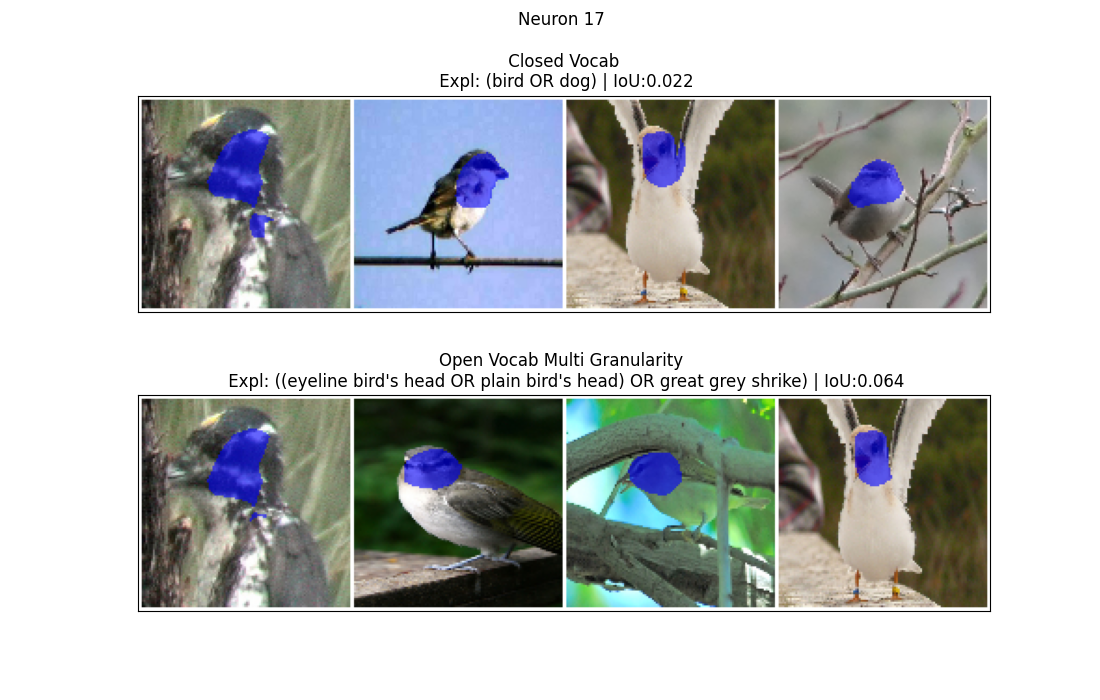}}
    \end{subfloat}
    \caption{Explanations associated with Cluster 5 of neurons from 15 to 17 by the \textit{Closed} approach~\cite{bau2020units} and our framework. In blue are areas of neuron activation within the considered range.}
    \label{fig:neuron15-17}
\end{figure*}

\begin{figure*}
    \centering
        \begin{subfloat}{
\includegraphics[width=0.7\linewidth]{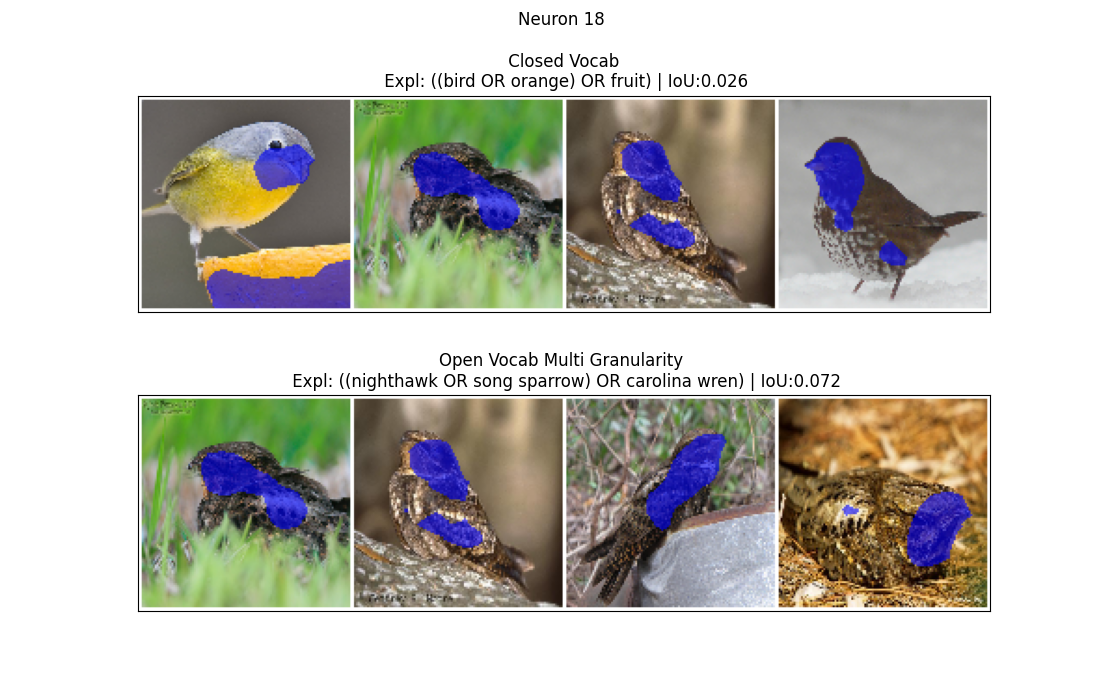}}
    \end{subfloat}
        \begin{subfloat}{
\includegraphics[width=0.7\linewidth]{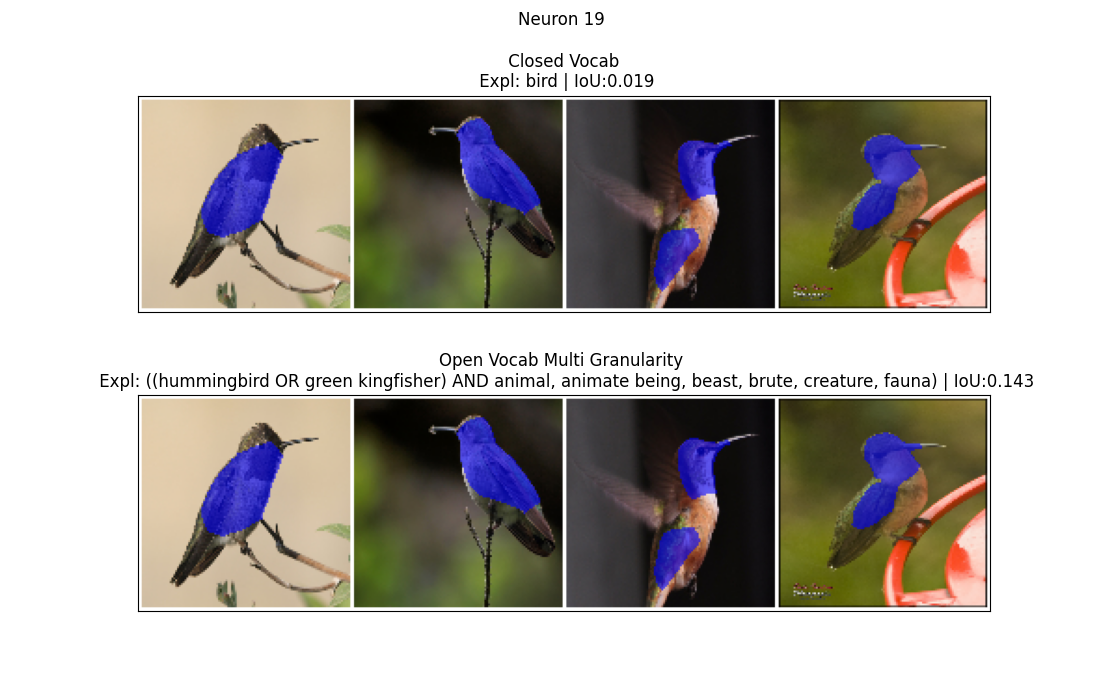}}
    \end{subfloat}
    \caption{Explanations associated with Cluster 5 of neurons from 18 to 19 by the \textit{Closed} approach~\cite{bau2020units}} and our framework. In blue are areas of neuron activation within the considered range.
    \label{fig:neuron18-19}
\end{figure*}

This section includes a visual comparison of the explanations generated by the \textit{Closed} approach and our framework on the CUB dataset. Specifically, we show the explanations generated for the first 20 neurons of the CUB model described in Section 4 for the highest cluster (\Cref{fig:neuron0-2,fig:neuron3-5,fig:neuron6-8,fig:neuron9-11,fig:neuron12-14,fig:neuron15-17,fig:neuron18-19}). 

As noted in the main text, we can observe that the \textit{Closed} approach~\cite{bau2020units} fails to recognize the specific concepts captured by the activation range and its explanations are comparable only when the neuron focuses on background elements or general concepts (e.g., water, sky), thus highlighting the lack of flexibility of this approach.

\end{document}